\documentclass[pdflatex,sn-mathphys-num]{sn-jnl}

\usepackage{graphicx}%
\usepackage{multirow}%
\usepackage{amsmath,amssymb,amsfonts}%
\usepackage{amsthm}%
\usepackage{mathrsfs}%
\usepackage[title]{appendix}%
\usepackage{xcolor}%
\usepackage{textcomp}%
\usepackage{manyfoot}%
\usepackage{nameref}%
\usepackage{booktabs}%
\usepackage{algorithm}%
\usepackage{algorithmicx}%
\usepackage{algpseudocode}%
\usepackage{listings}%

\usepackage{xr}

\makeatletter
\@input{xx.tex}
\makeatother

\definecolor{codegreen}{rgb}{0,0.6,0}
\definecolor{codegray}{rgb}{0.5,0.5,0.5}
\definecolor{codepurple}{rgb}{0.58,0,0.82}
\definecolor{backcolour}{rgb}{0.95,0.95,0.92}

\lstdefinestyle{mystyle}{
    backgroundcolor=\color{backcolour},   
    commentstyle=\color{codegreen},
    keywordstyle=\color{magenta},
    numberstyle=\tiny\color{codegray},
    stringstyle=\color{codepurple},
    basicstyle=\ttfamily\footnotesize,
    breakatwhitespace=false,         
    breaklines=true,                 
    captionpos=b,                    
    keepspaces=true,                 
    numbers=left,                    
    numbersep=5pt,                  
    showspaces=false,                
    showstringspaces=false,
    showtabs=false,                  
    tabsize=2
}

\theoremstyle{thmstyleone}%
\theoremstyle{thmstyletwo}%

\theoremstyle{thmstylethree}%

\begin{document}

\title[Article Title]{ASKCOS: an open source software suite for synthesis planning}


\author[1]{\fnm{Zhengkai} \sur{Tu}}\email{ztu@mit.edu}
\author[2]{\fnm{Sourabh J.} \sur{Choure}}\email{sjchoure@mit.edu}
\author[2]{\fnm{Mun Hong} \sur{Fong}}\email{fong410@mit.edu}
\author[2]{\fnm{Jihye} \sur{Roh}}\email{jroh99@mit.edu}
\author[3]{\fnm{Itai} \sur{Levin}}\email{itail@mit.edu}
\author[4]{\fnm{Kevin} \sur{Yu}}\email{kyu3@mit.edu}
\author[2]{\fnm{Joonyoung F.} \sur{Joung}}\email{jjoung@mit.edu}
\author[2]{\fnm{Nathan} \sur{Morgan}}\email{knathan@mit.edu}
\author[2]{\fnm{Shih-Cheng} \sur{Li}}\email{scli@mit.edu}
\author[2]{\fnm{Xiaoqi} \sur{Sun}}\email{xiaoqis@mit.edu}
\author[2]{\fnm{Huiqian} \sur{Lin}}\email{linhq@mit.edu}
\author[2]{\fnm{Mark} \sur{Murnin}}\email{murninm@mit.edu}
\author[2]{\fnm{Jordan P.} \sur{Liles}}\email{jliles24@mit.edu}
\author[5]{\fnm{Thomas J.} \sur{Struble}}\email{Thomas.Struble@bms.com}
\author[6]{\fnm{Michael E.} \sur{Fortunato}}\email{mike.fortunato@novartis.com}
\author{\fnm{Mengjie} \sur{Liu}\textsuperscript{2,}\footnote[2]{Current affiliation: AstraZeneca. Work done while at MIT.}}\email{mjliu@mit.edu}
\author[2]{\fnm{William H.} \sur{Green}}\email{whgreen@mit.edu}
\author[2]{\fnm{Klavs F.} \sur{Jensen}}\email{kfjensen@mit.edu}
\author*[1,2]{\fnm{Connor W.} \sur{Coley}}\email{ccoley@mit.edu}

\affil[1]{\orgdiv{Department of Electrical Engineering and Computer Science}, \orgname{Massachusetts Institute of Technology}, \orgaddress{\street{77 Massachusetts Ave}, \city{Cambridge}, \state{MA}, \postcode{02139}, \country{USA}}}

\affil[2]{\orgdiv{Department of Chemical Engineering}, \orgname{Massachusetts Institute of Technology}, \orgaddress{\street{77 Massachusetts Ave}, \city{Cambridge}, \state{MA}, \postcode{02139}, \country{USA}}}

\affil[3]{\orgdiv{Department of Biological Engineering}, \orgname{Massachusetts Institute of Technology}, \orgaddress{\street{77 Massachusetts Ave}, \city{Cambridge}, \state{MA}, \postcode{02139}, \country{USA}}}

\affil[4]{\orgdiv{Center for Computational Science and Engineering}, \orgname{Massachusetts Institute of Technology}, \orgaddress{\street{77 Massachusetts Ave}, \city{Cambridge}, \state{MA}, \postcode{02139}, \country{USA}}}

\affil[5]{\orgname{Bristol Myers Squibb}, \orgaddress{\street{250 Water Street, \city{Cambridge}, \state{MA}, \postcode{02141}, \country{USA}}}}

\affil[6]{\orgname{Novartis Institutes for BioMedical Research, Inc.}, \orgaddress{\street{250 Massachusetts Avenue}, \city{Cambridge}, \state{MA}, \postcode{02139}, \country{USA}}}

\abstract{The advancement of machine learning and the availability of large-scale reaction datasets have accelerated the development of data-driven models for computer-aided synthesis planning (CASP) in the past decade. Here, we detail the newest version of ASKCOS, an open source software suite for synthesis planning that makes available several research advances in a freely available, practical tool. Four one-step retrosynthesis models form the basis of both interactive planning and automatic planning modes. Retrosynthetic planning is complemented by other modules for feasibility assessment and pathway evaluation, including reaction condition recommendation, reaction outcome prediction, and auxiliary capabilities such as solubility prediction and quantum mechanical descriptor prediction. ASKCOS has assisted hundreds of medicinal, synthetic, and process chemists in their day-to-day tasks, complementing expert decision making. It is our belief that CASP tools like ASKCOS are an important part of modern chemistry research, and that they offer ever-increasing utility and accessibility.}

\keywords{synthesis planning, open source software, data-driven predictive chemistry}



\maketitle

\section{Introduction}\label{introduction}


Synthesis planning describes a broad category of approaches for selecting experimental pathways and procedures during target-oriented synthesis.  While planning a synthesis campaign may require significant chemistry expertise and benefits from the years of training that many experienced chemists undergo, the well-defined yet combinatorially complex nature of synthesis planning renders this task particularly amenable to algorithmic reasoning. Formally, computer-aided synthesis planning (CASP) integrates a variety of computational methodologies that assist chemists with different tasks in this process, including identifying viable synthetic routes through retrosynthetic analysis, recommending reaction conditions, and predicting reaction outcomes.

Since the 1960s, chemists have sought to encode the rules of organic synthesis into automated computational systems~\citep{corey_computer-assisted_1969,warr_short_2014}. Early CASP tools generally relied on expert-encoded reaction rules and heuristics for making suggestions. In particular, for one-step retrosynthetic analysis~\citep{corey_robert_1988}, expert systems such as LHASA~\citep{corey_computer-assisted_1972} and SECS~\citep{wipke_simulation_1978} made use of reaction templates that encode chemical reaction rules based on molecular pattern matching; AIPHOS~\citep{funatsu_computer-assisted_1988} and WODCA~\citep{gasteiger_wodca_1990} were among the first to combine retrosynthesis, reaction condition suggestion, and product prediction into an integrated system. More recent advancements include tools like Chematica (now Synthia)~\citep{grzybowski_chematica_2018}, which leverages modern computational capabilities alongside expert-curated rules and heuristics to propose transformations, generating synthetic pathways for complex molecules that have been successfully implemented in the laboratory~\citep{klucznik_efficient_2018,mikulak-klucznik_computational_2020}. 


With the development of machine learning and the availability of reaction datasets containing millions of entries, there has been renewed interest in CASP with data-driven approaches~\citep{tu_predictive_2023,schwaller_machine_2022}. Many data-driven models for one-step retrosynthesis have been developed with formulations based on reaction template prediction~\citep{dai_retrosynthesis_2019,chen_deep_2021} or retrieval~\citep{coley_computer-assisted_2017,xie_retrosynthesis_2023}, machine translation~\citep{tu_permutation_2022,tetko_state---art_2020,zhong_root-aligned_2022}, graph edit prediction~\citep{sacha_molecule_2021,somnath_learning_2021}, as well as other generative models~\citep{igashov_retrobridge_2023,gainski_retrogfn_2024}. These one-step predictors have been integrated with various tree-search algorithms to navigate through the network of hypothetical reactions to identify synthetic pathways in which all starting materials are accessible~\citep{heifets_construction_2012,segler_planning_2018,kishimoto_depth-first_2019,schwaller_predicting_2020,chen_retro_2020}. Machine learning models have also been applied to other elements of synthesis planning such as reaction condition recommendation~\citep{gao_using_2018,maser_multilabel_2021,kwon_generative_2022,qian_predictive_2023,chen_enhancing_2024} and reaction outcome prediction~\citep{schwaller_molecular_2019,tu_permutation_2022,coley_graph-convolutional_2019,sacha_molecule_2021,bradshaw_generative_2019,bi_non-autoregressive_2021}. Like in typical supervised learning settings, these approaches are trained on historical reaction data and try to generalize to unseen targets.


Parallel to the development of \emph{algorithms} for CASP has been the emergence of many software tools. Among many proprietary examples are LHASA~\citep{corey_computer-assisted_1972}, Synthia~\citep{grzybowski_chematica_2018}, Chemical.ai~\citep{Chemical.ai}, IBM's RXN for Chemistry~\citep{RXNforChemistry}, Spaya by Iktos~\citep{Spaya}, Manifold by PostEra~\citep{Manifold}, Molecule.one~\citep{Molecule.one}, SciFinder (integrating technology from ChemPlanner~\citep{ChemPlanner} and Molecule.one), and Reaxys (integrating technology from Iktos and Pending.AI)~\citep{ReaxysRetro}; these are complemented by ASKCOS~\citep{coley_robotic_2019}, AiZynthFinder~\citep{genheden_aizynthfinder_2020}, and Syntheseus~\citep{maziarz_re-evaluating_2024} as open source offerings. ASKCOS is distinct in its breadth and attempt to cover a wide range of different tasks in synthesis planning, rather than focusing mostly on retrosynthetic analysis. 

The ASKCOS software has been in development since 2016. It was originally designed to use algorithmically-extracted templates and simple tree search algorithms such as depth-first search or best-first search using heuristics. Over the past eight years, it has been expanded significantly, deployed, used, and evaluated by the community and, in particular, at dozens of pharmaceutical and chemical companies within and beyond the Machine Learning for Pharmaceutical Discovery and Synthesis (MLPDS) consortium~\citep{MLPDS}. Chemists have integrated various modules of ASKCOS into their workflows for molecule and route ideation, at times adapting individual components into proprietary design tools~\citep{struble_current_2020}. While ASKCOS has proven useful for many chemists and researchers~\citep{coley_robotic_2019,struble_current_2020,levin_merging_2022,soukaina_design_2024,avila_chemistry_2024,fromer_algorithmic_2024,sankaranarayanan_computer-assisted_2023,koscher_autonomous_2023,nambiar_bayesian_2022,mahjour_ideation_2024,seierstad_novel_2021,qi_optimized_2023,pasquini2023linchemin}, it has only been formally (and briefly) described in the context of robotic flow chemistry in \citet{coley_robotic_2019}.

\begin{figure}[h!]
\centering
\includegraphics[width=1.0\textwidth]{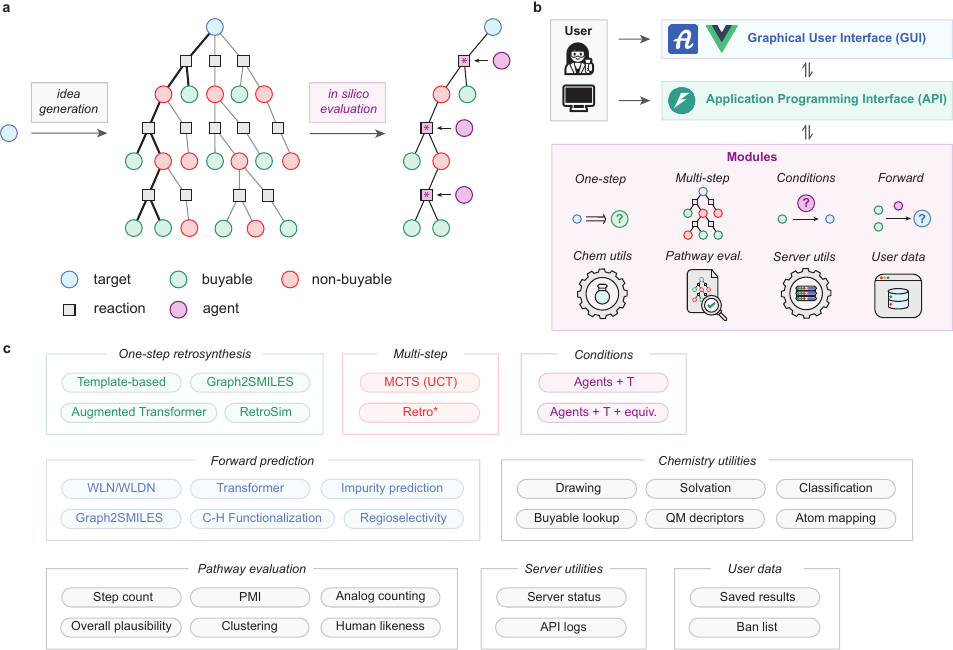}
\caption{ASKCOS overview. \textbf{a)} The typical target-oriented synthesis workflow for which ASKCOS is designed. A target molecule (blue circle) is recursively expanded retrosynthetically until buyable starting materials (green circles) are reached. Agents/conditions can be predicted for each proposed reaction, which can then undergo further evaluation, e.g., to anticipate reaction products. \textbf{b)} High-level flowchart for ASKCOS usage. A user can either interact via the graphical user interface or programmatic endpoints, which call various prediction modules. \textbf{c)} Summary of modules available in ASKCOS, which fall under themes including one-step retrosynthesis (green), multi-step search (red), condition recommendation (purple), reaction outcome prediction (blue), as well as utilities and supplementary capabilities (black). The modularity of the software design enables the straightforward extension of functionality (e.g., to new data-driven models) as they are developed in a research setting and made production-ready.}\label{fig_overview}
\end{figure}

Here, we report the latest version of the open source ASKCOS software suite, reflecting the variety of new functionalities and improvements that have contributed to its growth. At a high level, ASKCOS has two modes of operation for its core retrosynthesis functionality: an interactive mode with the Interactive Path Planner (IPP) and an automatic mode with the Tree Builder. Both modes can use multiple one-step strategies individually or simultaneously to combine the strengths of each. In addition to the retrosynthesis functionality, ASKCOS provides partial solutions to many adjacent tasks within synthesis planning. The open source and modular nature of ASKCOS also allows for easy access to various prediction modules, either via the web-based interface or via application programming interfaces (APIs), which are available under permissive MIT licenses. In the rest of this article, we present these functionalities in terms of both their scientific aspects and their usage in ASKCOS (Figure \ref{fig_overview}).

\section{Results}\label{results}

\subsection{One-step retrosynthetic expansion}\label{results_one_step}

The one-step retrosynthetic expansion engine lies at the core of retrosynthetic analysis in ASKCOS. Given a target molecule, a list of candidate precursors is first predicted with user-specified one-step model(s). A fast plausibility filter based on the in-scope filter described by \citet{segler_planning_2018} removes unlikely precursors, and the remaining list is reranked based on buyability and complexity (e.g., by heavy atom count, ring count, or by SCScore~\citep{coley_scscore_2018}). Thereafter, the candidates undergo a series of optional post-processing steps. Because several predictions from the list of candidate precursors may correspond to highly similar strategies (e.g., differing only by leaving groups), precursors can be clustered based on their structures or reaction classes to return a diverse set of suggestions to the user; atom mapping and template extraction can be performed on reactions proposed by template-free model(s), which help check the chemical validity of the suggestions; precursors can be filtered or grouped based on which atoms are involved in the reaction centers; selectivity checks can automatically flag reactions with potential regioselectivity issues.

\begin{figure}[h!]
\centering
\includegraphics[width=1.0\textwidth]{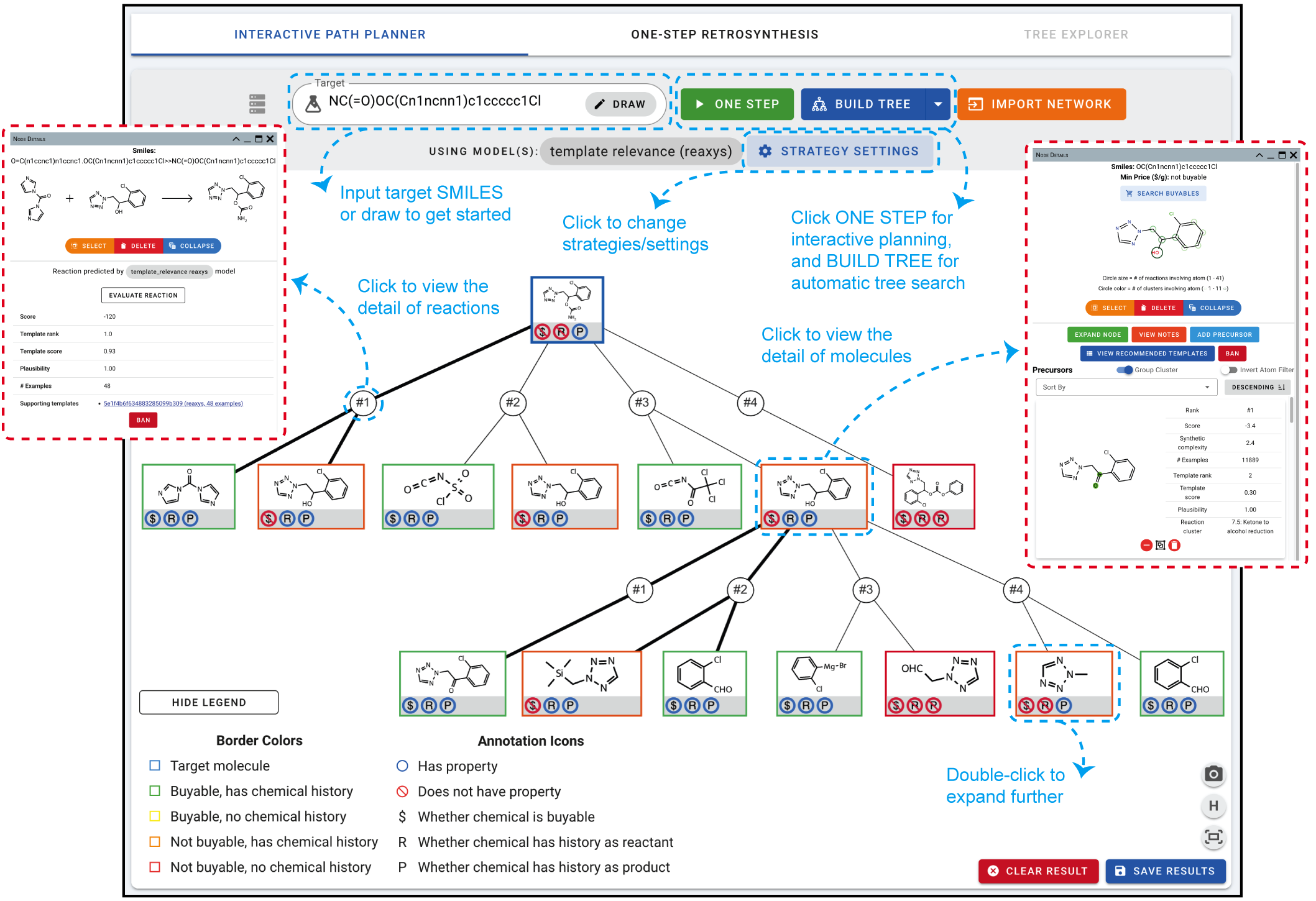}
\caption{Annotated screenshot of the interactive planning canvas in ASKCOS. The input target molecule shown is cenobamate, a drug for partial-onset seizures, defined by entering the SMILES string \texttt{NC(=O)OC(Cn1ncnn1)c1ccccc1Cl} in the search bar or using an external name resolver if permitted by security policies. An interactive retrosynthetic search can be initiated with the \texttt{ONE STEP} button. An automated multi-step retrosynthetic search can be initiated with the \texttt{BUILD TREE} button. Search settings for both modes can be set with the \texttt{STRATEGY SETTINGS} button. Here, results are shown for an interactive search performed using the template-based single step model trained on the Reaxys dataset. The top 4 suggestions by the model are displayed on the canvas. A legend for the frame colors is displayed at the bottom left of the page. The displayed two-tier tree was generated by following the initial single step expansion of the target molecule with the expansion of one of the suggested precursors. Additional retrosynthetic expansions can be performed by double clicking on a chemical node. The chemical node info window provides users with additional information and actions (e.g., adding a comment for that chemical or banning it from future suggestions). If a one-step model has already been used to suggest precursors for that chemical, the node info window displays (right inset, red dashed frame) suggested precursors and allows users to sort or filter these suggestions using a variety of metrics (e.g., score, synthetic complexity, number of rings). A reaction node info window (left inset, red dashed frame)  provides users with information about the suggested reaction, including predicted scores and database precedents if they exist. The window also allows users to perform actions on the node such as removing or hiding it. Results can be saved, exported, or reorganized into different graphical views using additional tools in the bottom-right of the canvas.}\label{fig_ipp}
\end{figure}

\subsection{Interactive planning with the template relevance model}\label{results_ipp_template}


The use of reaction templates to suggest retrosynthetic disconnections has remained a popular choice since its origination in the early CASP tools of the 1960s~\citep{corey_computer-assisted_1969,wipke_simulation_1978,corey_computer-assisted_1972}. Template-based models within ASKCOS follow the neural-symbolic approach of \citet{segler_neural-symbolic_2017} wherein a policy network is trained to rank which templates appear most \emph{strategic} and \emph{chemically plausible} given the target. ASKCOS contains a variety of such models trained to use template sets derived from reactions in Pistachio~\citep{Pistachio}, CAS Content~\citep{CASContent}, USPTO~\citep{lowe_extraction_2012}, and Reaxys~\citep{Reaxys} using RDChiral~\citep{coley_rdchiral_2019} (see Section \nameref{methods}). Specialized models trained on enzymatic reaction data in BKMS~\citep{BKMS} and a specialized ``ring-breaker'' Pistachio model are also available and described in \citet{levin_merging_2022} and \citet{thakkar_ring_2020}, respectively. Because template-based models propose candidate precursors using templates extracted from published reactions, model suggestions can be traced back to reaction precedents and are therefore somewhat explainable.


Figure \ref{fig_ipp} shows a sample planning step with the original template-based model trained on Reaxys in 2016 in the \emph{Interactive Path Planner (IPP)}. The target is specified by a simplified molecular input line entry system (SMILES) string~\citep{weininger_smiles_1988} but can also be defined by drawing its structure with Ketcher or by common name using the PubChem API~\citep{PubChemAPI}. One-step expansion can then be triggered by pressing the green button. The top few (5 by default) suggestions will be added to the canvas as the child nodes of the target, with circles representing the reaction nodes and rectangles representing the molecule nodes. Clicking on the nodes display more context-specific details (shown as inset figures in red dotted lines in Figure \ref{fig_ipp}) and provide access to additional features, such as banning the reaction/molecule (thereby preventing them from appearing in future searches) or deleting/collapsing all child nodes of that node. Reaction nodes report the template scores (i.e., the template probabilities returned by the model), plausibility (evaluated by the fast binary filter as mentioned in Section \nameref{results_one_step}), and links to template details (which include links to the reaction precedents that the templates were extracted from). 

Each reaction can be analyzed further with the \texttt{EVALUATE REACTION} button, such as recommending conditions for the reaction or predicting reaction outcomes (see Sections \nameref{results_condition} and \nameref{results_forward} for details). Molecule nodes report their price and are additionally color-coded to reflect whether they are buyable or appear in known reactions stored in a reference database, as explained in the bottom-left IPP legend. Molecule node-specific features are also available, including expanding the nodes, adding and saving notes, and viewing recommended templates for the molecule. Once molecule nodes are expanded, their node details additionally list \emph{all} predicted precursors. These precursors can then be sorted according to different criteria (e.g., heuristic scores, synthetic complexity, number of precedents), grouped into clusters (if already clustered as mentioned in Section \nameref{results_one_step}), or filtered by reaction center by selecting the green-circled atom in the rendering of the molecule. Specific precursors can be added or removed from the canvas with the green \texttt{+} button or the red \texttt{-} button, respectively. 

After this first expansion of the target molecule, it is up to the user to decide which molecule node(s) to expand further, hence the \emph{interactive} nature of this view. For instance, users can choose a non-buyable molecule to expand next and continue this process until an appropriate pathway is identified. Precursors can be added to the molecule manually with the \texttt{ADD PRECURSOR} button if a user wishes to supplement model predictions with their own ideas. The explored network and notes can be saved to a user's profile for future access or exported as a JSON file for offline processing. 

\subsection{Interactive planning with multiple models, including template-free models}\label{results_ipp_multiple}


There is an abundance of one-step retrosynthetic models reported in the past several years that forgo the use of templates and instead learn to predict reactant structures from product structures in a more flexible end-to-end manner. ASKCOS currently contains four categories of one-step strategies, including Transformer~\citep{lin_automatic_2020,tetko_state---art_2020,schwaller_molecular_2019}, Graph2SMILES~\citep{tu_permutation_2022}, Retrosim~\citep{coley_computer-assisted_2017}, and the aforementioned template relevance strategy~\citep{segler_neural-symbolic_2017}. Transformer and Graph2SMILES use template-free approaches and model retrosynthesis as SMILES-to-SMILES and graph-to-SMILES translation tasks, respectively. Retrosim offers a retrieval-based, learning-free approach in which reactions are suggested based on analogy to most-similar precedents. Each model may exhibit distinct strengths and failure modes not reflected in their quantitative performance on standard benchmark tasks (Table~\ref{tab1_one_step}). For this reason, ASKCOS supports the consolidation of recommendations from multiple strategies; when multiple models recommend the same precursors, this can be interpreted as a sign of confidence in that recommendation.


Mixing and matching different one-step strategies is performed within the \texttt{STRATEGY SETTINGS} menu (Figure \ref{fig_ipp}). Different strategies are queried in sequence and combined results are de-duplicated before appearing on the page. Each strategy has its own settings, such as the maximum number of templates to be applied for the template relevance model and the training set to use. When viewing node details, suggested reactions from template-free and Retrosim models contain different metadata (e.g., Retrosim results display the reaction precedents if that information is included on deployment). When multiple models predict the same precursor, the metadata from all models are merged to show all template and/or reaction precedent information where applicable. Most of these strategies can be retrained using new reaction databases, e.g., proprietary collections from an internal electronic lab notebook system, which may provide better coverage of different reaction and substrate types.

\begin{table}[h!]
\caption{Previously reported top-k accuracies (\%) of one-step models on two commonly-used benchmark datasets: USPTO-50k~\citep{schneider_whats_2016} and USPTO-full~\citep{dai_retrosynthesis_2019}.}\label{tab1_one_step}
\begin{tabular*}{\textwidth}{@{\extracolsep\fill}llcccccc}
\toprule
& & \multicolumn{3}{@{}c@{}}{USPTO-50k} & \multicolumn{3}{@{}c@{}}{USPTO-full} \\
\cmidrule{3-5}\cmidrule{6-8}
Model name & Type & Top 1 & Top 10 & Ref. 
& Top 1 & Top 10 & Ref. \\
\midrule
Template relevance
& Template-based    
& 45.2 
& 83.5 &~\citep{seidl_improving_2022} 
& 35.8 
& 60.8 &~\citep{dai_retrosynthesis_2019} \\
Retrosim
& Template-based    
& 37.3 
& 74.1 &~\citep{coley_computer-assisted_2017}
& 32.8
& 56.1 &~\citep{dai_retrosynthesis_2019} \\
Transformer w/o aug. 
& Template-free     
& 43.1 
& 78.7 &~\citep{lin_automatic_2020}
& 42.9 
& 66.8 &~\citep{zhu_dual-view_2023} \\
Transformer w/ aug. 
& Template-free     
& 53.2 
& 85.2 &~\citep{tetko_state---art_2020}
& 44.4 
& 73.3 &~\citep{tetko_state---art_2020} \\
Graph2SMILES 
& Template-free     
& 52.9 
& 79.5 &~\citep{tu_permutation_2022}
& 45.7 
& 63.4 &~\citep{tu_permutation_2022} \\
\botrule
\end{tabular*}
\footnotetext{Note: for the template relevance model, accuracies on the original NeuralSym model are reported. Best values from multiple references are recorded. Transformer w/ aug. refers to the variant where equivalent reaction SMILES are used to \emph{augment} the training set, which is a known empirical technique for boosting accuracy~\citep{bjerrum_smiles_2017}. Due to the vast amount of literature related to Transformer and SMILES augmentation, we did not attempt to do an exhaustive literature survey for Transformer-based models.}
\end{table}

\subsection{Automatic multi-step planning with the Tree Builder}\label{results_tree_builder}

\begin{figure}[h!]
\centering
\includegraphics[width=1.0\textwidth]{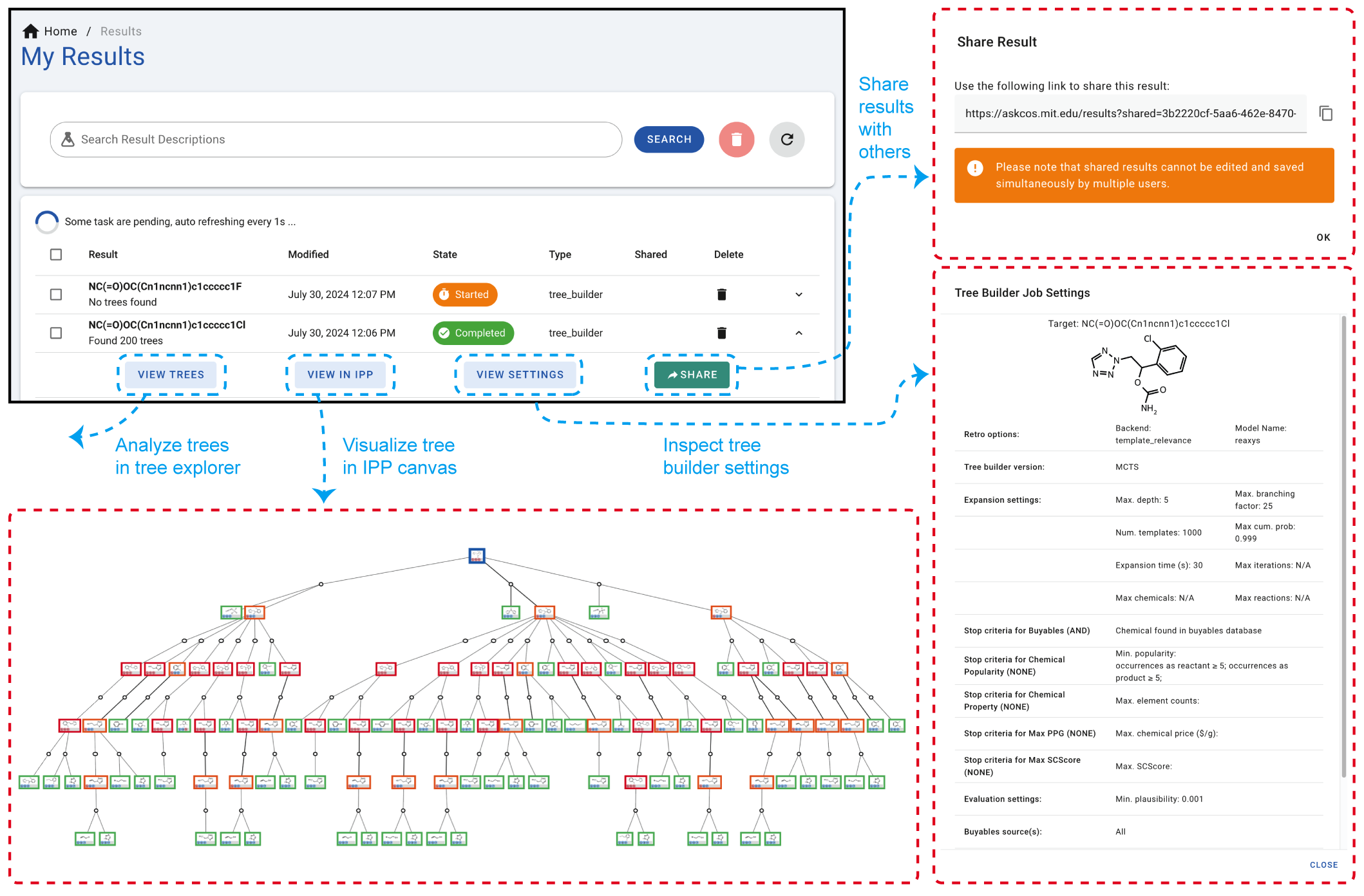}
\caption{Annotated screenshot of tree search results in ASKCOS. The Tree Builder job can be initiated with the \texttt{BUILD TREE} button on the interactive path planning page. In \texttt{My Results} page, the status for the Tree Builder job is shown as \texttt{Started}. Once the job is complete, the status changes to \texttt{Completed}, and the trees found can be visualized/analyzed in the Tree Explorer (\texttt{VIEW TREES}) or the IPP canvas (\texttt{VIEW IN IPP}). The settings used for the Tree Builder job can be viewed with the \texttt{VIEW SETTINGS} button, and the results can be shared with others with the \texttt{SHARE} button. Here, the Tree Builder job for the target molecule \texttt{NC(=O)OC(Cn1ncnn1)c1ccccc1Cl} is performed using the MCTS algorithm with the template-based single step model trained on the Reaxys dataset, using a maximum depth of 5, a maximum branching factor of 25, and an expansion time of 30s (right bottom inset, red dashed frame). The results can be visualized in the IPP canvas {(left bottom inset, red dashed frame)}.}\label{fig_tree_results}
\end{figure}


In addition to interactive planning where users guide the selection of which molecule nodes to expand, \emph{automatic} planning can be more convenient, particularly when there are many target molecules of interest. Retrosynthetic searches can be run for thousands of targets through asynchronous requests, albeit at the expense of losing explicit control on the direction of expansion. Formally, automatic multi-step planning has been formulated as tree search or graph search problems. In each iteration of the search, a molecule is selected for one-step retrosynthetic expansion. New hypothetical reactions and their corresponding reactants are added to the search tree. This process is repeated until some termination criterion is reached, for example, until a synthetic pathway is found in which all starting materials are buyable. The well-known baseline algorithms include Monte Carlo tree search (MCTS)~\citep{segler_planning_2018,lin_automatic_2020}, Proof Number Search (PNS)~\citep{heifets_construction_2012,kishimoto_depth-first_2019} and A* Search~\citep{chen_retro_2020} which tend to be inspired by other artificial intelligence applications such as AlphaGo~\citep{silver_mastering_2016}. More recent work in the field on such search algorithms has mainly focused on improvement of selection policies, for example, with reinforcement learning~\citep{liu_retrosynthetic_2023, yu_grasp_2022,schreck_learning_2019, wang_towards_2020}, supervised learning~\citep{xie_retrograph_2022,zhao_efficient_2024, hong_retrosynthetic_2023}, or chemical heuristics~\citep{schwaller_predicting_2020, kreutter_multistep_2023}. A number of studies have also sought to improve the expansion policy by focusing on the one-step model in the context of multi-step planning~\citep{liu_fusionretro_2023,kim_self-improved_2021}. Others have approached synthesis planning outside of the typical tree search formulation, such as through sequence generation~\citep{shee_directmultistep_2024} or as a bidirectional search~\citep{yu_double-ended_2024}.

ASKCOS supports automatic planning through the \emph{Tree Builder} module which currently provides two search algorithms, MCTS~\citep{segler_planning_2018} and Retro*~\citep{chen_retro_2020}. Either can be selected from \texttt{STRATEGY SETTINGS}. After specifying the target in the IPP canvas, the user can click the \texttt{BUILD TREE} button in Figure \ref{fig_ipp} to initiate a Tree Builder job, which will run asynchronously in the background. A pop-up window will inform the user when the results are ready and available for viewing. All tree results can be found under the \texttt{My Results} page as shown on the top left corner of Figure \ref{fig_tree_results}, where the status of any result entry will change from \texttt{started} to \texttt{completed} once the Tree Builder job is finished. From here, the user can visualize the result either as a full retrosynthesis tree in the IPP canvas, or as individual routes in the Tree Explorer (see Section \nameref{results_pathway_analysis} for further analysis). The settings used for the job can be inspected, and the result can be shared via the green sharing button in Figure \ref{fig_tree_results}.

\subsection{Reaction condition recommendation}\label{results_condition}

\begin{figure}[h!]
\centering
\includegraphics[width=1.0\textwidth]{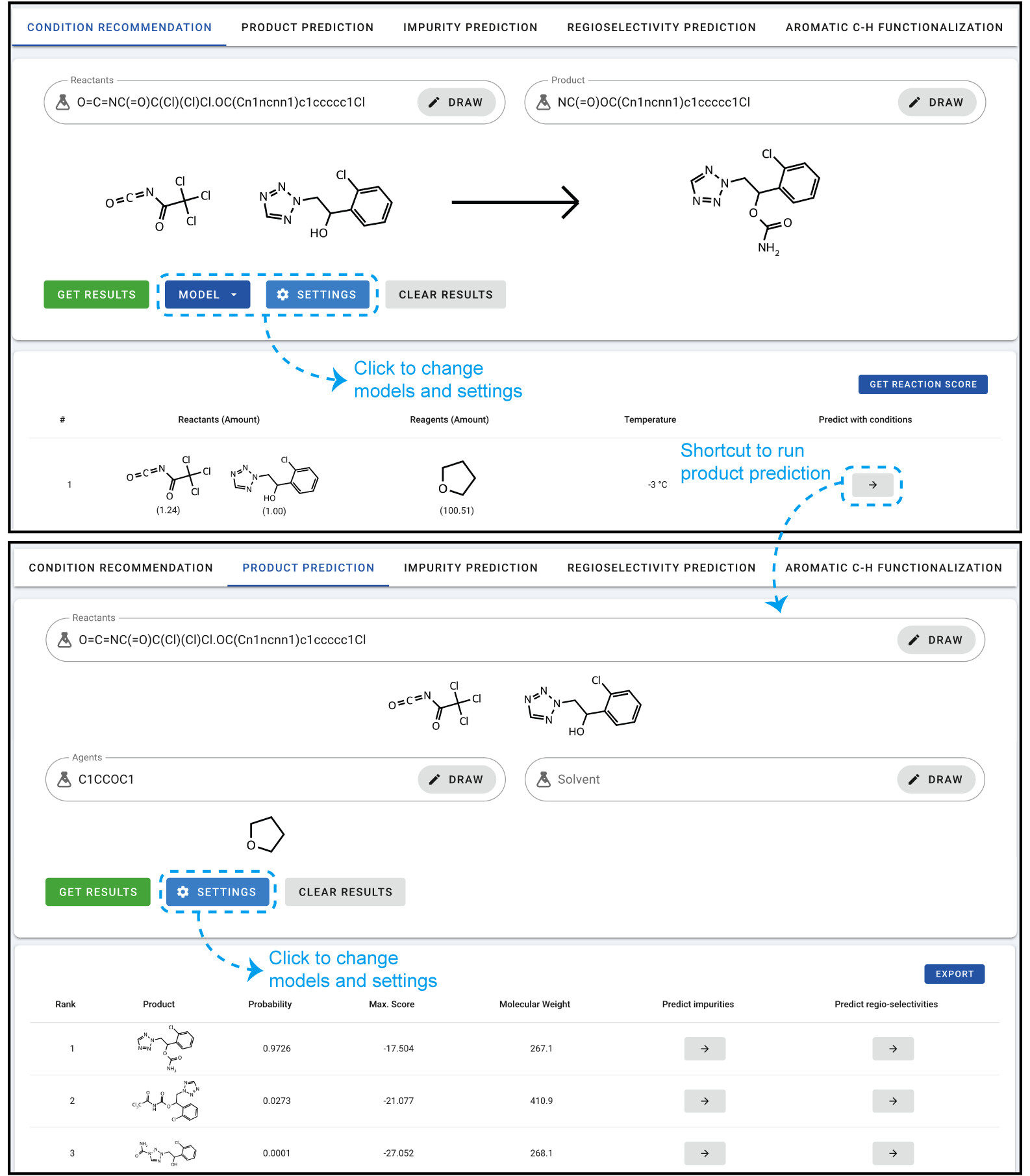}
\caption{Annotated screenshot of the condition recommender (top) and the forward predictor (bottom) in ASKCOS. In the condition recommender, the SMILES strings of reactants and products are input in the \texttt{Reactants} and \texttt{Products} panels. Condition recommendation can be initiated with the \texttt{GET RESULTS} button. Prediction models and their training set can be set with the \texttt{MODEL} and \texttt{SETTINGS} buttons, respectively. Each proposed condition has shortcuts to run product prediction. In the forward predictor, the SMILES strings of reactants and agents are pre-populated using the shortcut from the condition recommendation page. Product prediction can be initiated with the \texttt{GET RESULTS} button. Prediction models and their training sets are chosen via \texttt{SETTINGS} button. Here, the results are shown for three predicted products with their probabilities, which should be interpreted only qualitatively. Each prediction has shortcuts to run impurity and regioselectivity prediction as additional evaluations.}\label{fig_forward}
\end{figure}


The prediction of reaction conditions, including the identity of agents (catalysts, reagents, solvents) and operating conditions such as temperature and equivalence ratios, is an often overlooked aspect of synthesis planning. Condition recommendation is essential for any suggested reactions to be experimentally validated, and reaction outcomes are strongly influenced by reaction conditions. Reaction type specific data-driven models have been built, for example, to predict classes of solvent and catalyst for Michael additions~\citep{marcou_expert_2015}, ligands for Pd-catalyzed C-N coupling~\citep{li_making_2019}, and reaction contexts including temperature and pressure for four families of substrate-specific cross-coupling reactions~\citep{maser_multilabel_2021,kwon_generative_2022}. \emph{Global} models which are not specific to reaction families have also been developed using a variety of approaches, including multi-class classifier chains~\citep{gao_using_2018} and its variant using a Transformer encoder with SMILES inputs~\citep{wang_generic_2023},  retrieval-augmented prediction using text descriptions of similar reactions~\citep{qian_predictive_2023}, and a two-stage model with candidate generation and ranking as distinct steps~\citep{chen_enhancing_2024}. 


ASKCOS includes a data-driven condition recommendation model based on \citet{gao_using_2018} as well as a second version being developed in ongoing work which additionally predicts equivalence ratios. This model formulates condition recommendation in terms of four sub-problems (1) predicting agent identities as multi-label classification; (2) predicting temperature as binned classification; (3) predicting reactant equivalence ratios as multi-target regression; and (4) predicting agent equivalence ratios as multi-target regression. A screenshot of the condition recommendation page in ASKCOS is shown at the top of Figure \ref{fig_forward}. As with other pages, the reactants and product can be specified either by SMILES strings or by drawing. After generating model predictions by clicking on \texttt{GET RESULTS}, several different condition settings are displayed in a tabular format. The general layout of the reaction condition recommendation page is used across several other modules for consistency: the top panel is for the inputs and the bottom panel is for the prediction results; the model and settings can be modified by clicking on the blue buttons for \texttt{MODEL} and \texttt{SETTINGS}.

\subsection{Reaction outcome prediction}\label{results_forward}


Another component of synthesis planning beyond retrosynthesis is the prediction of reaction outcome(s). In our workflows, these predictions mostly serve to identify chemically infeasible or unfavorable reactions, which the user can choose to prune from the synthesis tree in an interactive setting. Many data-driven approaches simplify this task as predicting the identity of the major product, making it equivalent to a molecule-to-molecule transformation. Similar to one-step retrosynthesis, some studies have formulated it as forward template prediction~\citep{coley_prediction_2017,chen_generalized-template-based_2022}, whereas later developments have been dominated by graph-edit based~\citep{coley_graph-convolutional_2019,sacha_molecule_2021}, electron-flow based~\citep{bradshaw_generative_2019,bi_non-autoregressive_2021}, and translation-based~\citep{schwaller_molecular_2019,tetko_state---art_2020,tu_permutation_2022,zhong_root-aligned_2022} template-free methods. Outcome prediction can answer more fine-grained questions about reaction outcomes, such as the site selectivity of aromatic C-H functionalization reactions~\citep{struble_multitask_2020} or other situations where multiple regioisomers appear possible based on reaction templates~\citep{guan_regio-selectivity_2021}. Another application of outcome prediction is the analysis of potential impurities. When impurities are defined as the minor products of the main reactions or the products of side reactions, they can be predicted by considering several predicted reaction outcomes with lower probabilities or by predicting the outcomes of new reactant sets containing a product of the original reactions (i.e., anticipating potential over-reaction).


The major product prediction page in ASKCOS is shown in the bottom of Figure \ref{fig_forward}. In this specific example, the top outcome is carbamate formation arising from nucleophilic attack of the alcohol into the isocyanate (followed by hydrolysis~\citep{hirama_carbamate_1982}) with a probability of 0.9726. Non-hydrolyzed product and nucleophilic attack by a tetrazole nitrogen are predicted to be the second and third most likely products, but with lower probabilities of 0.0273 and 0.0001, respectively. Past studies have shown these probabilities to correlate with accuracy \emph{on average} but may not be a robust measure of confidence for a particular result~\citep{coley_graph-convolutional_2019,schwaller_molecular_2019,neves_global_2023}. Modules for impurity prediction, regioselectivity, and C-H site-selectivity are accessible via other tabs on the same page. The impurity prediction module relies on the major product predictor and considers minor products, over-reaction, dimerization, solvent adducts, and subsets of reactants; the details of regio- and site-selectivity predictions are reported in \citet{guan_regio-selectivity_2021} and \citet{struble_multitask_2020}, respectively.

\subsection{Pathway scoring and ranking}\label{results_pathway_analysis}

\begin{figure}[h!]
\centering
\includegraphics[width=1.0\textwidth]{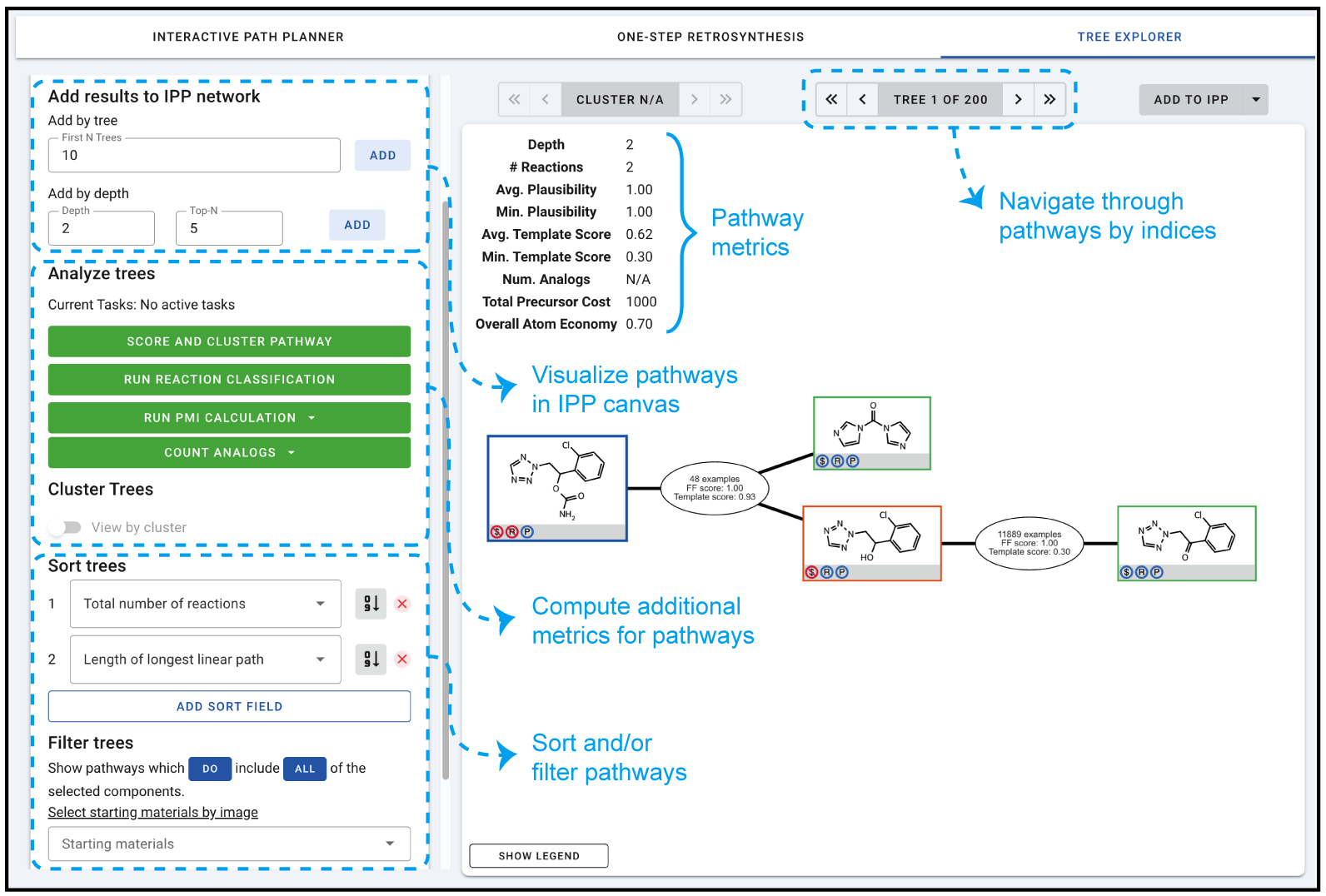}
\caption{Annotated screenshot of the Tree Explorer in ASKCOS. One of 200 routes returned for the target molecule cenobamate is displayed in the main panel (right) of the window. Pathway metrics for the route are shown in the top left corner of the main panel. The left panel provides users with additional actions to perform on the returned synthetic routes. Users can transfer information from the Tree Builder to the Interactive Path Planner to continue exploring retrosynthetic suggestions beyond what was returned (top); users can initiate longer-running pathway-level scoring calculations (middle); finally, users can sort or filter the discovered routes based on various calculated metrics or the presence or absence of specific starting materials and intermediates (bottom). }\label{fig_tree_explorer}
\end{figure}

When many putative synthetic pathways are found for a target molecule, a new challenge arises: to identify the routes that best satisfy a \emph{chemist's }goals, not merely to identify \emph{any} route. It is impractical to triage routes manually when the number of suggestions becomes too large. Strategically similar pathways can first be clustered or grouped based on trained pathway embeddings~\citep{mo_evaluating_2021} or the reactions types they involve (e.g., using NameRxn~\citep{NameRXN} categories or approximations thereof). Pathway-level evaluations then help prioritize promising synthetic routes, though there are many criteria by which a synthetic pathway could be judged~\citep{hoffmann_ranking_2009}. Simple, readily-calculable metrics include step count, longest linear sequence, atom economy, cost of starting materials, or diversifiability based on an estimate of the size of an analog space achieved through building block enumeration~\citep{levin_computer-aided_2023}. More complex metrics that may be derived from other predictive models include estimates of human likeness~\citep{mo_evaluating_2021}, overall perceived likelihood of feasibility, pathway greenness based on solvent usage~\citep{wang_towards_2020}, and process mass intensity (PMI). Assessing purification/isolation requirements in a robust manner remains elusive~\citep{kuznetsov_extractionscore_2021}.

Pathway evaluation in ASKCOS is intended to be performed on pathways returned by the Tree Builder and is accessible from the results page by clicking \texttt{VIEW TREES} as shown in Figure \ref{fig_tree_results}. Since none of the aforementioned evaluation criteria is perfect or sufficient on its own, ASKCOS makes all of them available as part of the \emph{Tree Explorer} (Figure \ref{fig_tree_explorer}), with operations and options on the left panel and the canvas on the right for displaying the pathway with its metrics. In this specific example, 200 pathways have been returned (the default maximum); the first pathway is shown along with automatically-calculated metrics such as the depth, the average plausibility, and the analog count of the pathway. The left panel is organized into three sections that allow the user to visualize multiple best-ranked pathways in the IPP canvas; calculate additional evaluation metrics on-request (rather than automatically due to their computational cost); and sort or filter pathways, e.g., based on certain starting materials of interest.

\subsection{Utilities and supplementary predictive models}\label{results_utilities}


Beyond these ``core'' synthesis planning capabilities, ASKCOS contains additional complementary tools. These include basic drawing functionality (\texttt{Drawing}) and buyable building block search by SMILES or SMARTS (\texttt{Buyable Look-up}). The latter makes use of a predefined commercial catalog that is easily customizable when ASKCOS is deployed. The utilities page also offers two additional machine learning models for solvation prediction (\texttt{Solubility Prediction} and \texttt{Solvent Screening})~\citep{vermeire_predicting_2022} and the prediction of atom- and bond-level descriptors calculated by DFT (\texttt{QM Descriptor})~\citep{guan_regio-selectivity_2021,li_when_2024}. Solubility prediction is part of a long-term goal of improving the relevance of CASP for process chemistry~\citep{griffin_opportunities_2023} as it helps guide the selection of solvents for reactions, liquid-liquid extractions, or crystallizations. ML-estimated QM descriptors can be used as features by other predictive models~\citep{stuyver2022quantum} or standalone, e.g., for human assessment of selectivity. 

The pages for solubility prediction, solvent screening, and QM feature prediction are organized under the \texttt{Utilities} tab, each with a standard layout as in Figure \ref{fig_forward}. Solubility prediction requires a solute,  solvent, and temperature as inputs; solvent screening expects a single solute, a list of solvents, and a list of temperatures; QM prediction needs the SMILES of the target molecule. The screenshots of these pages are shown in Supplementary Figure \ref{fig_solubility_prediction}, \ref{fig_solvent_screening} and \ref{fig_qm_descriptors}, with more detailed explanation on the theoretical aspect and usage in Supplementary Sections \nameref{results_solubility} and \nameref{results_qm}.

\section{Discussion}\label{discussion}



As a broad, extensible software suite for synthesis planning, ASKCOS has been adopted by various organizations within and beyond the context of the MLPDS consortium. While not all usage of ASKCOS is publicly described, several use cases where ASKCOS has aided chemists' workflows have been discussed in the 2020 review by \citet{struble_current_2020}. Particularly well-received features include the interactive planning mode when automatic tree search fails, as well as the possibility for retrosynthetic suggestions to be linked back to literature precedents. ASKCOS has helped serve as a foundation for components of AiZynthFinder~\citep{genheden_aizynthfinder_2020,shields_aizynth_2024} (e.g., in its template extraction strategy). Discovery chemists from Janssen have made use of various modules in ASKCOS at the lead optimization stage. In particular, the one-step retrosynthesis API helped narrow a library of 222k alcohols to 15.7k on the basis of synthesizability~\citep{seierstad_novel_2021}. Pfizer has reported the use of ASKCOS to augment human ideas in their internal graph databases~\citep{avila_chemistry_2024}, and Syngenta has incorporated ASKCOS as one of several idea generation tools when comparing synthetic routes \cite{pasquini2023linchemin}. Beyond the industrial setting, various researchers have used ASKCOS to propose synthetic routes for candidate protease inhibitors~\citep{soukaina_design_2024} and other potential anti-COVID-19 drugs~\citep{qi_optimized_2023}.


As is true of other computational tools, most if not all of the functionalities in ASKCOS aim to \emph{assist} and not replace expert chemists. Ultimately, these models are influenced by the data on which they are trained and may recapitulate popular patterns and trends in data (albeit in useful ways) without understanding the underlying physical sciences. The interpretability of our models fall on a wide spectrum. For the one-step retrosynthesis models, for example, template-based approaches ensure traceability to literature precedents. In contrast, translation-based approaches make predictions in a more black-box manner, which can be creative but to the extent of being alchemical, e.g., by inappropriately adding atoms in the generated SMILES. We refer the reader to individual manuscripts for each prediction module for more in-depth discussions on their strengths and limitations. 

As an illustration of how ASKCOS can be used as an assistive tool, we conduct a synthesis planning exercise, which is described further in Supplementary Section \nameref{results_fda}. We start with automatic retrosynthetic planning using the Tree Builder for all targets. Using typical search settings with template relevance models trained on Reaxys and Pistachio, the default buyable database (consisting of a few hundred thousand molecules from eMolecules~\citep{EMolecules}, Sigma Aldrich~\citep{SigmaAldrich}, LabNetwork~\citep{LabNetwork}, Mcule~\citep{Mcule}, and ChemBridge~\citep{ChemBridge}), and a limit on the number of chemical nodes in the search tree of 5000, hypothetical retrosynthetic pathways are found for many of the targets. Sample shortest routes are presented in Supplementary Figures \ref{fig:fda_study_1}, \ref{fig:fda_study_2}, \ref{fig:fda_study_3}, and \ref{fig:fda_study_4} \emph{exactly as returned} along with the top-1 proposed conditions from the V1 condition recommender. We then demonstrate the use of additional modules in ASKCOS for further analysis when proposed steps are counter-intuitive or appear implausible, e.g., by cross-referencing literature precedents or examining lower-ranked suggested conditions. Thereafter, we show how the flexibility of ASKCOS helps us handle the cases where automatic planning with typical settings fails, by providing a variety of tree search options, and more importantly, a user-friendly interface to directly modify proposed routes. Additional example pathways proposed by ASKCOS from three re-runs with manual edits where appropriate are shown in Supplementary Figures \ref{fig:fda_study_interesting_1}, \ref{fig:fda_study_interesting_2}, and \ref{fig:fda_study_interesting_3}. Examples of targets for which ASKCOS could not automatically find pathways even after these re-runs are shown in Supplementary Figure \ref{fig:fda_study_fail}, which may require interactive planning starting from the targets as described in Sections \nameref{results_ipp_template} and \nameref{results_ipp_multiple}. Synthesis planning tools can produce a range of suggestions, some of which can be high-risk while others are high-confidence. The ideal balance between creativity and conservatism is a matter of personal judgment, and we find that cross-referencing proposals with literature within or outside of ASKCOS is often fruitful.


The utility of a CASP tool depends on not only the modules and tools it contains, but also how it is deployed and customized within an organization. The open source nature of ASKCOS allows for local deployment and, in particular, deployment behind a firewall when working with proprietary data. Deployment is easily customized so that specific modules can be enabled or disabled as needed to save computational resources. Other possible customization includes retrosynthetic model retraining and integration, as well as replacing the default building block database, which are elaborated in Sections \nameref{method_retraining}, and \nameref{method_customization}, respectively. Retraining of translation-based forward predictors with in-house data is similarly possible for the Transformer~\citep{tetko_state---art_2020} and Graph2SMILES~\citep{tu_permutation_2022} models, which has been shown to boost prediction performance for company-specific reactions~\citep{lee_molecular_2019}.


The development of ASKCOS has been guided by a combination of suggestions from collaborators, colleagues, and the community. With this refreshed open source release, we envision a gradual shift to more community-driven development. We laid the groundwork for easy future extension with a major backend refactor in late-2023, in which a microservice-based architecture was formalized and functionalities were re-modularized, as discussed in Supplementary Section \nameref{method_software}. The refactor has made new model addition straightforward, not only for the ASKCOS team but also for advanced users who would like to replace or extend ASKCOS modules with their own. 


We believe that CASP---and computer-assisted chemistry more broadly--is an important part of modern chemistry research that deserves to be precompetitive and accessible to all. At the outset of our work to develop ASKCOS, the landscape of open source solutions was stark; even today, the vast majority of solutions remain commercial even when based on methods described in the open literature. With its ease of use and deployment, customizability, and extensibility, we hope that ASKCOS will find increased adoption and continue to provide a sustainable framework for open source yet production-ready CASP tools.

\section{Methods}\label{methods}

\subsection{Overview}

The majority of the usage of ASKCOS from end-users' perspectives has been covered in Section \nameref{results}. In the rest of Section \nameref{methods}, we will elaborate on the details of modeling and computation, which were only briefly discussed in Section \nameref{results}. Considerations for software development and for the 2023 refactor are elaborated in Supplementary Section \nameref{method_software}. Other non-central but very useful features for more advanced users including model retraining and customization are described in Supplementary Section \nameref{advanced_features}.

The cheminformatics-related computations throughout all ASKCOS modules rely heavily upon RDKit~\citep{RDKit}. Machine learning capabilities are implemented with commonly used packages including but not limited to PyTorch, Tensorflow, Numpy, and Pandas. NetworkX is used for modeling most trees and/or graphs. Clustering is done with scikit-learn, hdbscan, or reaction types from a baseline reaction classification model trained on NameRxn labels~\citep{NameRXN}. Atom mapping is mostly done with RXNMapper~\citep{schwaller_extraction_2021} or Indigo~\citep{Indigo}.

\subsection{Technical details of the template relevance model and MCTS tree search}

Our implementations of the template relevance model and of MCTS tree search deviate from what is described in \citet{segler_planning_2018} with several modifications. Specifically, most of the trained template relevance models we provide use simple feedforward neural networks, which we found to have comparable performance to the original but more complex \emph{highway networks}~\citep{srivastava_training_2015} on larger datasets (e.g., with hundreds of thousands of training reactions). We use RDKit for computing Morgan fingerprints of the input targets, and RDChiral~\citep{coley_rdchiral_2019} for reaction template extraction. The template classification model with feedforward networks is then implemented, trained, and evaluated using PyTorch~\citep{PyTorch}.

Our MCTS tree search differs significantly from \citet{segler_planning_2018} and we use a simplified formulation for Upper Confidence bound applied to Trees (UCT)~\citep{kocsis_bandit_2006}. In particular, we do not use a \emph{rollout} phase. The UCT score for a given reaction node in the search tree is calculated as

\begin{align}
    a_r &= Q_r + c * U_r \\
    &= \frac{s_r * v_r}{n_r} + c * \sqrt{\frac{ln\,N_r}{n_r}}
\end{align}

where the score ($a_r$) takes into consideration an exploitation term ($Q_r$) and an exploration term ($c*U_r$) with $c$ being the weight for exploration. $Q_r$ can be interpreted as a heuristic score with $s_r$ being the \emph{reaction score} from model output (e.g., the template probability for template relevance model), $v_r$ being the average buyability score of all children (1.0 for buyables and 0.0 for non-buyables), and $n_r$ having its typical definition of node visit counts. The $N_r$ in the exploration term is the visit counts of the parent chemical node. The tree search operates in a select-expand-update loop starting from the root node. The search network will keep expanding until the termination criteria is reached, for example, if reaching the time limit. A path enumeration phase identifies all synthesis pathways in the search tree. Optionally, the search can be configured to terminate once the first viable pathway is found.

\subsection{Technical details of other modules}

The implementations of other modules are summarized below.

\begin{itemize}
    \item \textbf{Augmented Transformer~\citep{tetko_state---art_2020} for retrosynthesis and forward prediction}: we re-implement using PyTorch, the OpenNMT~\citep{klein_opennmt_2017} package, and a regex tokenizer based on previous work by Schwaller~\citep{schwaller_found_2018,schwaller_molecular_2019} to tokenize SMILES strings into input and output tokens.
    \item \textbf{Graph2SMILES~\citep{tu_permutation_2022} for outcome prediction and retrosynthesis}: no deviation from the published version.
    \item \textbf{Retrosim~\citep{coley_computer-assisted_2017} for one-step retrosynthesis}: instead of extracting the templates on-the-fly \emph{after} retrieving similar targets in the original implementation, we pre-extract all templates and store them in the database for later use, which speeds up inference at the expense of storage.
    \item \textbf{Retro*~\citep{chen_retro_2020} for multi-step search}: while remaining faithful to the original algorithm, the code structure of our implementation is heavily tailored towards that of the MCTS for consistency.
    \item \textbf{WLDN5~\citep{coley_graph-convolutional_2019} for reaction outcome prediction}: no deviation from the published version.
    \item \textbf{The reaction condition recommender~\citep{gao_using_2018}}: no deviation from the published version for the V1 model. The V2 model is experimental and undergoing active development with publication underway.
    \item \textbf{The analog counting module from \citet{levin_computer-aided_2023}}: no deviation from the published version.
    \item \textbf{The regio-selectivity predictor from \citet{guan_regio-selectivity_2021}}: no deviation from the published version.
    \item \textbf{The site-selectivity predictor from \citet{struble_multitask_2020}}: no deviation from the published version.
    \item \textbf{The synthesis pathway scorer from \citet{mo_evaluating_2021}}: no deviation from the published version.
    \item \textbf{The SCScorer from \citet{coley_scscore_2018}}: no deviation from the published version.
    \item \textbf{The solubility prediction module from \citet{vermeire_predicting_2022}}: no deviation from the published version.
    \item \textbf{The QM descriptor predictor from \citet{li_when_2024}}: no deviation from the published version.
\end{itemize}

\backmatter

\bmhead{Supplementary information}




Supplementary information is available for this manuscript, which provides more details on solubility prediction, solvent screening, and QM descriptor prediction, as well as other advanced features including model retraining and customization. It also includes sections to elaborate on software engineering considerations and to describe in detail the case study on FDA-approved small molecule drugs mentioned in Section \nameref{discussion}.

\bmhead{Acknowledgements}

The authors thank the Machine Learning for Pharmaceutical Discovery and Synthesis consortium for numerous discussions over the years. We thank CAS and NextMove Software for providing large-scale reaction data on which various prediction models have been trained. We thank Itlize Global, LLC and TOC Research for providing software development services in the 2023 refactor. We thank all the past and current contributors to ASKCOS who are too numerous to name. The full contributor list is included in our public instance at \href{https://askcos.mit.edu}{https://askcos.mit.edu} and will be continuously updated.


\section*{Declarations}

\subsection{Funding}

This work was supported by the DARPA Make-It program under Contract ARO W911NF-16-2-0023, the Machine Learning for Pharmaceutical Discovery and Synthesis (MLPDS) consortium, and the National Institutes of Health under grant 1U18TR004149. The continued development of ASKCOS is supported by the MLPDS consortium. Z.T. received additional funding from the MolSSI Fellowship Program and the NSERC PGS-D fellowship under Application No. 577866-2023.

\subsection{Competing interests}

The authors declare no competing interests

\subsection{Ethics approval and consent to participate}
Not applicable.
\subsection{Consent for publication}
Not applicable.
\subsection{Data availability}

All data and trained model weights are shared under MIT licenses, with the only exceptions being models trained on data from the CAS Content Collection~\citep{CASContent} which are only accessible to MLPDS members, as well as the template relevance model trained on the Reaxys dataset ca. 2016 and its associated template set, which are released under the non-commercial CC BY-NC 4.0 license.

We refer the reader to the respective manuscripts for various models for the availability of original training data. In particular, data derived from US patents is generally openly available, including but not limited to USPTO\_50k and USPTO\_full (from the GLN repository~\citep{GLNRepo}), USPTO\_480k (from the WLN repository~\citep{WLNRepo}), as well as USPTO\_STEREO (from the Molecular Transformer repository~\citep{MTRepo}). Proprietary data from the CAS Content Collection, Pistachio, or Reaxys are not able to be shared.

\subsection{Materials availability}
Not applicable.

\subsection{Code availability}

All of the code associated with ASKCOS is fully open sourced under MIT licenses and is available at \href{https://gitlab.com/mlpds\_mit/askcosv2}{https://gitlab.com/mlpds\_mit/askcosv2}, with the main entry point being the \href{https://gitlab.com/mlpds\_mit/askcosv2/askcos2\_core}{askcos2\_core} repository. A snapshot of all repositories including all data and model checkpoints at the time of writing has be archived and is available at \href{https://doi.org/10.5281/zenodo.13929900}{https://doi.org/10.5281/zenodo.13929900}. The ASKCOS wiki is accessible at \href{https://gitlab.com/mlpds\_mit/askcosv2/askcos-docs/-/wikis/home}{https://gitlab.com/mlpds\_mit/askcosv2/askcos-docs/-/wikis/home}.

\subsection{Author contributions}

Z.T., C.W.C., M.L., M.E.F., T.J.S., and M.M. conceived the idea of the 2023 ASKCOS refactor, on top of historical development efforts led by C.W.C., M.E.F., and M.L. Z.T. led the design and implementation of microservice-based ASKCOS. Z.T., S.J.C., M.H.F., and H.L developed various codes for the refactor. Z.T. and C.W.C. led the manuscript writing. Every author contributed to and approved the manuscript. In particular, J.R. and K.Y. made the figures and polished the introduction. The main contributors for the result sections include J.R. and I.L. (for retrosynthesis and pathway evaluation), X.S. (for reaction condition recommendation), and J.F.J. (for reaction outcome prediction). N.M. and S.-C.L. led the discussion of the solubility modules and of the QM descriptor module, respectively. S.J.C., M.H.F., H.L., and M.M. contributed to various sections for software engineering details in the SI. J.R., Z.T., and J.P.L. conducted the synthesis planning exercise and discussed the results in the SI. W.H.G, K.F.J., and C.W.C. have provided oversight, organization, and fundraising to support the development of ASKCOS over the years.

\bibliography{sn-bibliography}


\begin{thebibliography}{125}
\ifx \bisbn   \undefined \def \bisbn  #1{ISBN #1}\fi
\ifx \binits  \undefined \def \binits#1{#1}\fi
\ifx \bauthor  \undefined \def \bauthor#1{#1}\fi
\ifx \batitle  \undefined \def \batitle#1{#1}\fi
\ifx \bjtitle  \undefined \def \bjtitle#1{#1}\fi
\ifx \bvolume  \undefined \def \bvolume#1{\textbf{#1}}\fi
\ifx \byear  \undefined \def \byear#1{#1}\fi
\ifx \bissue  \undefined \def \bissue#1{#1}\fi
\ifx \bfpage  \undefined \def \bfpage#1{#1}\fi
\ifx \blpage  \undefined \def \blpage #1{#1}\fi
\ifx \burl  \undefined \def \burl#1{\textsf{#1}}\fi
\ifx \doiurl  \undefined \def \doiurl#1{\url{https://doi.org/#1}}\fi
\ifx \betal  \undefined \def \betal{\textit{et al.}}\fi
\ifx \binstitute  \undefined \def \binstitute#1{#1}\fi
\ifx \binstitutionaled  \undefined \def \binstitutionaled#1{#1}\fi
\ifx \bctitle  \undefined \def \bctitle#1{#1}\fi
\ifx \beditor  \undefined \def \beditor#1{#1}\fi
\ifx \bpublisher  \undefined \def \bpublisher#1{#1}\fi
\ifx \bbtitle  \undefined \def \bbtitle#1{#1}\fi
\ifx \bedition  \undefined \def \bedition#1{#1}\fi
\ifx \bseriesno  \undefined \def \bseriesno#1{#1}\fi
\ifx \blocation  \undefined \def \blocation#1{#1}\fi
\ifx \bsertitle  \undefined \def \bsertitle#1{#1}\fi
\ifx \bsnm \undefined \def \bsnm#1{#1}\fi
\ifx \bsuffix \undefined \def \bsuffix#1{#1}\fi
\ifx \bparticle \undefined \def \bparticle#1{#1}\fi
\ifx \barticle \undefined \def \barticle#1{#1}\fi
\bibcommenthead
\ifx \bconfdate \undefined \def \bconfdate #1{#1}\fi
\ifx \botherref \undefined \def \botherref #1{#1}\fi
\ifx \url \undefined \def \url#1{\textsf{#1}}\fi
\ifx \bchapter \undefined \def \bchapter#1{#1}\fi
\ifx \bbook \undefined \def \bbook#1{#1}\fi
\ifx \bcomment \undefined \def \bcomment#1{#1}\fi
\ifx \oauthor \undefined \def \oauthor#1{#1}\fi
\ifx \citeauthoryear \undefined \def \citeauthoryear#1{#1}\fi
\ifx \endbibitem  \undefined \def \endbibitem {}\fi
\ifx \bconflocation  \undefined \def \bconflocation#1{#1}\fi
\ifx \arxivurl  \undefined \def \arxivurl#1{\textsf{#1}}\fi
\csname PreBibitemsHook\endcsname

\bibitem[\protect\citeauthoryear{Corey and Wipke}{1969}]{corey_computer-assisted_1969}
\begin{barticle}
\bauthor{\bsnm{Corey}, \binits{E.J.}},
\bauthor{\bsnm{Wipke}, \binits{W.T.}}:
\batitle{Computer-{Assisted} {Design} of {Complex} {Organic} {Syntheses}}.
\bjtitle{Science}
\bvolume{166}(\bissue{3902}),
\bfpage{178}--\blpage{192}
(\byear{1969})
\doiurl{10.1126/science.166.3902.178} .
\bcomment{Publisher: American Association for the Advancement of Science}
\end{barticle}
\endbibitem

\bibitem[\protect\citeauthoryear{Warr}{2014}]{warr_short_2014}
\begin{barticle}
\bauthor{\bsnm{Warr}, \binits{W.A.}}:
\batitle{A {Short} {Review} of {Chemical} {Reaction} {Database} {Systems}, {Computer}-{Aided} {Synthesis} {Design}, {Reaction} {Prediction} and {Synthetic} {Feasibility}}.
\bjtitle{Molecular Informatics}
\bvolume{33}(\bissue{6-7}),
\bfpage{469}--\blpage{476}
(\byear{2014})
\doiurl{10.1002/minf.201400052}
\end{barticle}
\endbibitem

\bibitem[\protect\citeauthoryear{Corey}{1988}]{corey_robert_1988}
\begin{barticle}
\bauthor{\bsnm{Corey}, \binits{E.J.}}:
\batitle{Robert {Robinson} {Lecture}. {Retrosynthetic} thinking—essentials and examples}.
\bjtitle{Chemical Society Reviews}
\bvolume{17}(\bissue{0}),
\bfpage{111}--\blpage{133}
(\byear{1988})
\doiurl{10.1039/CS9881700111} .
\bcomment{Publisher: The Royal Society of Chemistry}
\end{barticle}
\endbibitem

\bibitem[\protect\citeauthoryear{Corey et~al.}{1972}]{corey_computer-assisted_1972}
\begin{barticle}
\bauthor{\bsnm{Corey}, \binits{E.J.}},
\bauthor{\bsnm{Cramer}, \binits{R.D.I.}},
\bauthor{\bsnm{Howe}, \binits{W.J.}}:
\batitle{Computer-assisted synthetic analysis for complex molecules. {Methods} and procedures for machine generation of synthetic intermediates}.
\bjtitle{Journal of the American Chemical Society}
\bvolume{94}(\bissue{2}),
\bfpage{440}--\blpage{459}
(\byear{1972})
\doiurl{10.1021/ja00757a022} .
\bcomment{Publisher: American Chemical Society}
\end{barticle}
\endbibitem

\bibitem[\protect\citeauthoryear{Wipke et~al.}{1978}]{wipke_simulation_1978}
\begin{barticle}
\bauthor{\bsnm{Wipke}, \binits{W.T.}},
\bauthor{\bsnm{Ouchi}, \binits{G.I.}},
\bauthor{\bsnm{Krishnan}, \binits{S.}}:
\batitle{Simulation and evaluation of chemical synthesis—{SECS}: {An} application of artificial intelligence techniques}.
\bjtitle{Artificial Intelligence}
\bvolume{11}(\bissue{1}),
\bfpage{173}--\blpage{193}
(\byear{1978})
\doiurl{10.1016/0004-3702(78)90016-4}
\end{barticle}
\endbibitem

\bibitem[\protect\citeauthoryear{Funatsu and Sasaki}{1988}]{funatsu_computer-assisted_1988}
\begin{barticle}
\bauthor{\bsnm{Funatsu}, \binits{K.}},
\bauthor{\bsnm{Sasaki}, \binits{S.-I.}}:
\batitle{Computer-assisted organic synthesis design and reaction prediction system, “{AIPHOS}”}.
\bjtitle{Tetrahedron Computer Methodology}
\bvolume{1}(\bissue{1}),
\bfpage{27}--\blpage{37}
(\byear{1988})
\doiurl{10.1016/0898-5529(88)90006-1}
\end{barticle}
\endbibitem

\bibitem[\protect\citeauthoryear{Gasteiger and Ihlenfeldt}{1990}]{gasteiger_wodca_1990}
\begin{bchapter}
\bauthor{\bsnm{Gasteiger}, \binits{J.}},
\bauthor{\bsnm{Ihlenfeldt}, \binits{W.D.}}:
\bctitle{The {WODCA} {System}}.
In: \beditor{\bsnm{Gasteiger}, \binits{J.}} (ed.)
\bbtitle{Software {Development} in {Chemistry} 4},
pp. \bfpage{57}--\blpage{65}.
\bpublisher{Springer},
\blocation{Berlin, Heidelberg}
(\byear{1990}).
\doiurl{10.1007/978-3-642-75430-2_7}
\end{bchapter}
\endbibitem

\bibitem[\protect\citeauthoryear{Grzybowski et~al.}{2018}]{grzybowski_chematica_2018}
\begin{barticle}
\bauthor{\bsnm{Grzybowski}, \binits{B.A.}},
\bauthor{\bsnm{Szymkuć}, \binits{S.}},
\bauthor{\bsnm{Gajewska}, \binits{E.P.}},
\bauthor{\bsnm{Molga}, \binits{K.}},
\bauthor{\bsnm{Dittwald}, \binits{P.}},
\bauthor{\bsnm{Wołos}, \binits{A.}},
\bauthor{\bsnm{Klucznik}, \binits{T.}}:
\batitle{Chematica: {A} {Story} of {Computer} {Code} {That} {Started} to {Think} like a {Chemist}}.
\bjtitle{Chem}
\bvolume{4}(\bissue{3}),
\bfpage{390}--\blpage{398}
(\byear{2018})
\doiurl{10.1016/j.chempr.2018.02.024} .
\bcomment{Publisher: Elsevier}
\end{barticle}
\endbibitem

\bibitem[\protect\citeauthoryear{Klucznik et~al.}{2018}]{klucznik_efficient_2018}
\begin{barticle}
\bauthor{\bsnm{Klucznik}, \binits{T.}},
\bauthor{\bsnm{Mikulak-Klucznik}, \binits{B.}},
\bauthor{\bsnm{McCormack}, \binits{M.P.}},
\bauthor{\bsnm{Lima}, \binits{H.}},
\bauthor{\bsnm{Szymkuć}, \binits{S.}},
\bauthor{\bsnm{Bhowmick}, \binits{M.}},
\bauthor{\bsnm{Molga}, \binits{K.}},
\bauthor{\bsnm{Zhou}, \binits{Y.}},
\bauthor{\bsnm{Rickershauser}, \binits{L.}},
\bauthor{\bsnm{Gajewska}, \binits{E.P.}},
\bauthor{\bsnm{Toutchkine}, \binits{A.}},
\bauthor{\bsnm{Dittwald}, \binits{P.}},
\bauthor{\bsnm{Startek}, \binits{M.P.}},
\bauthor{\bsnm{Kirkovits}, \binits{G.J.}},
\bauthor{\bsnm{Roszak}, \binits{R.}},
\bauthor{\bsnm{Adamski}, \binits{A.}},
\bauthor{\bsnm{Sieredzińska}, \binits{B.}},
\bauthor{\bsnm{Mrksich}, \binits{M.}},
\bauthor{\bsnm{Trice}, \binits{S.L.J.}},
\bauthor{\bsnm{Grzybowski}, \binits{B.A.}}:
\batitle{Efficient {Syntheses} of {Diverse}, {Medicinally} {Relevant} {Targets} {Planned} by {Computer} and {Executed} in the {Laboratory}}.
\bjtitle{Chem}
\bvolume{4}(\bissue{3}),
\bfpage{522}--\blpage{532}
(\byear{2018})
\doiurl{10.1016/j.chempr.2018.02.002} .
\bcomment{Publisher: Elsevier}
\end{barticle}
\endbibitem

\bibitem[\protect\citeauthoryear{Mikulak-Klucznik et~al.}{2020}]{mikulak-klucznik_computational_2020}
\begin{barticle}
\bauthor{\bsnm{Mikulak-Klucznik}, \binits{B.}},
\bauthor{\bsnm{Gołebiowska}, \binits{P.}},
\bauthor{\bsnm{Bayly}, \binits{A.A.}},
\bauthor{\bsnm{Popik}, \binits{O.}},
\bauthor{\bsnm{Klucznik}, \binits{T.}},
\bauthor{\bsnm{Szymkuć}, \binits{S.}},
\bauthor{\bsnm{Gajewska}, \binits{E.P.}},
\bauthor{\bsnm{Dittwald}, \binits{P.}},
\bauthor{\bsnm{Staszewska-Krajewska}, \binits{O.}},
\bauthor{\bsnm{Beker}, \binits{W.}},
\bauthor{\bsnm{Badowski}, \binits{T.}},
\bauthor{\bsnm{Scheidt}, \binits{K.A.}},
\bauthor{\bsnm{Molga}, \binits{K.}},
\bauthor{\bsnm{Mlynarski}, \binits{J.}},
\bauthor{\bsnm{Mrksich}, \binits{M.}},
\bauthor{\bsnm{Grzybowski}, \binits{B.A.}}:
\batitle{Computational planning of the synthesis of complex natural products}.
\bjtitle{Nature}
\bvolume{588}(\bissue{7836}),
\bfpage{83}--\blpage{88}
(\byear{2020})
\doiurl{10.1038/s41586-020-2855-y} .
\bcomment{Number: 7836 Publisher: Nature Publishing Group}
\end{barticle}
\endbibitem

\bibitem[\protect\citeauthoryear{Tu et~al.}{2023}]{tu_predictive_2023}
\begin{barticle}
\bauthor{\bsnm{Tu}, \binits{Z.}},
\bauthor{\bsnm{Stuyver}, \binits{T.}},
\bauthor{\bsnm{Coley}, \binits{C.W.}}:
\batitle{Predictive chemistry: machine learning for reaction deployment, reaction development, and reaction discovery}.
\bjtitle{Chemical Science}
\bvolume{14}(\bissue{2}),
\bfpage{226}--\blpage{244}
(\byear{2023})
\doiurl{10.1039/D2SC05089G} .
\bcomment{Publisher: The Royal Society of Chemistry}
\end{barticle}
\endbibitem

\bibitem[\protect\citeauthoryear{Schwaller et~al.}{2022}]{schwaller_machine_2022}
\begin{barticle}
\bauthor{\bsnm{Schwaller}, \binits{P.}},
\bauthor{\bsnm{Vaucher}, \binits{A.C.}},
\bauthor{\bsnm{Laplaza}, \binits{R.}},
\bauthor{\bsnm{Bunne}, \binits{C.}},
\bauthor{\bsnm{Krause}, \binits{A.}},
\bauthor{\bsnm{Corminboeuf}, \binits{C.}},
\bauthor{\bsnm{Laino}, \binits{T.}}:
\batitle{Machine intelligence for chemical reaction space}.
\bjtitle{WIREs Computational Molecular Science}
\bvolume{12}(\bissue{5}),
\bfpage{1604}
(\byear{2022})
\doiurl{10.1002/wcms.1604}
\end{barticle}
\endbibitem

\bibitem[\protect\citeauthoryear{Dai et~al.}{2019}]{dai_retrosynthesis_2019}
\begin{bchapter}
\bauthor{\bsnm{Dai}, \binits{H.}},
\bauthor{\bsnm{Li}, \binits{C.}},
\bauthor{\bsnm{Coley}, \binits{C.}},
\bauthor{\bsnm{Dai}, \binits{B.}},
\bauthor{\bsnm{Song}, \binits{L.}}:
\bctitle{Retrosynthesis {Prediction} with {Conditional} {Graph} {Logic} {Network}}.
In: \bbtitle{Advances in {Neural} {Information} {Processing} {Systems}},
vol. \bseriesno{32}.
\bpublisher{Curran Associates, Inc.},
\blocation{Vancouver, British Columbia, Canada}
(\byear{2019})
\end{bchapter}
\endbibitem

\bibitem[\protect\citeauthoryear{Chen and Jung}{2021}]{chen_deep_2021}
\begin{barticle}
\bauthor{\bsnm{Chen}, \binits{S.}},
\bauthor{\bsnm{Jung}, \binits{Y.}}:
\batitle{Deep {Retrosynthetic} {Reaction} {Prediction} using {Local} {Reactivity} and {Global} {Attention}}.
\bjtitle{JACS Au}
\bvolume{1}(\bissue{10}),
\bfpage{1612}--\blpage{1620}
(\byear{2021})
\doiurl{10.1021/jacsau.1c00246} .
\bcomment{Publisher: American Chemical Society}
\end{barticle}
\endbibitem

\bibitem[\protect\citeauthoryear{Coley et~al.}{2017}]{coley_computer-assisted_2017}
\begin{barticle}
\bauthor{\bsnm{Coley}, \binits{C.W.}},
\bauthor{\bsnm{Rogers}, \binits{L.}},
\bauthor{\bsnm{Green}, \binits{W.H.}},
\bauthor{\bsnm{Jensen}, \binits{K.F.}}:
\batitle{Computer-{Assisted} {Retrosynthesis} {Based} on {Molecular} {Similarity}}.
\bjtitle{ACS Central Science}
\bvolume{3}(\bissue{12}),
\bfpage{1237}--\blpage{1245}
(\byear{2017})
\doiurl{10.1021/acscentsci.7b00355} .
\bcomment{Publisher: American Chemical Society}
\end{barticle}
\endbibitem

\bibitem[\protect\citeauthoryear{Xie et~al.}{2023}]{xie_retrosynthesis_2023}
\begin{bchapter}
\bauthor{\bsnm{Xie}, \binits{S.}},
\bauthor{\bsnm{Yan}, \binits{R.}},
\bauthor{\bsnm{Guo}, \binits{J.}},
\bauthor{\bsnm{Xia}, \binits{Y.}},
\bauthor{\bsnm{Wu}, \binits{L.}},
\bauthor{\bsnm{Qin}, \binits{T.}}:
\bctitle{Retrosynthesis prediction with local template retrieval}.
In: \bbtitle{Proceedings of the {Thirty}-{Seventh} {AAAI} {Conference} on {Artificial} {Intelligence} and {Thirty}-{Fifth} {Conference} on {Innovative} {Applications} of {Artificial} {Intelligence} and {Thirteenth} {Symposium} on {Educational} {Advances} in {Artificial} {Intelligence}}.
\bsertitle{{AAAI}'23/{IAAI}'23/{EAAI}'23},
vol. \bseriesno{37},
pp. \bfpage{5330}--\blpage{5338}.
\bpublisher{AAAI Press},
\blocation{Washington DC, US}
(\byear{2023}).
\doiurl{10.1609/aaai.v37i4.25664}
\end{bchapter}
\endbibitem

\bibitem[\protect\citeauthoryear{Tu and Coley}{2022}]{tu_permutation_2022}
\begin{barticle}
\bauthor{\bsnm{Tu}, \binits{Z.}},
\bauthor{\bsnm{Coley}, \binits{C.W.}}:
\batitle{Permutation {Invariant} {Graph}-to-{Sequence} {Model} for {Template}-{Free} {Retrosynthesis} and {Reaction} {Prediction}}.
\bjtitle{Journal of Chemical Information and Modeling}
\bvolume{62}(\bissue{15}),
\bfpage{3503}--\blpage{3513}
(\byear{2022})
\doiurl{10.1021/acs.jcim.2c00321} .
\bcomment{Publisher: American Chemical Society}
\end{barticle}
\endbibitem

\bibitem[\protect\citeauthoryear{Tetko et~al.}{2020}]{tetko_state---art_2020}
\begin{barticle}
\bauthor{\bsnm{Tetko}, \binits{I.V.}},
\bauthor{\bsnm{Karpov}, \binits{P.}},
\bauthor{\bsnm{Van~Deursen}, \binits{R.}},
\bauthor{\bsnm{Godin}, \binits{G.}}:
\batitle{State-of-the-art augmented {NLP} transformer models for direct and single-step retrosynthesis}.
\bjtitle{Nature Communications}
\bvolume{11}(\bissue{1}),
\bfpage{5575}
(\byear{2020})
\doiurl{10.1038/s41467-020-19266-y} .
\bcomment{Number: 1 Publisher: Nature Publishing Group}
\end{barticle}
\endbibitem

\bibitem[\protect\citeauthoryear{Zhong et~al.}{2022}]{zhong_root-aligned_2022}
\begin{barticle}
\bauthor{\bsnm{Zhong}, \binits{Z.}},
\bauthor{\bsnm{Song}, \binits{J.}},
\bauthor{\bsnm{Feng}, \binits{Z.}},
\bauthor{\bsnm{Liu}, \binits{T.}},
\bauthor{\bsnm{Jia}, \binits{L.}},
\bauthor{\bsnm{Yao}, \binits{S.}},
\bauthor{\bsnm{Wu}, \binits{M.}},
\bauthor{\bsnm{Hou}, \binits{T.}},
\bauthor{\bsnm{Song}, \binits{M.}}:
\batitle{Root-aligned {SMILES}: a tight representation for chemical reaction prediction}.
\bjtitle{Chemical Science}
\bvolume{13}(\bissue{31}),
\bfpage{9023}--\blpage{9034}
(\byear{2022})
\doiurl{10.1039/D2SC02763A} .
\bcomment{Publisher: Royal Society of Chemistry}
\end{barticle}
\endbibitem

\bibitem[\protect\citeauthoryear{Sacha et~al.}{2021}]{sacha_molecule_2021}
\begin{barticle}
\bauthor{\bsnm{Sacha}, \binits{M.}},
\bauthor{\bsnm{Błaż}, \binits{M.}},
\bauthor{\bsnm{Byrski}, \binits{P.}},
\bauthor{\bsnm{Dabrowski-Tumański}, \binits{P.}},
\bauthor{\bsnm{Chromiński}, \binits{M.}},
\bauthor{\bsnm{Loska}, \binits{R.}},
\bauthor{\bsnm{Włodarczyk-Pruszyński}, \binits{P.}},
\bauthor{\bsnm{Jastrzebski}, \binits{S.}}:
\batitle{Molecule {Edit} {Graph} {Attention} {Network}: {Modeling} {Chemical} {Reactions} as {Sequences} of {Graph} {Edits}}.
\bjtitle{Journal of Chemical Information and Modeling}
\bvolume{61}(\bissue{7}),
\bfpage{3273}--\blpage{3284}
(\byear{2021})
\doiurl{10.1021/acs.jcim.1c00537} .
\bcomment{Publisher: American Chemical Society}
\end{barticle}
\endbibitem

\bibitem[\protect\citeauthoryear{Somnath et~al.}{2021}]{somnath_learning_2021}
\begin{bchapter}
\bauthor{\bsnm{Somnath}, \binits{V.R.}},
\bauthor{\bsnm{Bunne}, \binits{C.}},
\bauthor{\bsnm{Coley}, \binits{C.}},
\bauthor{\bsnm{Krause}, \binits{A.}},
\bauthor{\bsnm{Barzilay}, \binits{R.}}:
\bctitle{Learning {Graph} {Models} for {Retrosynthesis} {Prediction}}.
In: \bbtitle{Advances in {Neural} {Information} {Processing} {Systems}},
vol. \bseriesno{34},
pp. \bfpage{9405}--\blpage{9415}.
\bpublisher{Curran Associates, Inc.},
\blocation{virtual}
(\byear{2021})
\end{bchapter}
\endbibitem

\bibitem[\protect\citeauthoryear{Igashov et~al.}{2023}]{igashov_retrobridge_2023}
\begin{bchapter}
\bauthor{\bsnm{Igashov}, \binits{I.}},
\bauthor{\bsnm{Schneuing}, \binits{A.}},
\bauthor{\bsnm{Segler}, \binits{M.}},
\bauthor{\bsnm{Bronstein}, \binits{M.M.}},
\bauthor{\bsnm{Correia}, \binits{B.}}:
\bctitle{{RetroBridge}: {Modeling} {Retrosynthesis} with {Markov} {Bridges}}.
In: \bbtitle{The Twelfth International Conference on Learning Representations}
(\byear{2023})
\end{bchapter}
\endbibitem

\bibitem[\protect\citeauthoryear{Gaiński et~al.}{2024}]{gainski_retrogfn_2024}
\begin{bchapter}
\bauthor{\bsnm{Gaiński}, \binits{P.}},
\bauthor{\bsnm{Koziarski}, \binits{M.}},
\bauthor{\bsnm{Maziarz}, \binits{K.}},
\bauthor{\bsnm{Segler}, \binits{M.}},
\bauthor{\bsnm{Tabor}, \binits{J.}},
\bauthor{\bsnm{Śmieja}, \binits{M.}}:
\bctitle{{RetroGFN}: {Diverse} and {Feasible} {Retrosynthesis} using {GFlowNets}}.
In: \bbtitle{ICLR 2024 Workshop on Generative and Experimental Perspectives for Biomolecular Design}
(\byear{2024})
\end{bchapter}
\endbibitem

\bibitem[\protect\citeauthoryear{Heifets and Jurisica}{2012}]{heifets_construction_2012}
\begin{bchapter}
\bauthor{\bsnm{Heifets}, \binits{A.}},
\bauthor{\bsnm{Jurisica}, \binits{I.}}:
\bctitle{Construction of new medicines via game proof search}.
In: \bbtitle{Proceedings of the {Twenty}-{Sixth} {AAAI} {Conference} on {Artificial} {Intelligence}}.
\bsertitle{{AAAI}'12},
pp. \bfpage{1564}--\blpage{1570}.
\bpublisher{AAAI Press},
\blocation{Toronto, Ontario, Canada}
(\byear{2012})
\end{bchapter}
\endbibitem

\bibitem[\protect\citeauthoryear{Segler et~al.}{2018}]{segler_planning_2018}
\begin{barticle}
\bauthor{\bsnm{Segler}, \binits{M.H.S.}},
\bauthor{\bsnm{Preuss}, \binits{M.}},
\bauthor{\bsnm{Waller}, \binits{M.P.}}:
\batitle{Planning chemical syntheses with deep neural networks and symbolic {AI}}.
\bjtitle{Nature}
\bvolume{555}(\bissue{7698}),
\bfpage{604}--\blpage{610}
(\byear{2018})
\doiurl{10.1038/nature25978} .
\bcomment{Number: 7698 Publisher: Nature Publishing Group}
\end{barticle}
\endbibitem

\bibitem[\protect\citeauthoryear{Kishimoto et~al.}{2019}]{kishimoto_depth-first_2019}
\begin{bchapter}
\bauthor{\bsnm{Kishimoto}, \binits{A.}},
\bauthor{\bsnm{Buesser}, \binits{B.}},
\bauthor{\bsnm{Chen}, \binits{B.}},
\bauthor{\bsnm{Botea}, \binits{A.}}:
\bctitle{Depth-{First} {Proof}-{Number} {Search} with {Heuristic} {Edge} {Cost} and {Application} to {Chemical} {Synthesis} {Planning}}.
In: \bbtitle{Advances in {Neural} {Information} {Processing} {Systems}},
vol. \bseriesno{32}.
\bpublisher{Curran Associates, Inc.},
\blocation{Vancouver, British Columbia, Canada}
(\byear{2019})
\end{bchapter}
\endbibitem

\bibitem[\protect\citeauthoryear{Schwaller et~al.}{2020}]{schwaller_predicting_2020}
\begin{barticle}
\bauthor{\bsnm{Schwaller}, \binits{P.}},
\bauthor{\bsnm{Petraglia}, \binits{R.}},
\bauthor{\bsnm{Zullo}, \binits{V.}},
\bauthor{\bsnm{Nair}, \binits{V.H.}},
\bauthor{\bsnm{Haeuselmann}, \binits{R.A.}},
\bauthor{\bsnm{Pisoni}, \binits{R.}},
\bauthor{\bsnm{Bekas}, \binits{C.}},
\bauthor{\bsnm{Iuliano}, \binits{A.}},
\bauthor{\bsnm{Laino}, \binits{T.}}:
\batitle{Predicting retrosynthetic pathways using transformer-based models and a hyper-graph exploration strategy}.
\bjtitle{Chemical Science}
\bvolume{11}(\bissue{12}),
\bfpage{3316}--\blpage{3325}
(\byear{2020})
\doiurl{10.1039/C9SC05704H} .
\bcomment{Publisher: The Royal Society of Chemistry}
\end{barticle}
\endbibitem

\bibitem[\protect\citeauthoryear{Chen et~al.}{2020}]{chen_retro_2020}
\begin{bchapter}
\bauthor{\bsnm{Chen}, \binits{B.}},
\bauthor{\bsnm{Li}, \binits{C.}},
\bauthor{\bsnm{Dai}, \binits{H.}},
\bauthor{\bsnm{Song}, \binits{L.}}:
\bctitle{Retro*: {Learning} {Retrosynthetic} {Planning} with {Neural} {Guided} {A}* {Search}}.
In: \bbtitle{Proceedings of the 37th {International} {Conference} on {Machine} {Learning}},
pp. \bfpage{1608}--\blpage{1616}.
\bpublisher{PMLR},
\blocation{virtual}
(\byear{2020}).
\bcomment{ISSN: 2640-3498}
\end{bchapter}
\endbibitem

\bibitem[\protect\citeauthoryear{Gao et~al.}{2018}]{gao_using_2018}
\begin{barticle}
\bauthor{\bsnm{Gao}, \binits{H.}},
\bauthor{\bsnm{Struble}, \binits{T.J.}},
\bauthor{\bsnm{Coley}, \binits{C.W.}},
\bauthor{\bsnm{Wang}, \binits{Y.}},
\bauthor{\bsnm{Green}, \binits{W.H.}},
\bauthor{\bsnm{Jensen}, \binits{K.F.}}:
\batitle{Using {Machine} {Learning} {To} {Predict} {Suitable} {Conditions} for {Organic} {Reactions}}.
\bjtitle{ACS Central Science}
\bvolume{4}(\bissue{11}),
\bfpage{1465}--\blpage{1476}
(\byear{2018})
\doiurl{10.1021/acscentsci.8b00357} .
\bcomment{Publisher: American Chemical Society}
\end{barticle}
\endbibitem

\bibitem[\protect\citeauthoryear{Maser et~al.}{2021}]{maser_multilabel_2021}
\begin{barticle}
\bauthor{\bsnm{Maser}, \binits{M.R.}},
\bauthor{\bsnm{Cui}, \binits{A.Y.}},
\bauthor{\bsnm{Ryou}, \binits{S.}},
\bauthor{\bsnm{DeLano}, \binits{T.J.}},
\bauthor{\bsnm{Yue}, \binits{Y.}},
\bauthor{\bsnm{Reisman}, \binits{S.E.}}:
\batitle{Multilabel {Classification} {Models} for the {Prediction} of {Cross}-{Coupling} {Reaction} {Conditions}}.
\bjtitle{Journal of Chemical Information and Modeling}
\bvolume{61}(\bissue{1}),
\bfpage{156}--\blpage{166}
(\byear{2021})
\doiurl{10.1021/acs.jcim.0c01234} .
\bcomment{Publisher: American Chemical Society}
\end{barticle}
\endbibitem

\bibitem[\protect\citeauthoryear{Kwon et~al.}{2022}]{kwon_generative_2022}
\begin{barticle}
\bauthor{\bsnm{Kwon}, \binits{Y.}},
\bauthor{\bsnm{Kim}, \binits{S.}},
\bauthor{\bsnm{Choi}, \binits{Y.-S.}},
\bauthor{\bsnm{Kang}, \binits{S.}}:
\batitle{Generative {Modeling} to {Predict} {Multiple} {Suitable} {Conditions} for {Chemical} {Reactions}}.
\bjtitle{Journal of Chemical Information and Modeling}
(\byear{2022})
\doiurl{10.1021/acs.jcim.2c01085} .
\bcomment{Publisher: American Chemical Society}
\end{barticle}
\endbibitem

\bibitem[\protect\citeauthoryear{Qian et~al.}{2023}]{qian_predictive_2023}
\begin{bchapter}
\bauthor{\bsnm{Qian}, \binits{Y.}},
\bauthor{\bsnm{Li}, \binits{Z.}},
\bauthor{\bsnm{Tu}, \binits{Z.}},
\bauthor{\bsnm{Coley}, \binits{C.}},
\bauthor{\bsnm{Barzilay}, \binits{R.}}:
\bctitle{Predictive {Chemistry} {Augmented} with {Text} {Retrieval}}.
In: \beditor{\bsnm{Bouamor}, \binits{H.}},
\beditor{\bsnm{Pino}, \binits{J.}},
\beditor{\bsnm{Bali}, \binits{K.}} (eds.)
\bbtitle{Proceedings of the 2023 {Conference} on {Empirical} {Methods} in {Natural} {Language} {Processing}},
pp. \bfpage{12731}--\blpage{12745}.
\bpublisher{Association for Computational Linguistics},
\blocation{Singapore}
(\byear{2023}).
\doiurl{10.18653/v1/2023.emnlp-main.784}
\end{bchapter}
\endbibitem

\bibitem[\protect\citeauthoryear{Chen and Li}{2024}]{chen_enhancing_2024}
\begin{barticle}
\bauthor{\bsnm{Chen}, \binits{L.-Y.}},
\bauthor{\bsnm{Li}, \binits{Y.-P.}}:
\batitle{Enhancing chemical synthesis: a two-stage deep neural network for predicting feasible reaction conditions}.
\bjtitle{Journal of Cheminformatics}
\bvolume{16}(\bissue{1}),
\bfpage{11}
(\byear{2024})
\doiurl{10.1186/s13321-024-00805-4}
\end{barticle}
\endbibitem

\bibitem[\protect\citeauthoryear{Schwaller et~al.}{2019}]{schwaller_molecular_2019}
\begin{barticle}
\bauthor{\bsnm{Schwaller}, \binits{P.}},
\bauthor{\bsnm{Laino}, \binits{T.}},
\bauthor{\bsnm{Gaudin}, \binits{T.}},
\bauthor{\bsnm{Bolgar}, \binits{P.}},
\bauthor{\bsnm{Hunter}, \binits{C.A.}},
\bauthor{\bsnm{Bekas}, \binits{C.}},
\bauthor{\bsnm{Lee}, \binits{A.A.}}:
\batitle{Molecular {Transformer}: {A} {Model} for {Uncertainty}-{Calibrated} {Chemical} {Reaction} {Prediction}}.
\bjtitle{ACS Central Science}
\bvolume{5}(\bissue{9}),
\bfpage{1572}--\blpage{1583}
(\byear{2019})
\doiurl{10.1021/acscentsci.9b00576} .
\bcomment{Publisher: American Chemical Society}
\end{barticle}
\endbibitem

\bibitem[\protect\citeauthoryear{Coley et~al.}{2019}]{coley_graph-convolutional_2019}
\begin{barticle}
\bauthor{\bsnm{Coley}, \binits{C.W.}},
\bauthor{\bsnm{Jin}, \binits{W.}},
\bauthor{\bsnm{Rogers}, \binits{L.}},
\bauthor{\bsnm{Jamison}, \binits{T.F.}},
\bauthor{\bsnm{Jaakkola}, \binits{T.S.}},
\bauthor{\bsnm{Green}, \binits{W.H.}},
\bauthor{\bsnm{Barzilay}, \binits{R.}},
\bauthor{\bsnm{Jensen}, \binits{K.F.}}:
\batitle{A graph-convolutional neural network model for the prediction of chemical reactivity}.
\bjtitle{Chemical Science}
\bvolume{10}(\bissue{2}),
\bfpage{370}--\blpage{377}
(\byear{2019})
\doiurl{10.1039/C8SC04228D} .
\bcomment{Publisher: The Royal Society of Chemistry}
\end{barticle}
\endbibitem

\bibitem[\protect\citeauthoryear{Bradshaw et~al.}{2019}]{bradshaw_generative_2019}
\begin{bchapter}
\bauthor{\bsnm{Bradshaw}, \binits{J.}},
\bauthor{\bsnm{Kusner}, \binits{M.J.}},
\bauthor{\bsnm{Paige}, \binits{B.}},
\bauthor{\bsnm{Segler}, \binits{M.H.S.}},
\bauthor{\bsnm{Hernández-Lobato}, \binits{J.M.}}:
\bctitle{A {Generative} {Model} {For} {Electron} {Paths}}.
In: \bbtitle{The Seventh International Conference on Learning Representations}
(\byear{2019})
\end{bchapter}
\endbibitem

\bibitem[\protect\citeauthoryear{Bi et~al.}{2021}]{bi_non-autoregressive_2021}
\begin{bchapter}
\bauthor{\bsnm{Bi}, \binits{H.}},
\bauthor{\bsnm{Wang}, \binits{H.}},
\bauthor{\bsnm{Shi}, \binits{C.}},
\bauthor{\bsnm{Coley}, \binits{C.}},
\bauthor{\bsnm{Tang}, \binits{J.}},
\bauthor{\bsnm{Guo}, \binits{H.}}:
\bctitle{Non-{Autoregressive} {Electron} {Redistribution} {Modeling} for {Reaction} {Prediction}}.
In: \bbtitle{Proceedings of the 38th {International} {Conference} on {Machine} {Learning}},
pp. \bfpage{904}--\blpage{913}.
\bpublisher{PMLR},
\blocation{virtual}
(\byear{2021}).
\bcomment{ISSN: 2640-3498}
\end{bchapter}
\endbibitem

\bibitem[\protect\citeauthoryear{}{}]{Chemical.ai}
\begin{botherref}
Chemical.ai.
\url{https://chemical.ai/}
Accessed 2024-07-08
\end{botherref}
\endbibitem

\bibitem[\protect\citeauthoryear{}{}]{RXNforChemistry}
\begin{botherref}
RXN for Chemistry.
\url{https://rxn.app.accelerate.science/rxn}
Accessed 2024-07-08
\end{botherref}
\endbibitem

\bibitem[\protect\citeauthoryear{}{}]{Spaya}
\begin{botherref}
Spaya.
\url{https://spaya.ai/}
Accessed 2024-07-08
\end{botherref}
\endbibitem

\bibitem[\protect\citeauthoryear{}{}]{Manifold}
\begin{botherref}
MANIFOLD.
\url{https://app.postera.ai/}
Accessed 2024-08-27
\end{botherref}
\endbibitem

\bibitem[\protect\citeauthoryear{}{}]{Molecule.one}
\begin{botherref}
Molecule.one.
\url{https://www.molecule.one/}
Accessed 2024-08-27
\end{botherref}
\endbibitem

\bibitem[\protect\citeauthoryear{}{}]{ChemPlanner}
\begin{botherref}
Speed synthetic planning.
\url{https://www.cas.org/solutions/cas-scifinder-discovery-platform/cas-scifinder/synthesis-planning}
Accessed 2024-07-08
\end{botherref}
\endbibitem

\bibitem[\protect\citeauthoryear{}{}]{ReaxysRetro}
\begin{botherref}
Reaxys - Predictive Retrosynthesis.
\url{https://www.elsevier.com/products/reaxys/predictive-retrosynthesis}
Accessed 2024-11-18
\end{botherref}
\endbibitem

\bibitem[\protect\citeauthoryear{Coley et~al.}{2019}]{coley_robotic_2019}
\begin{barticle}
\bauthor{\bsnm{Coley}, \binits{C.W.}},
\bauthor{\bsnm{Thomas}, \binits{D.A.}},
\bauthor{\bsnm{Lummiss}, \binits{J.A.M.}},
\bauthor{\bsnm{Jaworski}, \binits{J.N.}},
\bauthor{\bsnm{Breen}, \binits{C.P.}},
\bauthor{\bsnm{Schultz}, \binits{V.}},
\bauthor{\bsnm{Hart}, \binits{T.}},
\bauthor{\bsnm{Fishman}, \binits{J.S.}},
\bauthor{\bsnm{Rogers}, \binits{L.}},
\bauthor{\bsnm{Gao}, \binits{H.}},
\bauthor{\bsnm{Hicklin}, \binits{R.W.}},
\bauthor{\bsnm{Plehiers}, \binits{P.P.}},
\bauthor{\bsnm{Byington}, \binits{J.}},
\bauthor{\bsnm{Piotti}, \binits{J.S.}},
\bauthor{\bsnm{Green}, \binits{W.H.}},
\bauthor{\bsnm{Hart}, \binits{A.J.}},
\bauthor{\bsnm{Jamison}, \binits{T.F.}},
\bauthor{\bsnm{Jensen}, \binits{K.F.}}:
\batitle{A robotic platform for flow synthesis of organic compounds informed by {AI} planning}.
\bjtitle{Science}
\bvolume{365}(\bissue{6453}),
\bfpage{1566}
(\byear{2019})
\doiurl{10.1126/science.aax1566} .
\bcomment{Publisher: American Association for the Advancement of Science}
\end{barticle}
\endbibitem

\bibitem[\protect\citeauthoryear{Genheden et~al.}{2020}]{genheden_aizynthfinder_2020}
\begin{barticle}
\bauthor{\bsnm{Genheden}, \binits{S.}},
\bauthor{\bsnm{Thakkar}, \binits{A.}},
\bauthor{\bsnm{Chadimová}, \binits{V.}},
\bauthor{\bsnm{Reymond}, \binits{J.-L.}},
\bauthor{\bsnm{Engkvist}, \binits{O.}},
\bauthor{\bsnm{Bjerrum}, \binits{E.}}:
\batitle{{AiZynthFinder}: a fast, robust and flexible open-source software for retrosynthetic planning}.
\bjtitle{Journal of Cheminformatics}
\bvolume{12}(\bissue{1}),
\bfpage{70}
(\byear{2020})
\doiurl{10.1186/s13321-020-00472-1}
\end{barticle}
\endbibitem

\bibitem[\protect\citeauthoryear{Maziarz et~al.}{2024}]{maziarz_re-evaluating_2024}
\begin{barticle}
\bauthor{\bsnm{Maziarz}, \binits{K.}},
\bauthor{\bsnm{Tripp}, \binits{A.}},
\bauthor{\bsnm{Liu}, \binits{G.}},
\bauthor{\bsnm{Stanley}, \binits{M.}},
\bauthor{\bsnm{Xie}, \binits{S.}},
\bauthor{\bsnm{Gainski}, \binits{P.}},
\bauthor{\bsnm{Seidl}, \binits{P.}},
\bauthor{\bsnm{Segler}, \binits{M.}}:
\batitle{Re-evaluating {Retrosynthesis} {Algorithms} with {Syntheseus}}.
\bjtitle{Faraday Discussions}
(\byear{2024})
\doiurl{10.1039/D4FD00093E} .
\bcomment{Publisher: The Royal Society of Chemistry}
\end{barticle}
\endbibitem

\bibitem[\protect\citeauthoryear{}{}]{MLPDS}
\begin{botherref}
MLPDS, Machine Learning for Pharmaceutical Discovery and Synthesis Consortium.
\url{https://mlpds.mit.edu}
Accessed 2024-07-08
\end{botherref}
\endbibitem

\bibitem[\protect\citeauthoryear{Struble et~al.}{2020}]{struble_current_2020}
\begin{barticle}
\bauthor{\bsnm{Struble}, \binits{T.J.}},
\bauthor{\bsnm{Alvarez}, \binits{J.C.}},
\bauthor{\bsnm{Brown}, \binits{S.P.}},
\bauthor{\bsnm{Chytil}, \binits{M.}},
\bauthor{\bsnm{Cisar}, \binits{J.}},
\bauthor{\bsnm{DesJarlais}, \binits{R.L.}},
\bauthor{\bsnm{Engkvist}, \binits{O.}},
\bauthor{\bsnm{Frank}, \binits{S.A.}},
\bauthor{\bsnm{Greve}, \binits{D.R.}},
\bauthor{\bsnm{Griffin}, \binits{D.J.}},
\bauthor{\bsnm{Hou}, \binits{X.}},
\bauthor{\bsnm{Johannes}, \binits{J.W.}},
\bauthor{\bsnm{Kreatsoulas}, \binits{C.}},
\bauthor{\bsnm{Lahue}, \binits{B.}},
\bauthor{\bsnm{Mathea}, \binits{M.}},
\bauthor{\bsnm{Mogk}, \binits{G.}},
\bauthor{\bsnm{Nicolaou}, \binits{C.A.}},
\bauthor{\bsnm{Palmer}, \binits{A.D.}},
\bauthor{\bsnm{Price}, \binits{D.J.}},
\bauthor{\bsnm{Robinson}, \binits{R.I.}},
\bauthor{\bsnm{Salentin}, \binits{S.}},
\bauthor{\bsnm{Xing}, \binits{L.}},
\bauthor{\bsnm{Jaakkola}, \binits{T.}},
\bauthor{\bsnm{Green}, \binits{W.H.}},
\bauthor{\bsnm{Barzilay}, \binits{R.}},
\bauthor{\bsnm{Coley}, \binits{C.W.}},
\bauthor{\bsnm{Jensen}, \binits{K.F.}}:
\batitle{Current and {Future} {Roles} of {Artificial} {Intelligence} in {Medicinal} {Chemistry} {Synthesis}}.
\bjtitle{Journal of Medicinal Chemistry}
\bvolume{63}(\bissue{16}),
\bfpage{8667}--\blpage{8682}
(\byear{2020})
\doiurl{10.1021/acs.jmedchem.9b02120} .
\bcomment{Publisher: American Chemical Society}
\end{barticle}
\endbibitem

\bibitem[\protect\citeauthoryear{Levin et~al.}{2022}]{levin_merging_2022}
\begin{barticle}
\bauthor{\bsnm{Levin}, \binits{I.}},
\bauthor{\bsnm{Liu}, \binits{M.}},
\bauthor{\bsnm{Voigt}, \binits{C.A.}},
\bauthor{\bsnm{Coley}, \binits{C.W.}}:
\batitle{Merging enzymatic and synthetic chemistry with computational synthesis planning}.
\bjtitle{Nature Communications}
\bvolume{13}(\bissue{1}),
\bfpage{7747}
(\byear{2022})
\doiurl{10.1038/s41467-022-35422-y} .
\bcomment{Number: 1 Publisher: Nature Publishing Group}
\end{barticle}
\endbibitem

\bibitem[\protect\citeauthoryear{Soukaina et~al.}{2024}]{soukaina_design_2024}
\begin{barticle}
\bauthor{\bsnm{Soukaina}, \binits{E.}},
\bauthor{\bsnm{Wissal}, \binits{L.}},
\bauthor{\bsnm{Yassine}, \binits{K.}},
\bauthor{\bsnm{Achraf}, \binits{E.A.}},
\bauthor{\bsnm{Guenoun}, \binits{F.}},
\bauthor{\bsnm{Bouachrine}, \binits{M.}}:
\batitle{Design of new dipeptide inhibitors against {SARS}-{CoV} {3CLpro}: {3D}-{QSAR}, molecular docking, {MD} simulation, {ADMET} studies and retrosynthesis strategy}.
\bjtitle{Arabian Journal of Chemistry}
\bvolume{17}(\bissue{2}),
\bfpage{105584}
(\byear{2024})
\doiurl{10.1016/j.arabjc.2023.105584}
\end{barticle}
\endbibitem

\bibitem[\protect\citeauthoryear{Avila et~al.}{2024}]{avila_chemistry_2024}
\begin{barticle}
\bauthor{\bsnm{Avila}, \binits{C.}},
\bauthor{\bsnm{West}, \binits{A.}},
\bauthor{\bsnm{C. Vicini}, \binits{A.}},
\bauthor{\bsnm{Waddington}, \binits{W.}},
\bauthor{\bsnm{Brearley}, \binits{C.}},
\bauthor{\bsnm{Clarke}, \binits{J.}},
\bauthor{\bsnm{M. Derrick}, \binits{A.}}:
\batitle{Chemistry in a graph: modern insights into commercial organic synthesis planning}.
\bjtitle{Digital Discovery}
(\byear{2024})
\doiurl{10.1039/D4DD00120F} .
\bcomment{Publisher: Royal Society of Chemistry}
\end{barticle}
\endbibitem

\bibitem[\protect\citeauthoryear{Fromer and Coley}{2024}]{fromer_algorithmic_2024}
\begin{barticle}
\bauthor{\bsnm{Fromer}, \binits{J.C.}},
\bauthor{\bsnm{Coley}, \binits{C.W.}}:
\batitle{An algorithmic framework for synthetic cost-aware decision making in molecular design}.
\bjtitle{Nature Computational Science}
\bvolume{4}(\bissue{6}),
\bfpage{440}--\blpage{450}
(\byear{2024})
\doiurl{10.1038/s43588-024-00639-y} .
\bcomment{Publisher: Nature Publishing Group}
\end{barticle}
\endbibitem

\bibitem[\protect\citeauthoryear{Sankaranarayanan and Jensen}{2023}]{sankaranarayanan_computer-assisted_2023}
\begin{barticle}
\bauthor{\bsnm{Sankaranarayanan}, \binits{K.}},
\bauthor{\bsnm{Jensen}, \binits{K.F.}}:
\batitle{Computer-assisted multistep chemoenzymatic retrosynthesis using a chemical synthesis planner}.
\bjtitle{Chemical Science}
\bvolume{14}(\bissue{23}),
\bfpage{6467}--\blpage{6475}
(\byear{2023})
\doiurl{10.1039/D3SC01355C} .
\bcomment{Publisher: The Royal Society of Chemistry}
\end{barticle}
\endbibitem

\bibitem[\protect\citeauthoryear{Koscher et~al.}{2023}]{koscher_autonomous_2023}
\begin{barticle}
\bauthor{\bsnm{Koscher}, \binits{B.A.}},
\bauthor{\bsnm{Canty}, \binits{R.B.}},
\bauthor{\bsnm{McDonald}, \binits{M.A.}},
\bauthor{\bsnm{Greenman}, \binits{K.P.}},
\bauthor{\bsnm{McGill}, \binits{C.J.}},
\bauthor{\bsnm{Bilodeau}, \binits{C.L.}},
\bauthor{\bsnm{Jin}, \binits{W.}},
\bauthor{\bsnm{Wu}, \binits{H.}},
\bauthor{\bsnm{Vermeire}, \binits{F.H.}},
\bauthor{\bsnm{Jin}, \binits{B.}},
\bauthor{\bsnm{Hart}, \binits{T.}},
\bauthor{\bsnm{Kulesza}, \binits{T.}},
\bauthor{\bsnm{Li}, \binits{S.-C.}},
\bauthor{\bsnm{Jaakkola}, \binits{T.S.}},
\bauthor{\bsnm{Barzilay}, \binits{R.}},
\bauthor{\bsnm{Gómez-Bombarelli}, \binits{R.}},
\bauthor{\bsnm{Green}, \binits{W.H.}},
\bauthor{\bsnm{Jensen}, \binits{K.F.}}:
\batitle{Autonomous, multiproperty-driven molecular discovery: {From} predictions to measurements and back}.
\bjtitle{Science}
\bvolume{382}(\bissue{6677}),
\bfpage{1407}
(\byear{2023})
\doiurl{10.1126/science.adi1407} .
\bcomment{Publisher: American Association for the Advancement of Science}
\end{barticle}
\endbibitem

\bibitem[\protect\citeauthoryear{Nambiar et~al.}{2022}]{nambiar_bayesian_2022}
\begin{barticle}
\bauthor{\bsnm{Nambiar}, \binits{A.M.K.}},
\bauthor{\bsnm{Breen}, \binits{C.P.}},
\bauthor{\bsnm{Hart}, \binits{T.}},
\bauthor{\bsnm{Kulesza}, \binits{T.}},
\bauthor{\bsnm{Jamison}, \binits{T.F.}},
\bauthor{\bsnm{Jensen}, \binits{K.F.}}:
\batitle{Bayesian {Optimization} of {Computer}-{Proposed} {Multistep} {Synthetic} {Routes} on an {Automated} {Robotic} {Flow} {Platform}}.
\bjtitle{ACS Central Science}
\bvolume{8}(\bissue{6}),
\bfpage{825}--\blpage{836}
(\byear{2022})
\doiurl{10.1021/acscentsci.2c00207} .
\bcomment{Publisher: American Chemical Society}
\end{barticle}
\endbibitem

\bibitem[\protect\citeauthoryear{Mahjour et~al.}{2024}]{mahjour_ideation_2024}
\begin{botherref}
\oauthor{\bsnm{Mahjour}, \binits{B.}},
\oauthor{\bsnm{Lu}, \binits{J.}},
\oauthor{\bsnm{Fromer}, \binits{J.}},
\oauthor{\bsnm{Casetti}, \binits{N.}},
\oauthor{\bsnm{Coley}, \binits{C.}}:
Ideation and {Evaluation} of {Novel} {Multicomponent} {Reactions} via {Mechanistic} {Network} {Analysis} and {Automation}
(2024).
\doiurl{10.26434/chemrxiv-2024-qfjh9-v2}
\end{botherref}
\endbibitem

\bibitem[\protect\citeauthoryear{Seierstad et~al.}{2021}]{seierstad_novel_2021}
\begin{barticle}
\bauthor{\bsnm{Seierstad}, \binits{M.}},
\bauthor{\bsnm{Tichenor}, \binits{M.S.}},
\bauthor{\bsnm{DesJarlais}, \binits{R.L.}},
\bauthor{\bsnm{Na}, \binits{J.}},
\bauthor{\bsnm{Bacani}, \binits{G.M.}},
\bauthor{\bsnm{Chung}, \binits{D.M.}},
\bauthor{\bsnm{Mercado-Marin}, \binits{E.V.}},
\bauthor{\bsnm{Steffens}, \binits{H.C.}},
\bauthor{\bsnm{Mirzadegan}, \binits{T.}}:
\batitle{Novel {Reagent} {Space}: {Identifying} {Unorderable} but {Readily} {Synthesizable} {Building} {Blocks}}.
\bjtitle{ACS Medicinal Chemistry Letters}
\bvolume{12}(\bissue{11}),
\bfpage{1853}--\blpage{1860}
(\byear{2021})
\doiurl{10.1021/acsmedchemlett.1c00340} .
\bcomment{Publisher: American Chemical Society}
\end{barticle}
\endbibitem

\bibitem[\protect\citeauthoryear{Qi et~al.}{2023}]{qi_optimized_2023}
\begin{barticle}
\bauthor{\bsnm{Qi}, \binits{W.}},
\bauthor{\bsnm{Zhai}, \binits{D.}},
\bauthor{\bsnm{Song}, \binits{D.}},
\bauthor{\bsnm{Liu}, \binits{C.}},
\bauthor{\bsnm{Yang}, \binits{J.}},
\bauthor{\bsnm{Sun}, \binits{L.}},
\bauthor{\bsnm{Li}, \binits{Y.}},
\bauthor{\bsnm{Li}, \binits{X.}},
\bauthor{\bsnm{Deng}, \binits{W.-Q.}}:
\batitle{Optimized synthesis of the anti-{COVID}-19 drugs aided by retrosynthesis software}.
\bjtitle{RSC Medicinal Chemistry}
(\byear{2023})
\doiurl{10.1039/D2MD00444E} .
\bcomment{Publisher: RSC}
\end{barticle}
\endbibitem

\bibitem[\protect\citeauthoryear{Pasquini and Stenta}{2023}]{pasquini2023linchemin}
\begin{barticle}
\bauthor{\bsnm{Pasquini}, \binits{M.}},
\bauthor{\bsnm{Stenta}, \binits{M.}}:
\batitle{Linchemin: Syngraph—a data model and a toolkit to analyze and compare synthetic routes}.
\bjtitle{Journal of Cheminformatics}
\bvolume{15}(\bissue{1}),
\bfpage{41}
(\byear{2023})
\end{barticle}
\endbibitem

\bibitem[\protect\citeauthoryear{Coley et~al.}{2018}]{coley_scscore_2018}
\begin{barticle}
\bauthor{\bsnm{Coley}, \binits{C.W.}},
\bauthor{\bsnm{Rogers}, \binits{L.}},
\bauthor{\bsnm{Green}, \binits{W.H.}},
\bauthor{\bsnm{Jensen}, \binits{K.F.}}:
\batitle{{SCScore}: {Synthetic} {Complexity} {Learned} from a {Reaction} {Corpus}}.
\bjtitle{Journal of Chemical Information and Modeling}
\bvolume{58}(\bissue{2}),
\bfpage{252}--\blpage{261}
(\byear{2018})
\doiurl{10.1021/acs.jcim.7b00622} .
\bcomment{Publisher: American Chemical Society}
\end{barticle}
\endbibitem

\bibitem[\protect\citeauthoryear{Segler and Waller}{2017}]{segler_neural-symbolic_2017}
\begin{barticle}
\bauthor{\bsnm{Segler}, \binits{M.H.S.}},
\bauthor{\bsnm{Waller}, \binits{M.P.}}:
\batitle{Neural-{Symbolic} {Machine} {Learning} for {Retrosynthesis} and {Reaction} {Prediction}}.
\bjtitle{Chemistry – A European Journal}
\bvolume{23}(\bissue{25}),
\bfpage{5966}--\blpage{5971}
(\byear{2017})
\doiurl{10.1002/chem.201605499}
\end{barticle}
\endbibitem

\bibitem[\protect\citeauthoryear{}{}]{Pistachio}
\begin{botherref}
The Pistachio dataset.
\url{https://www.nextmovesoftware.com/pistachio.html}
Accessed 2024-07-08
\end{botherref}
\endbibitem

\bibitem[\protect\citeauthoryear{}{}]{CASContent}
\begin{botherref}
The CAS reactions collection.
\url{https://www.cas.org/cas-data/cas-reactions}
Accessed 2024-07-08
\end{botherref}
\endbibitem

\bibitem[\protect\citeauthoryear{Lowe}{2012}]{lowe_extraction_2012}
\begin{botherref}
\oauthor{\bsnm{Lowe}, \binits{D.M.}}:
Extraction of chemical structures and reactions from the literature.
Thesis,
University of Cambridge
(October 2012).
\doiurl{10.17863/CAM.16293} .
Accepted: 2013-07-23T08:23:10Z
\end{botherref}
\endbibitem

\bibitem[\protect\citeauthoryear{}{}]{Reaxys}
\begin{botherref}
Reaxys.
\url{https://www.elsevier.com/promotions/chemistry-database}
Accessed 2024-07-08
\end{botherref}
\endbibitem

\bibitem[\protect\citeauthoryear{Coley et~al.}{2019}]{coley_rdchiral_2019}
\begin{barticle}
\bauthor{\bsnm{Coley}, \binits{C.W.}},
\bauthor{\bsnm{Green}, \binits{W.H.}},
\bauthor{\bsnm{Jensen}, \binits{K.F.}}:
\batitle{{RDChiral}: {An} {RDKit} {Wrapper} for {Handling} {Stereochemistry} in {Retrosynthetic} {Template} {Extraction} and {Application}}.
\bjtitle{Journal of Chemical Information and Modeling}
\bvolume{59}(\bissue{6}),
\bfpage{2529}--\blpage{2537}
(\byear{2019})
\doiurl{10.1021/acs.jcim.9b00286} .
\bcomment{Publisher: American Chemical Society}
\end{barticle}
\endbibitem

\bibitem[\protect\citeauthoryear{}{}]{BKMS}
\begin{botherref}
BKMS-react: an integrated and non-redundant biochemical reaction database.
\url{https://bkms.brenda-enzymes.org/index.php}
Accessed 2024-08-27
\end{botherref}
\endbibitem

\bibitem[\protect\citeauthoryear{Thakkar et~al.}{2020}]{thakkar_ring_2020}
\begin{barticle}
\bauthor{\bsnm{Thakkar}, \binits{A.}},
\bauthor{\bsnm{Selmi}, \binits{N.}},
\bauthor{\bsnm{Reymond}, \binits{J.-L.}},
\bauthor{\bsnm{Engkvist}, \binits{O.}},
\bauthor{\bsnm{Bjerrum}, \binits{E.J.}}:
\batitle{“{Ring} {Breaker}”: {Neural} {Network} {Driven} {Synthesis} {Prediction} of the {Ring} {System} {Chemical} {Space}}.
\bjtitle{Journal of Medicinal Chemistry}
\bvolume{63}(\bissue{16}),
\bfpage{8791}--\blpage{8808}
(\byear{2020})
\doiurl{10.1021/acs.jmedchem.9b01919} .
\bcomment{Publisher: American Chemical Society}
\end{barticle}
\endbibitem

\bibitem[\protect\citeauthoryear{Weininger}{1988}]{weininger_smiles_1988}
\begin{barticle}
\bauthor{\bsnm{Weininger}, \binits{D.}}:
\batitle{{SMILES}, a chemical language and information system. 1. {Introduction} to methodology and encoding rules}.
\bjtitle{Journal of Chemical Information and Computer Sciences}
\bvolume{28}(\bissue{1}),
\bfpage{31}--\blpage{36}
(\byear{1988})
\doiurl{10.1021/ci00057a005} .
\bcomment{Publisher: American Chemical Society}
\end{barticle}
\endbibitem

\bibitem[\protect\citeauthoryear{}{}]{PubChemAPI}
\begin{botherref}
PUG REST documentation.
\url{https://pubchem.ncbi.nlm.nih.gov/docs/pug-rest#section=Compound-Property-Tables}
Accessed 2024-07-09
\end{botherref}
\endbibitem

\bibitem[\protect\citeauthoryear{Lin et~al.}{2020}]{lin_automatic_2020}
\begin{barticle}
\bauthor{\bsnm{Lin}, \binits{K.}},
\bauthor{\bsnm{Xu}, \binits{Y.}},
\bauthor{\bsnm{Pei}, \binits{J.}},
\bauthor{\bsnm{Lai}, \binits{L.}}:
\batitle{Automatic retrosynthetic route planning using template-free models}.
\bjtitle{Chemical Science}
\bvolume{11}(\bissue{12}),
\bfpage{3355}--\blpage{3364}
(\byear{2020})
\doiurl{10.1039/C9SC03666K} .
\bcomment{Publisher: The Royal Society of Chemistry}
\end{barticle}
\endbibitem

\bibitem[\protect\citeauthoryear{Schneider et~al.}{2016}]{schneider_whats_2016}
\begin{barticle}
\bauthor{\bsnm{Schneider}, \binits{N.}},
\bauthor{\bsnm{Stiefl}, \binits{N.}},
\bauthor{\bsnm{Landrum}, \binits{G.A.}}:
\batitle{What’s {What}: {The} ({Nearly}) {Definitive} {Guide} to {Reaction} {Role} {Assignment}}.
\bjtitle{Journal of Chemical Information and Modeling}
\bvolume{56}(\bissue{12}),
\bfpage{2336}--\blpage{2346}
(\byear{2016})
\doiurl{10.1021/acs.jcim.6b00564} .
\bcomment{Publisher: American Chemical Society}
\end{barticle}
\endbibitem

\bibitem[\protect\citeauthoryear{Seidl et~al.}{2022}]{seidl_improving_2022}
\begin{barticle}
\bauthor{\bsnm{Seidl}, \binits{P.}},
\bauthor{\bsnm{Renz}, \binits{P.}},
\bauthor{\bsnm{Dyubankova}, \binits{N.}},
\bauthor{\bsnm{Neves}, \binits{P.}},
\bauthor{\bsnm{Verhoeven}, \binits{J.}},
\bauthor{\bsnm{Wegner}, \binits{J.K.}},
\bauthor{\bsnm{Segler}, \binits{M.}},
\bauthor{\bsnm{Hochreiter}, \binits{S.}},
\bauthor{\bsnm{Klambauer}, \binits{G.}}:
\batitle{Improving {Few}- and {Zero}-{Shot} {Reaction} {Template} {Prediction} {Using} {Modern} {Hopfield} {Networks}}.
\bjtitle{Journal of Chemical Information and Modeling}
\bvolume{62}(\bissue{9}),
\bfpage{2111}--\blpage{2120}
(\byear{2022})
\doiurl{10.1021/acs.jcim.1c01065} .
\bcomment{Publisher: American Chemical Society}
\end{barticle}
\endbibitem

\bibitem[\protect\citeauthoryear{Zhu et~al.}{2023}]{zhu_dual-view_2023}
\begin{bchapter}
\bauthor{\bsnm{Zhu}, \binits{J.}},
\bauthor{\bsnm{Xia}, \binits{Y.}},
\bauthor{\bsnm{Wu}, \binits{L.}},
\bauthor{\bsnm{Xie}, \binits{S.}},
\bauthor{\bsnm{Zhou}, \binits{W.}},
\bauthor{\bsnm{Qin}, \binits{T.}},
\bauthor{\bsnm{Li}, \binits{H.}},
\bauthor{\bsnm{Liu}, \binits{T.-Y.}}:
\bctitle{Dual-view {Molecular} {Pre}-training}.
In: \bbtitle{Proceedings of the 29th {ACM} {SIGKDD} {Conference} on {Knowledge} {Discovery} and {Data} {Mining}}.
\bsertitle{{KDD} '23},
pp. \bfpage{3615}--\blpage{3627}.
\bpublisher{Association for Computing Machinery},
\blocation{New York, NY, USA}
(\byear{2023}).
\doiurl{10.1145/3580305.3599317}
\end{bchapter}
\endbibitem

\bibitem[\protect\citeauthoryear{Bjerrum}{2017}]{bjerrum_smiles_2017}
\begin{botherref}
\oauthor{\bsnm{Bjerrum}, \binits{E.J.}}:
{SMILES} {Enumeration} as {Data} {Augmentation} for {Neural} {Network} {Modeling} of {Molecules}.
arXiv.
arXiv:1703.07076 [cs]
(2017).
\doiurl{10.48550/arXiv.1703.07076}
\end{botherref}
\endbibitem

\bibitem[\protect\citeauthoryear{Silver et~al.}{2016}]{silver_mastering_2016}
\begin{barticle}
\bauthor{\bsnm{Silver}, \binits{D.}},
\bauthor{\bsnm{Huang}, \binits{A.}},
\bauthor{\bsnm{Maddison}, \binits{C.J.}},
\bauthor{\bsnm{Guez}, \binits{A.}},
\bauthor{\bsnm{Sifre}, \binits{L.}},
\bauthor{\bsnm{Driessche}, \binits{G.}},
\bauthor{\bsnm{Schrittwieser}, \binits{J.}},
\bauthor{\bsnm{Antonoglou}, \binits{I.}},
\bauthor{\bsnm{Panneershelvam}, \binits{V.}},
\bauthor{\bsnm{Lanctot}, \binits{M.}},
\bauthor{\bsnm{Dieleman}, \binits{S.}},
\bauthor{\bsnm{Grewe}, \binits{D.}},
\bauthor{\bsnm{Nham}, \binits{J.}},
\bauthor{\bsnm{Kalchbrenner}, \binits{N.}},
\bauthor{\bsnm{Sutskever}, \binits{I.}},
\bauthor{\bsnm{Lillicrap}, \binits{T.}},
\bauthor{\bsnm{Leach}, \binits{M.}},
\bauthor{\bsnm{Kavukcuoglu}, \binits{K.}},
\bauthor{\bsnm{Graepel}, \binits{T.}},
\bauthor{\bsnm{Hassabis}, \binits{D.}}:
\batitle{Mastering the game of {Go} with deep neural networks and tree search}.
\bjtitle{Nature}
\bvolume{529}(\bissue{7587}),
\bfpage{484}--\blpage{489}
(\byear{2016})
\doiurl{10.1038/nature16961} .
\bcomment{Number: 7587 Publisher: Nature Publishing Group}
\end{barticle}
\endbibitem

\bibitem[\protect\citeauthoryear{Liu et~al.}{2023}]{liu_retrosynthetic_2023}
\begin{bchapter}
\bauthor{\bsnm{Liu}, \binits{G.}},
\bauthor{\bsnm{Xue}, \binits{D.}},
\bauthor{\bsnm{Xie}, \binits{S.}},
\bauthor{\bsnm{Xia}, \binits{Y.}},
\bauthor{\bsnm{Tripp}, \binits{A.}},
\bauthor{\bsnm{Maziarz}, \binits{K.}},
\bauthor{\bsnm{Segler}, \binits{M.}},
\bauthor{\bsnm{Qin}, \binits{T.}},
\bauthor{\bsnm{Zhang}, \binits{Z.}},
\bauthor{\bsnm{Liu}, \binits{T.-Y.}}:
\bctitle{Retrosynthetic {Planning} with {Dual} {Value} {Networks}}.
In: \bbtitle{Proceedings of the 40th {International} {Conference} on {Machine} {Learning}},
pp. \bfpage{22266}--\blpage{22276}.
\bpublisher{PMLR},
\blocation{Honolulu, Hawaii, US}
(\byear{2023}).
\bcomment{ISSN: 2640-3498}
\end{bchapter}
\endbibitem

\bibitem[\protect\citeauthoryear{Yu et~al.}{2022}]{yu_grasp_2022}
\begin{barticle}
\bauthor{\bsnm{Yu}, \binits{Y.}},
\bauthor{\bsnm{Wei}, \binits{Y.}},
\bauthor{\bsnm{Kuang}, \binits{K.}},
\bauthor{\bsnm{Huang}, \binits{Z.}},
\bauthor{\bsnm{Yao}, \binits{H.}},
\bauthor{\bsnm{Wu}, \binits{F.}}:
\batitle{{GRASP}: {Navigating} {Retrosynthetic} {Planning} with {Goal}-driven {Policy}}.
\bjtitle{Advances in Neural Information Processing Systems}
\bvolume{35},
\bfpage{10257}--\blpage{10268}
(\byear{2022})
\end{barticle}
\endbibitem

\bibitem[\protect\citeauthoryear{Schreck et~al.}{2019}]{schreck_learning_2019}
\begin{barticle}
\bauthor{\bsnm{Schreck}, \binits{J.S.}},
\bauthor{\bsnm{Coley}, \binits{C.W.}},
\bauthor{\bsnm{Bishop}, \binits{K.J.M.}}:
\batitle{Learning {Retrosynthetic} {Planning} through {Simulated} {Experience}}.
\bjtitle{ACS Central Science}
\bvolume{5}(\bissue{6}),
\bfpage{970}--\blpage{981}
(\byear{2019})
\doiurl{10.1021/acscentsci.9b00055} .
\bcomment{Publisher: American Chemical Society}
\end{barticle}
\endbibitem

\bibitem[\protect\citeauthoryear{Wang et~al.}{2020}]{wang_towards_2020}
\begin{barticle}
\bauthor{\bsnm{Wang}, \binits{X.}},
\bauthor{\bsnm{Qian}, \binits{Y.}},
\bauthor{\bsnm{Gao}, \binits{H.}},
\bauthor{\bsnm{W. Coley}, \binits{C.}},
\bauthor{\bsnm{Mo}, \binits{Y.}},
\bauthor{\bsnm{Barzilay}, \binits{R.}},
\bauthor{\bsnm{F. Jensen}, \binits{K.}}:
\batitle{Towards efficient discovery of green synthetic pathways with {Monte} {Carlo} tree search and reinforcement learning}.
\bjtitle{Chemical Science}
\bvolume{11}(\bissue{40}),
\bfpage{10959}--\blpage{10972}
(\byear{2020})
\doiurl{10.1039/D0SC04184J} .
\bcomment{Publisher: Royal Society of Chemistry}
\end{barticle}
\endbibitem

\bibitem[\protect\citeauthoryear{Xie et~al.}{2022}]{xie_retrograph_2022}
\begin{bchapter}
\bauthor{\bsnm{Xie}, \binits{S.}},
\bauthor{\bsnm{Yan}, \binits{R.}},
\bauthor{\bsnm{Han}, \binits{P.}},
\bauthor{\bsnm{Xia}, \binits{Y.}},
\bauthor{\bsnm{Wu}, \binits{L.}},
\bauthor{\bsnm{Guo}, \binits{C.}},
\bauthor{\bsnm{Yang}, \binits{B.}},
\bauthor{\bsnm{Qin}, \binits{T.}}:
\bctitle{{RetroGraph}: {Retrosynthetic} {Planning} with {Graph} {Search}}.
In: \bbtitle{Proceedings of the 28th {ACM} {SIGKDD} {Conference} on {Knowledge} {Discovery} and {Data} {Mining}}.
\bsertitle{{KDD} '22},
pp. \bfpage{2120}--\blpage{2129}.
\bpublisher{Association for Computing Machinery},
\blocation{New York, NY, USA}
(\byear{2022}).
\doiurl{10.1145/3534678.3539446}
\end{bchapter}
\endbibitem

\bibitem[\protect\citeauthoryear{Zhao et~al.}{2024}]{zhao_efficient_2024}
\begin{barticle}
\bauthor{\bsnm{Zhao}, \binits{D.}},
\bauthor{\bsnm{Tu}, \binits{S.}},
\bauthor{\bsnm{Xu}, \binits{L.}}:
\batitle{Efficient retrosynthetic planning with {MCTS} exploration enhanced {A}* search}.
\bjtitle{Communications Chemistry}
\bvolume{7}(\bissue{1}),
\bfpage{1}--\blpage{12}
(\byear{2024})
\doiurl{10.1038/s42004-024-01133-2} .
\bcomment{Publisher: Nature Publishing Group}
\end{barticle}
\endbibitem

\bibitem[\protect\citeauthoryear{Hong et~al.}{2023}]{hong_retrosynthetic_2023}
\begin{barticle}
\bauthor{\bsnm{Hong}, \binits{S.}},
\bauthor{\bsnm{Zhuo}, \binits{H.H.}},
\bauthor{\bsnm{Jin}, \binits{K.}},
\bauthor{\bsnm{Shao}, \binits{G.}},
\bauthor{\bsnm{Zhou}, \binits{Z.}}:
\batitle{Retrosynthetic planning with experience-guided {Monte} {Carlo} tree search}.
\bjtitle{Communications Chemistry}
\bvolume{6}(\bissue{1}),
\bfpage{1}--\blpage{14}
(\byear{2023})
\doiurl{10.1038/s42004-023-00911-8} .
\bcomment{Publisher: Nature Publishing Group}
\end{barticle}
\endbibitem

\bibitem[\protect\citeauthoryear{Kreutter and Reymond}{2023}]{kreutter_multistep_2023}
\begin{barticle}
\bauthor{\bsnm{Kreutter}, \binits{D.}},
\bauthor{\bsnm{Reymond}, \binits{J.-L.}}:
\batitle{Multistep retrosynthesis combining a disconnection aware triple transformer loop with a route penalty score guided tree search}.
\bjtitle{Chemical Science}
\bvolume{14}(\bissue{36}),
\bfpage{9959}--\blpage{9969}
(\byear{2023})
\doiurl{10.1039/D3SC01604H} .
\bcomment{Publisher: The Royal Society of Chemistry}
\end{barticle}
\endbibitem

\bibitem[\protect\citeauthoryear{Liu et~al.}{2023}]{liu_fusionretro_2023}
\begin{bchapter}
\bauthor{\bsnm{Liu}, \binits{S.}},
\bauthor{\bsnm{Tu}, \binits{Z.}},
\bauthor{\bsnm{Xu}, \binits{M.}},
\bauthor{\bsnm{Zhang}, \binits{Z.}},
\bauthor{\bsnm{Lin}, \binits{L.}},
\bauthor{\bsnm{Ying}, \binits{R.}},
\bauthor{\bsnm{Tang}, \binits{J.}},
\bauthor{\bsnm{Zhao}, \binits{P.}},
\bauthor{\bsnm{Wu}, \binits{D.}}:
\bctitle{{FusionRetro}: {Molecule} {Representation} {Fusion} via {In}-{Context} {Learning} for {Retrosynthetic} {Planning}}.
In: \bbtitle{Proceedings of the 40th {International} {Conference} on {Machine} {Learning}},
pp. \bfpage{22028}--\blpage{22041}.
\bpublisher{PMLR},
\blocation{Honolulu, Hawaii, US}
(\byear{2023}).
\bcomment{ISSN: 2640-3498}
\end{bchapter}
\endbibitem

\bibitem[\protect\citeauthoryear{Kim et~al.}{2021}]{kim_self-improved_2021}
\begin{bchapter}
\bauthor{\bsnm{Kim}, \binits{J.}},
\bauthor{\bsnm{Ahn}, \binits{S.}},
\bauthor{\bsnm{Lee}, \binits{H.}},
\bauthor{\bsnm{Shin}, \binits{J.}}:
\bctitle{Self-{Improved} {Retrosynthetic} {Planning}}.
In: \bbtitle{Proceedings of the 38th {International} {Conference} on {Machine} {Learning}},
pp. \bfpage{5486}--\blpage{5495}.
\bpublisher{PMLR},
\blocation{virtual}
(\byear{2021}).
\bcomment{ISSN: 2640-3498}
\end{bchapter}
\endbibitem

\bibitem[\protect\citeauthoryear{Shee et~al.}{2024}]{shee_directmultistep_2024}
\begin{botherref}
\oauthor{\bsnm{Shee}, \binits{Y.}},
\oauthor{\bsnm{Li}, \binits{H.}},
\oauthor{\bsnm{Morgunov}, \binits{A.}},
\oauthor{\bsnm{Batista}, \binits{V.}}:
DirectMultiStep: Direct Route Generation for Multi-Step Retrosynthesis.
arXiv
(2024).
\doiurl{10.48550/arXiv.2405.13983}
\end{botherref}
\endbibitem

\bibitem[\protect\citeauthoryear{Yu et~al.}{2024}]{yu_double-ended_2024}
\begin{bchapter}
\bauthor{\bsnm{Yu}, \binits{K.}},
\bauthor{\bsnm{Roh}, \binits{J.}},
\bauthor{\bsnm{Li}, \binits{Z.}},
\bauthor{\bsnm{Gao}, \binits{W.}},
\bauthor{\bsnm{Wang}, \binits{R.}},
\bauthor{\bsnm{Coley}, \binits{C.W.}}:
\bctitle{Double-{Ended} {Synthesis} {Planning} with {Goal}-{Constrained} {Bidirectional} {Search}}.
In: \bbtitle{Advances in {Neural} {Information} {Processing} {Systems}}.
\bpublisher{Curran Associates, Inc.},
\blocation{Vancouver, British Columbia, Canada}
(\byear{2024})
\end{bchapter}
\endbibitem

\bibitem[\protect\citeauthoryear{Marcou et~al.}{2015}]{marcou_expert_2015}
\begin{barticle}
\bauthor{\bsnm{Marcou}, \binits{G.}},
\bauthor{\bsnm{Sousa}, \binits{J.}},
\bauthor{\bsnm{Latino}, \binits{D.A.R.S.}},
\bauthor{\bsnm{Luca}, \binits{A.}},
\bauthor{\bsnm{Horvath}, \binits{D.}},
\bauthor{\bsnm{Rietsch}, \binits{V.}},
\bauthor{\bsnm{Varnek}, \binits{A.}}:
\batitle{Expert {System} for {Predicting} {Reaction} {Conditions}: {The} {Michael} {Reaction} {Case}}.
\bjtitle{Journal of Chemical Information and Modeling}
\bvolume{55}(\bissue{2}),
\bfpage{239}--\blpage{250}
(\byear{2015})
\doiurl{10.1021/ci500698a} .
\bcomment{Publisher: American Chemical Society}
\end{barticle}
\endbibitem

\bibitem[\protect\citeauthoryear{Li and Eastgate}{2019}]{li_making_2019}
\begin{barticle}
\bauthor{\bsnm{Li}, \binits{J.}},
\bauthor{\bsnm{Eastgate}, \binits{M.D.}}:
\batitle{Making better decisions during synthetic route design: leveraging prediction to achieve greenness-by-design}.
\bjtitle{Reaction Chemistry \& Engineering}
\bvolume{4}(\bissue{9}),
\bfpage{1595}--\blpage{1607}
(\byear{2019})
\doiurl{10.1039/C9RE00019D} .
\bcomment{Publisher: The Royal Society of Chemistry}
\end{barticle}
\endbibitem

\bibitem[\protect\citeauthoryear{Wang et~al.}{2023}]{wang_generic_2023}
\begin{botherref}
\oauthor{\bsnm{Wang}, \binits{X.}},
\oauthor{\bsnm{Hsieh}, \binits{C.-Y.}},
\oauthor{\bsnm{Yin}, \binits{X.}},
\oauthor{\bsnm{Wang}, \binits{J.}},
\oauthor{\bsnm{Li}, \binits{Y.}},
\oauthor{\bsnm{Deng}, \binits{Y.}},
\oauthor{\bsnm{Jiang}, \binits{D.}},
\oauthor{\bsnm{Wu}, \binits{Z.}},
\oauthor{\bsnm{Du}, \binits{H.}},
\oauthor{\bsnm{Chen}, \binits{H.}},
\oauthor{\bsnm{Li}, \binits{Y.}},
\oauthor{\bsnm{Liu}, \binits{H.}},
\oauthor{\bsnm{Wang}, \binits{Y.}},
\oauthor{\bsnm{Luo}, \binits{P.}},
\oauthor{\bsnm{Hou}, \binits{T.}},
\oauthor{\bsnm{Yao}, \binits{X.}}:
Generic {Interpretable} {Reaction} {Condition} {Predictions} with {Open} {Reaction} {Condition} {Datasets} and {Unsupervised} {Learning} of {Reaction} {Center}.
Research
\textbf{0}(ja)
(2023)
\doiurl{10.34133/research.0231} .
Publisher: American Association for the Advancement of Science
\end{botherref}
\endbibitem

\bibitem[\protect\citeauthoryear{Coley et~al.}{2017}]{coley_prediction_2017}
\begin{barticle}
\bauthor{\bsnm{Coley}, \binits{C.W.}},
\bauthor{\bsnm{Barzilay}, \binits{R.}},
\bauthor{\bsnm{Jaakkola}, \binits{T.S.}},
\bauthor{\bsnm{Green}, \binits{W.H.}},
\bauthor{\bsnm{Jensen}, \binits{K.F.}}:
\batitle{Prediction of {Organic} {Reaction} {Outcomes} {Using} {Machine} {Learning}}.
\bjtitle{ACS Central Science}
\bvolume{3}(\bissue{5}),
\bfpage{434}--\blpage{443}
(\byear{2017})
\doiurl{10.1021/acscentsci.7b00064} .
\bcomment{Publisher: American Chemical Society}
\end{barticle}
\endbibitem

\bibitem[\protect\citeauthoryear{Chen and Jung}{2022}]{chen_generalized-template-based_2022}
\begin{botherref}
\oauthor{\bsnm{Chen}, \binits{S.}},
\oauthor{\bsnm{Jung}, \binits{Y.}}:
A generalized-template-based graph neural network for accurate organic reactivity prediction.
Nature Machine Intelligence,
1--9
(2022)
\doiurl{10.1038/s42256-022-00526-z} .
Publisher: Nature Publishing Group
\end{botherref}
\endbibitem

\bibitem[\protect\citeauthoryear{Struble et~al.}{2020}]{struble_multitask_2020}
\begin{barticle}
\bauthor{\bsnm{Struble}, \binits{T.J.}},
\bauthor{\bsnm{Coley}, \binits{C.W.}},
\bauthor{\bsnm{Jensen}, \binits{K.F.}}:
\batitle{Multitask prediction of site selectivity in aromatic {C}–{H} functionalization reactions}.
\bjtitle{Reaction Chemistry \& Engineering}
\bvolume{5}(\bissue{5}),
\bfpage{896}--\blpage{902}
(\byear{2020})
\doiurl{10.1039/D0RE00071J} .
\bcomment{Publisher: The Royal Society of Chemistry}
\end{barticle}
\endbibitem

\bibitem[\protect\citeauthoryear{Guan et~al.}{2021}]{guan_regio-selectivity_2021}
\begin{barticle}
\bauthor{\bsnm{Guan}, \binits{Y.}},
\bauthor{\bsnm{Coley}, \binits{C.W.}},
\bauthor{\bsnm{Wu}, \binits{H.}},
\bauthor{\bsnm{Ranasinghe}, \binits{D.}},
\bauthor{\bsnm{Heid}, \binits{E.}},
\bauthor{\bsnm{Struble}, \binits{T.J.}},
\bauthor{\bsnm{Pattanaik}, \binits{L.}},
\bauthor{\bsnm{Green}, \binits{W.H.}},
\bauthor{\bsnm{Jensen}, \binits{K.F.}}:
\batitle{Regio-selectivity prediction with a machine-learned reaction representation and on-the-fly quantum mechanical descriptors}.
\bjtitle{Chemical Science}
\bvolume{12}(\bissue{6}),
\bfpage{2198}--\blpage{2208}
(\byear{2021})
\doiurl{10.1039/D0SC04823B} .
\bcomment{Publisher: The Royal Society of Chemistry}
\end{barticle}
\endbibitem

\bibitem[\protect\citeauthoryear{Hirama and Uei}{1982}]{hirama_carbamate_1982}
\begin{barticle}
\bauthor{\bsnm{Hirama}, \binits{M.}},
\bauthor{\bsnm{Uei}, \binits{M.}}:
\batitle{Carbamate mediated 1,3-asymmetric induction. {A} stereoselective synthesis of acyclic 1,3-diol systems}.
\bjtitle{Tetrahedron Letters}
\bvolume{23}(\bissue{50}),
\bfpage{5307}--\blpage{5310}
(\byear{1982})
\doiurl{10.1016/S0040-4039(00)85825-6}
\end{barticle}
\endbibitem

\bibitem[\protect\citeauthoryear{Neves et~al.}{2023}]{neves_global_2023}
\begin{barticle}
\bauthor{\bsnm{Neves}, \binits{P.}},
\bauthor{\bsnm{McClure}, \binits{K.}},
\bauthor{\bsnm{Verhoeven}, \binits{J.}},
\bauthor{\bsnm{Dyubankova}, \binits{N.}},
\bauthor{\bsnm{Nugmanov}, \binits{R.}},
\bauthor{\bsnm{Gedich}, \binits{A.}},
\bauthor{\bsnm{Menon}, \binits{S.}},
\bauthor{\bsnm{Shi}, \binits{Z.}},
\bauthor{\bsnm{Wegner}, \binits{J.K.}}:
\batitle{Global reactivity models are impactful in industrial synthesis applications}.
\bjtitle{Journal of Cheminformatics}
\bvolume{15}(\bissue{1}),
\bfpage{20}
(\byear{2023})
\doiurl{10.1186/s13321-023-00685-0}
\end{barticle}
\endbibitem

\bibitem[\protect\citeauthoryear{Mo et~al.}{2021}]{mo_evaluating_2021}
\begin{barticle}
\bauthor{\bsnm{Mo}, \binits{Y.}},
\bauthor{\bsnm{Guan}, \binits{Y.}},
\bauthor{\bsnm{Verma}, \binits{P.}},
\bauthor{\bsnm{Guo}, \binits{J.}},
\bauthor{\bsnm{Fortunato}, \binits{M.E.}},
\bauthor{\bsnm{Lu}, \binits{Z.}},
\bauthor{\bsnm{Coley}, \binits{C.W.}},
\bauthor{\bsnm{Jensen}, \binits{K.F.}}:
\batitle{Evaluating and clustering retrosynthesis pathways with learned strategy}.
\bjtitle{Chemical Science}
\bvolume{12}(\bissue{4}),
\bfpage{1469}--\blpage{1478}
(\byear{2021})
\doiurl{10.1039/D0SC05078D} .
\bcomment{Publisher: The Royal Society of Chemistry}
\end{barticle}
\endbibitem

\bibitem[\protect\citeauthoryear{}{}]{NameRXN}
\begin{botherref}
NameRxn: Expert System for Named Reaction Identification and Classification.
\url{https://www.nextmovesoftware.com/namerxn.html}
Accessed 2024-07-10
\end{botherref}
\endbibitem

\bibitem[\protect\citeauthoryear{Hoffmann}{2009}]{hoffmann_ranking_2009}
\begin{bchapter}
\bauthor{\bsnm{Hoffmann}, \binits{R.W.}}:
\bctitle{Ranking of {Synthesis} {Plans}}.
In: \beditor{\bsnm{Hoffmann}, \binits{R.W.}} (ed.)
\bbtitle{Elements of {Synthesis} {Planning}},
pp. \bfpage{133}--\blpage{144}.
\bpublisher{Springer},
\blocation{Berlin, Heidelberg}
(\byear{2009}).
\doiurl{10.1007/978-3-540-79220-8_8}
\end{bchapter}
\endbibitem

\bibitem[\protect\citeauthoryear{Levin et~al.}{2023}]{levin_computer-aided_2023}
\begin{barticle}
\bauthor{\bsnm{Levin}, \binits{I.}},
\bauthor{\bsnm{Fortunato}, \binits{M.E.}},
\bauthor{\bsnm{Tan}, \binits{K.L.}},
\bauthor{\bsnm{Coley}, \binits{C.W.}}:
\batitle{Computer-aided evaluation and exploration of chemical spaces constrained by reaction pathways}.
\bjtitle{AIChE Journal}
\bvolume{69}(\bissue{12}),
\bfpage{18234}
(\byear{2023})
\doiurl{10.1002/aic.18234}
\end{barticle}
\endbibitem

\bibitem[\protect\citeauthoryear{Kuznetsov and Sahinidis}{2021}]{kuznetsov_extractionscore_2021}
\begin{barticle}
\bauthor{\bsnm{Kuznetsov}, \binits{A.}},
\bauthor{\bsnm{Sahinidis}, \binits{N.V.}}:
\batitle{{ExtractionScore}: {A} {Quantitative} {Framework} for {Evaluating} {Synthetic} {Routes} on {Predicted} {Liquid}–{Liquid} {Extraction} {Performance}}.
\bjtitle{Journal of Chemical Information and Modeling}
\bvolume{61}(\bissue{5}),
\bfpage{2274}--\blpage{2282}
(\byear{2021})
\doiurl{10.1021/acs.jcim.0c01426} .
\bcomment{Publisher: American Chemical Society}
\end{barticle}
\endbibitem

\bibitem[\protect\citeauthoryear{Vermeire et~al.}{2022}]{vermeire_predicting_2022}
\begin{barticle}
\bauthor{\bsnm{Vermeire}, \binits{F.H.}},
\bauthor{\bsnm{Chung}, \binits{Y.}},
\bauthor{\bsnm{Green}, \binits{W.H.}}:
\batitle{Predicting {Solubility} {Limits} of {Organic} {Solutes} for a {Wide} {Range} of {Solvents} and {Temperatures}}.
\bjtitle{Journal of the American Chemical Society}
\bvolume{144}(\bissue{24}),
\bfpage{10785}--\blpage{10797}
(\byear{2022})
\doiurl{10.1021/jacs.2c01768} .
\bcomment{Publisher: American Chemical Society}
\end{barticle}
\endbibitem

\bibitem[\protect\citeauthoryear{Li et~al.}{2024}]{li_when_2024}
\begin{barticle}
\bauthor{\bsnm{Li}, \binits{S.-C.}},
\bauthor{\bsnm{Wu}, \binits{H.}},
\bauthor{\bsnm{Menon}, \binits{A.}},
\bauthor{\bsnm{Spiekermann}, \binits{K.A.}},
\bauthor{\bsnm{Li}, \binits{Y.-P.}},
\bauthor{\bsnm{Green}, \binits{W.H.}}:
\batitle{When {Do} {Quantum} {Mechanical} {Descriptors} {Help} {Graph} {Neural} {Networks} to {Predict} {Chemical} {Properties}?}
\bjtitle{Journal of the American Chemical Society}
\bvolume{146}(\bissue{33}),
\bfpage{23103}--\blpage{23120}
(\byear{2024})
\doiurl{10.1021/jacs.4c04670} .
\bcomment{Publisher: American Chemical Society}
\end{barticle}
\endbibitem

\bibitem[\protect\citeauthoryear{Griffin et~al.}{2023}]{griffin_opportunities_2023}
\begin{barticle}
\bauthor{\bsnm{Griffin}, \binits{D.J.}},
\bauthor{\bsnm{Coley}, \binits{C.W.}},
\bauthor{\bsnm{Frank}, \binits{S.A.}},
\bauthor{\bsnm{Hawkins}, \binits{J.M.}},
\bauthor{\bsnm{Jensen}, \binits{K.F.}}:
\batitle{Opportunities for {Machine} {Learning} and {Artificial} {Intelligence} to {Advance} {Synthetic} {Drug} {Substance} {Process} {Development}}.
\bjtitle{Organic Process Research \& Development}
(\byear{2023})
\doiurl{10.1021/acs.oprd.3c00229} .
\bcomment{Publisher: American Chemical Society}
\end{barticle}
\endbibitem

\bibitem[\protect\citeauthoryear{Stuyver and Coley}{2022}]{stuyver2022quantum}
\begin{botherref}
\oauthor{\bsnm{Stuyver}, \binits{T.}},
\oauthor{\bsnm{Coley}, \binits{C.W.}}:
Quantum chemistry-augmented neural networks for reactivity prediction: Performance, generalizability, and explainability.
The Journal of Chemical Physics
\textbf{156}(8)
(2022)
\end{botherref}
\endbibitem

\bibitem[\protect\citeauthoryear{Shields et~al.}{2024}]{shields_aizynth_2024}
\begin{barticle}
\bauthor{\bsnm{Shields}, \binits{J.D.}},
\bauthor{\bsnm{Howells}, \binits{R.}},
\bauthor{\bsnm{Lamont}, \binits{G.}},
\bauthor{\bsnm{Leilei}, \binits{Y.}},
\bauthor{\bsnm{Madin}, \binits{A.}},
\bauthor{\bsnm{Reimann}, \binits{C.E.}},
\bauthor{\bsnm{Rezaei}, \binits{H.}},
\bauthor{\bsnm{Reuillon}, \binits{T.}},
\bauthor{\bsnm{Smith}, \binits{B.}},
\bauthor{\bsnm{Thomson}, \binits{C.}},
\bauthor{\bsnm{Zheng}, \binits{Y.}},
\bauthor{\bsnm{Ziegler}, \binits{R.E.}}:
\batitle{{AiZynth} impact on medicinal chemistry practice at {AstraZeneca}}.
\bjtitle{RSC Medicinal Chemistry}
\bvolume{15}(\bissue{4}),
\bfpage{1085}--\blpage{1095}
(\byear{2024})
\doiurl{10.1039/D3MD00651D} .
\bcomment{Publisher: RSC}
\end{barticle}
\endbibitem

\bibitem[\protect\citeauthoryear{}{}]{EMolecules}
\begin{botherref}
eMolecules.
\url{https://www.emolecules.com/}
Accessed 2024-11-05
\end{botherref}
\endbibitem

\bibitem[\protect\citeauthoryear{}{}]{SigmaAldrich}
\begin{botherref}
MilliporeSigma.
\url{https://www.sigmaaldrich.com/}
Accessed 2024-11-05
\end{botherref}
\endbibitem

\bibitem[\protect\citeauthoryear{}{}]{LabNetwork}
\begin{botherref}
WuXi LabNetwork.
\url{https://www.labnetwork.com/}
Accessed 2024-11-05
\end{botherref}
\endbibitem

\bibitem[\protect\citeauthoryear{}{}]{Mcule}
\begin{botherref}
Mcule: Advanced Tools to Find and Order Molecules Online.
\url{https://mcule.com/}
Accessed 2024-11-05
\end{botherref}
\endbibitem

\bibitem[\protect\citeauthoryear{}{}]{ChemBridge}
\begin{botherref}
ChemBridge: The Gold Standard in Small Molecule Libraries and Building Blocks.
\url{https://chembridge.com/}
Accessed 2024-11-05
\end{botherref}
\endbibitem

\bibitem[\protect\citeauthoryear{Lee et~al.}{2019}]{lee_molecular_2019}
\begin{barticle}
\bauthor{\bsnm{Lee}, \binits{A.A.}},
\bauthor{\bsnm{Yang}, \binits{Q.}},
\bauthor{\bsnm{Sresht}, \binits{V.}},
\bauthor{\bsnm{Bolgar}, \binits{P.}},
\bauthor{\bsnm{Hou}, \binits{X.}},
\bauthor{\bsnm{Klug-McLeod}, \binits{J.L.}},
\bauthor{\bsnm{Butler}, \binits{C.R.}}:
\batitle{Molecular {Transformer} unifies reaction prediction and retrosynthesis across pharma chemical space}.
\bjtitle{Chemical Communications}
\bvolume{55}(\bissue{81}),
\bfpage{12152}--\blpage{12155}
(\byear{2019})
\doiurl{10.1039/C9CC05122H} .
\bcomment{Publisher: The Royal Society of Chemistry}
\end{barticle}
\endbibitem

\bibitem[\protect\citeauthoryear{}{}]{RDKit}
\begin{botherref}
RDKit: Open-Source Cheminformatics Software.
\url{https://www.rdkit.org/}
Accessed 2024-07-25
\end{botherref}
\endbibitem

\bibitem[\protect\citeauthoryear{Schwaller et~al.}{2021}]{schwaller_extraction_2021}
\begin{barticle}
\bauthor{\bsnm{Schwaller}, \binits{P.}},
\bauthor{\bsnm{Hoover}, \binits{B.}},
\bauthor{\bsnm{Reymond}, \binits{J.-L.}},
\bauthor{\bsnm{Strobelt}, \binits{H.}},
\bauthor{\bsnm{Laino}, \binits{T.}}:
\batitle{Extraction of organic chemistry grammar from unsupervised learning of chemical reactions}.
\bjtitle{Science Advances}
\bvolume{7}(\bissue{15}),
\bfpage{4166}
(\byear{2021})
\doiurl{10.1126/sciadv.abe4166} .
\bcomment{Publisher: American Association for the Advancement of Science}
\end{barticle}
\endbibitem

\bibitem[\protect\citeauthoryear{}{}]{Indigo}
\begin{botherref}
Indigo Toolkit.
\url{https://lifescience.opensource.epam.com/indigo/api/index.html}
Accessed 2024-07-25
\end{botherref}
\endbibitem

\bibitem[\protect\citeauthoryear{Srivastava et~al.}{2015}]{srivastava_training_2015}
\begin{bchapter}
\bauthor{\bsnm{Srivastava}, \binits{R.K.}},
\bauthor{\bsnm{Greff}, \binits{K.}},
\bauthor{\bsnm{Schmidhuber}, \binits{J.}}:
\bctitle{Training {Very} {Deep} {Networks}}.
In: \bbtitle{Advances in {Neural} {Information} {Processing} {Systems}},
vol. \bseriesno{28}.
\bpublisher{Curran Associates, Inc.},
\blocation{Montréal, Québec, Canada}
(\byear{2015})
\end{bchapter}
\endbibitem

\bibitem[\protect\citeauthoryear{}{}]{PyTorch}
\begin{botherref}
PyTorch.
\url{https://pytorch.org/}
Accessed 2024-07-25
\end{botherref}
\endbibitem

\bibitem[\protect\citeauthoryear{Kocsis and Szepesvári}{2006}]{kocsis_bandit_2006}
\begin{bchapter}
\bauthor{\bsnm{Kocsis}, \binits{L.}},
\bauthor{\bsnm{Szepesvári}, \binits{C.}}:
\bctitle{Bandit based monte-carlo planning}.
In: \bbtitle{Proceedings of the 17th {European} Conference on {Machine} {Learning}}.
\bsertitle{{ECML}'06},
pp. \bfpage{282}--\blpage{293}.
\bpublisher{Springer},
\blocation{Berlin, Heidelberg}
(\byear{2006}).
\doiurl{10.1007/11871842_29}
\end{bchapter}
\endbibitem

\bibitem[\protect\citeauthoryear{Klein et~al.}{2017}]{klein_opennmt_2017}
\begin{bchapter}
\bauthor{\bsnm{Klein}, \binits{G.}},
\bauthor{\bsnm{Kim}, \binits{Y.}},
\bauthor{\bsnm{Deng}, \binits{Y.}},
\bauthor{\bsnm{Senellart}, \binits{J.}},
\bauthor{\bsnm{Rush}, \binits{A.}}:
\bctitle{{OpenNMT}: {Open}-{Source} {Toolkit} for {Neural} {Machine} {Translation}}.
In: \beditor{\bsnm{Bansal}, \binits{M.}},
\beditor{\bsnm{Ji}, \binits{H.}} (eds.)
\bbtitle{Proceedings of {ACL} 2017, {System} {Demonstrations}},
pp. \bfpage{67}--\blpage{72}.
\bpublisher{Association for Computational Linguistics},
\blocation{Vancouver, Canada}
(\byear{2017})
\end{bchapter}
\endbibitem

\bibitem[\protect\citeauthoryear{Schwaller et~al.}{2018}]{schwaller_found_2018}
\begin{barticle}
\bauthor{\bsnm{Schwaller}, \binits{P.}},
\bauthor{\bsnm{Gaudin}, \binits{T.}},
\bauthor{\bsnm{Lányi}, \binits{D.}},
\bauthor{\bsnm{Bekas}, \binits{C.}},
\bauthor{\bsnm{Laino}, \binits{T.}}:
\batitle{“{Found} in {Translation}”: predicting outcomes of complex organic chemistry reactions using neural sequence-to-sequence models}.
\bjtitle{Chemical Science}
\bvolume{9}(\bissue{28}),
\bfpage{6091}--\blpage{6098}
(\byear{2018})
\doiurl{10.1039/C8SC02339E} .
\bcomment{Publisher: The Royal Society of Chemistry}
\end{barticle}
\endbibitem

\bibitem[\protect\citeauthoryear{}{}]{GLNRepo}
\begin{botherref}
The GLN repository.
\url{https://github.com/Hanjun-Dai/GLN/}
Accessed 2024-10-14
\end{botherref}
\endbibitem

\bibitem[\protect\citeauthoryear{}{}]{WLNRepo}
\begin{botherref}
The WLN repository.
\url{https://github.com/wengong-jin/nips17-rexgen/}
Accessed 2024-10-14
\end{botherref}
\endbibitem

\bibitem[\protect\citeauthoryear{}{}]{MTRepo}
\begin{botherref}
The Molecular Transformer repository.
\url{https://github.com/pschwllr/MolecularTransformer/}
Accessed 2024-10-14
\end{botherref}
\endbibitem

\end{thebibliography}



\begin{thebibliography}{26}
\ifx \bisbn   \undefined \def \bisbn  #1{ISBN #1}\fi
\ifx \binits  \undefined \def \binits#1{#1}\fi
\ifx \bauthor  \undefined \def \bauthor#1{#1}\fi
\ifx \batitle  \undefined \def \batitle#1{#1}\fi
\ifx \bjtitle  \undefined \def \bjtitle#1{#1}\fi
\ifx \bvolume  \undefined \def \bvolume#1{\textbf{#1}}\fi
\ifx \byear  \undefined \def \byear#1{#1}\fi
\ifx \bissue  \undefined \def \bissue#1{#1}\fi
\ifx \bfpage  \undefined \def \bfpage#1{#1}\fi
\ifx \blpage  \undefined \def \blpage #1{#1}\fi
\ifx \burl  \undefined \def \burl#1{\textsf{#1}}\fi
\ifx \doiurl  \undefined \def \doiurl#1{\url{https://doi.org/#1}}\fi
\ifx \betal  \undefined \def \betal{\textit{et al.}}\fi
\ifx \binstitute  \undefined \def \binstitute#1{#1}\fi
\ifx \binstitutionaled  \undefined \def \binstitutionaled#1{#1}\fi
\ifx \bctitle  \undefined \def \bctitle#1{#1}\fi
\ifx \beditor  \undefined \def \beditor#1{#1}\fi
\ifx \bpublisher  \undefined \def \bpublisher#1{#1}\fi
\ifx \bbtitle  \undefined \def \bbtitle#1{#1}\fi
\ifx \bedition  \undefined \def \bedition#1{#1}\fi
\ifx \bseriesno  \undefined \def \bseriesno#1{#1}\fi
\ifx \blocation  \undefined \def \blocation#1{#1}\fi
\ifx \bsertitle  \undefined \def \bsertitle#1{#1}\fi
\ifx \bsnm \undefined \def \bsnm#1{#1}\fi
\ifx \bsuffix \undefined \def \bsuffix#1{#1}\fi
\ifx \bparticle \undefined \def \bparticle#1{#1}\fi
\ifx \barticle \undefined \def \barticle#1{#1}\fi
\bibcommenthead
\ifx \bconfdate \undefined \def \bconfdate #1{#1}\fi
\ifx \botherref \undefined \def \botherref #1{#1}\fi
\ifx \url \undefined \def \url#1{\textsf{#1}}\fi
\ifx \bchapter \undefined \def \bchapter#1{#1}\fi
\ifx \bbook \undefined \def \bbook#1{#1}\fi
\ifx \bcomment \undefined \def \bcomment#1{#1}\fi
\ifx \oauthor \undefined \def \oauthor#1{#1}\fi
\ifx \citeauthoryear \undefined \def \citeauthoryear#1{#1}\fi
\ifx \endbibitem  \undefined \def \endbibitem {}\fi
\ifx \bconflocation  \undefined \def \bconflocation#1{#1}\fi
\ifx \arxivurl  \undefined \def \arxivurl#1{\textsf{#1}}\fi
\csname PreBibitemsHook\endcsname

\bibitem[\protect\citeauthoryear{Chung et~al.}{2022}]{chung_group_2022}
\begin{barticle}
\bauthor{\bsnm{Chung}, \binits{Y.}},
\bauthor{\bsnm{H.~Vermeire}, \binits{F.}},
\bauthor{\bsnm{Wu}, \binits{H.}},
\bauthor{\bsnm{J.~Walker}, \binits{P.}},
\bauthor{\bsnm{Abraham}, \binits{M.H.}},
\bauthor{\bsnm{Green}, \binits{W.H.}}:
\batitle{Group {Contribution} and {Machine} {Learning} {Approaches} to {Predict} {Abraham} {Solute} {Parameters}, {Solvation} {Free} {Energy}, and {Solvation} {Enthalpy}}.
\bjtitle{Journal of Chemical Information and Modeling}
\bvolume{62}(\bissue{3}),
\bfpage{433}--\blpage{446}
(\byear{2022})
\doiurl{10.1021/acs.jcim.1c01103} .
\bcomment{Publisher: American Chemical Society}
\end{barticle}
\endbibitem

\bibitem[\protect\citeauthoryear{Guan et~al.}{2021}]{guan_regio-selectivity_2021}
\begin{barticle}
\bauthor{\bsnm{Guan}, \binits{Y.}},
\bauthor{\bsnm{Coley}, \binits{C.W.}},
\bauthor{\bsnm{Wu}, \binits{H.}},
\bauthor{\bsnm{Ranasinghe}, \binits{D.}},
\bauthor{\bsnm{Heid}, \binits{E.}},
\bauthor{\bsnm{Struble}, \binits{T.J.}},
\bauthor{\bsnm{Pattanaik}, \binits{L.}},
\bauthor{\bsnm{Green}, \binits{W.H.}},
\bauthor{\bsnm{Jensen}, \binits{K.F.}}:
\batitle{Regio-selectivity prediction with a machine-learned reaction representation and on-the-fly quantum mechanical descriptors}.
\bjtitle{Chemical Science}
\bvolume{12}(\bissue{6}),
\bfpage{2198}--\blpage{2208}
(\byear{2021})
\doiurl{10.1039/D0SC04823B} .
\bcomment{Publisher: The Royal Society of Chemistry}
\end{barticle}
\endbibitem

\bibitem[\protect\citeauthoryear{Li et~al.}{2024}]{li_when_2024}
\begin{barticle}
\bauthor{\bsnm{Li}, \binits{S.-C.}},
\bauthor{\bsnm{Wu}, \binits{H.}},
\bauthor{\bsnm{Menon}, \binits{A.}},
\bauthor{\bsnm{Spiekermann}, \binits{K.A.}},
\bauthor{\bsnm{Li}, \binits{Y.-P.}},
\bauthor{\bsnm{Green}, \binits{W.H.}}:
\batitle{When {Do} {Quantum} {Mechanical} {Descriptors} {Help} {Graph} {Neural} {Networks} to {Predict} {Chemical} {Properties}?}
\bjtitle{Journal of the American Chemical Society}
\bvolume{146}(\bissue{33}),
\bfpage{23103}--\blpage{23120}
(\byear{2024})
\doiurl{10.1021/jacs.4c04670} .
\bcomment{Publisher: American Chemical Society}
\end{barticle}
\endbibitem

\bibitem[\protect\citeauthoryear{Stuyver and Coley}{2022}]{stuyver2022quantum}
\begin{botherref}
\oauthor{\bsnm{Stuyver}, \binits{T.}},
\oauthor{\bsnm{Coley}, \binits{C.W.}}:
Quantum chemistry-augmented neural networks for reactivity prediction: Performance, generalizability, and explainability.
The Journal of Chemical Physics
\textbf{156}(8)
(2022)
\end{botherref}
\endbibitem

\bibitem[\protect\citeauthoryear{}{}]{RDKit}
\begin{botherref}
RDKit: Open-Source Cheminformatics Software.
\url{https://www.rdkit.org/}
Accessed 2024-07-25
\end{botherref}
\endbibitem

\bibitem[\protect\citeauthoryear{}{}]{FastAPI}
\begin{botherref}
FastAPI.
\url{https://fastapi.tiangolo.com/}
Accessed 2024-07-25
\end{botherref}
\endbibitem

\bibitem[\protect\citeauthoryear{}{}]{TorchServe}
\begin{botherref}
TorchServe.
\url{https://pytorch.org/serve/}
Accessed 2024-07-25
\end{botherref}
\endbibitem

\bibitem[\protect\citeauthoryear{}{}]{ASKCOSwiki}
\begin{botherref}
The ASKCOS wiki.
\url{https://gitlab.com/mlpds_mit/askcosv2/askcos-docs/-/wikis/home}
Accessed 2024-07-09
\end{botherref}
\endbibitem

\bibitem[\protect\citeauthoryear{}{}]{Vue}
\begin{botherref}
Vue.js.
\url{https://vuejs.org/}
Accessed 2024-09-22
\end{botherref}
\endbibitem

\bibitem[\protect\citeauthoryear{}{}]{Vuetify}
\begin{botherref}
Vuetify.
\url{https://vuetifyjs.com/}
Accessed 2024-09-22
\end{botherref}
\endbibitem

\bibitem[\protect\citeauthoryear{}{}]{Cypress}
\begin{botherref}
Cypress.
\url{https://www.cypress.io/}
Accessed 2024-09-22
\end{botherref}
\endbibitem

\bibitem[\protect\citeauthoryear{}{}]{Pinia}
\begin{botherref}
Pinia: The intuitive store for Vue.js.
\url{https://pinia.vuejs.org/}
Accessed 2024-09-22
\end{botherref}
\endbibitem

\bibitem[\protect\citeauthoryear{}{}]{VueRouter}
\begin{botherref}
VueRouter.
\url{https://router.vuejs.org/}
Accessed 2024-09-22
\end{botherref}
\endbibitem

\bibitem[\protect\citeauthoryear{}{}]{NIHNameResolver}
\begin{botherref}
Chemical Identifier Resolver from NIH.
\url{https://cactus.nci.nih.gov/chemical/structure}
Accessed 2024-11-04
\end{botherref}
\endbibitem

\bibitem[\protect\citeauthoryear{}{}]{FDAApproval}
\begin{botherref}
Compilation of CDER New Molecular Entity (NME) Drug and New Biologic Approvals.
\url{https://www.fda.gov/drugs/drug-approvals-and-databases/compilation-cder-new-molecular-entity-nme-drug-and-new-biologic-approvals/}
Accessed 2024-10-14
\end{botherref}
\endbibitem

\bibitem[\protect\citeauthoryear{Anderson et~al.}{1985}]{anderson_pyrrole_1985}
\begin{barticle}
\bauthor{\bsnm{Anderson}, \binits{H.J.}},
\bauthor{\bsnm{Loader}, \binits{C.E.}},
\bauthor{\bsnm{Xu}, \binits{R.X.}},
\bauthor{\bsnm{Lê}, \binits{N.}},
\bauthor{\bsnm{Gogan}, \binits{N.J.}},
\bauthor{\bsnm{McDonald}, \binits{R.}},
\bauthor{\bsnm{Edwards}, \binits{L.G.}}:
\batitle{Pyrrole chemistry. {XXVIII}. {Substitution} reactions of 1-(phenylsulfonyl)pyrrole and some derivatives}.
\bjtitle{Canadian Journal of Chemistry}
\bvolume{63}(\bissue{4}),
\bfpage{896}--\blpage{902}
(\byear{1985})
\doiurl{10.1139/v85-149} .
\bcomment{Publisher: NRC Research Press}
\end{barticle}
\endbibitem

\bibitem[\protect\citeauthoryear{Elliott et~al.}{2007}]{elliott_intramolecular_2007}
\begin{barticle}
\bauthor{\bsnm{Elliott}, \binits{L.D.}},
\bauthor{\bsnm{Berry}, \binits{M.}},
\bauthor{\bsnm{Orr-Ewing}, \binits{A.J.}},
\bauthor{\bsnm{Booker-Milburn}, \binits{K.I.}}:
\batitle{The {Intramolecular} {Photometathesis} of {Pyrroles}}.
\bjtitle{Journal of the American Chemical Society}
\bvolume{129}(\bissue{11}),
\bfpage{3078}--\blpage{3079}
(\byear{2007})
\doiurl{10.1021/ja070254l} .
\bcomment{Publisher: American Chemical Society}
\end{barticle}
\endbibitem

\bibitem[\protect\citeauthoryear{Koovits et~al.}{2016}]{koovits_conformationally_2016}
\begin{barticle}
\bauthor{\bsnm{Koovits}, \binits{P.J.}},
\bauthor{\bsnm{Knowles}, \binits{J.P.}},
\bauthor{\bsnm{Booker-Milburn}, \binits{K.I.}}:
\batitle{Conformationally {Driven} {Two}- and {Three}-{Photon} {Cascade} {Processes} in the {Stereoselective} {Photorearrangement} of {Pyrroles}}.
\bjtitle{Organic Letters}
\bvolume{18}(\bissue{21}),
\bfpage{5608}--\blpage{5611}
(\byear{2016})
\doiurl{10.1021/acs.orglett.6b02829} .
\bcomment{Publisher: American Chemical Society}
\end{barticle}
\endbibitem

\bibitem[\protect\citeauthoryear{Boogaard et~al.}{1994}]{boogaard_ring_1994}
\begin{barticle}
\bauthor{\bsnm{Boogaard}, \binits{A.T.}},
\bauthor{\bsnm{Pandit}, \binits{U.K.}},
\bauthor{\bsnm{Koome}, \binits{G.-J.}}:
\batitle{Ring {D} modifications of ellipticine. {Part} 1. {New} ellipticine derivatives from 1-cyano-6-methylellipticine.}
\bjtitle{Tetrahedron}
\bvolume{50}(\bissue{8}),
\bfpage{2551}--\blpage{2560}
(\byear{1994})
\doiurl{10.1016/S0040-4020(01)86971-4}
\end{barticle}
\endbibitem

\bibitem[\protect\citeauthoryear{Ornstein et~al.}{1991}]{ornstein_4-tetrazolylalkylpiperidine-2-carboxylic_1991}
\begin{barticle}
\bauthor{\bsnm{Ornstein}, \binits{P.L.}},
\bauthor{\bsnm{Schoepp}, \binits{D.D.}},
\bauthor{\bsnm{Arnold}, \binits{M.B.}},
\bauthor{\bsnm{Leander}, \binits{J.D.}},
\bauthor{\bsnm{Lodge}, \binits{D.}},
\bauthor{\bsnm{Paschal}, \binits{J.W.}},
\bauthor{\bsnm{Elzey}, \binits{T.}}:
\batitle{4-({Tetrazolylalkyl})piperidine-2-carboxylic acids. {Potent} and selective {N}-methyl-{D}-aspartic acid receptor antagonists with a short duration of action}.
\bjtitle{Journal of Medicinal Chemistry}
\bvolume{34}(\bissue{1}),
\bfpage{90}--\blpage{97}
(\byear{1991})
\doiurl{10.1021/jm00105a016} .
\bcomment{Publisher: American Chemical Society}
\end{barticle}
\endbibitem

\bibitem[\protect\citeauthoryear{Price et~al.}{2018}]{price_orally_2018}
\begin{barticle}
\bauthor{\bsnm{Price}, \binits{D.J.}},
\bauthor{\bsnm{Drewry}, \binits{D.H.}},
\bauthor{\bsnm{Schaller}, \binits{L.T.}},
\bauthor{\bsnm{Thompson}, \binits{B.D.}},
\bauthor{\bsnm{Reid}, \binits{P.R.}},
\bauthor{\bsnm{Maloney}, \binits{P.R.}},
\bauthor{\bsnm{Liang}, \binits{X.}},
\bauthor{\bsnm{Banker}, \binits{P.}},
\bauthor{\bsnm{Buckholz}, \binits{R.G.}},
\bauthor{\bsnm{Selley}, \binits{P.K.}},
\bauthor{\bsnm{McDonald}, \binits{O.B.}},
\bauthor{\bsnm{Smith}, \binits{J.L.}},
\bauthor{\bsnm{Shearer}, \binits{T.W.}},
\bauthor{\bsnm{Cox}, \binits{R.F.}},
\bauthor{\bsnm{Williams}, \binits{S.P.}},
\bauthor{\bsnm{Reid}, \binits{R.A.}},
\bauthor{\bsnm{Tacconi}, \binits{S.}},
\bauthor{\bsnm{Faggioni}, \binits{F.}},
\bauthor{\bsnm{Piubelli}, \binits{C.}},
\bauthor{\bsnm{Sartori}, \binits{I.}},
\bauthor{\bsnm{Tessari}, \binits{M.}},
\bauthor{\bsnm{Wang}, \binits{T.Y.}}:
\batitle{An orally available, brain-penetrant {CAMKK2} inhibitor reduces food intake in rodent model}.
\bjtitle{Bioorganic \& Medicinal Chemistry Letters}
\bvolume{28}(\bissue{10}),
\bfpage{1958}--\blpage{1963}
(\byear{2018})
\doiurl{10.1016/j.bmcl.2018.03.034}
\end{barticle}
\endbibitem

\bibitem[\protect\citeauthoryear{Henry et~al.}{1989}]{henry1989mitsunobu}
\begin{barticle}
\bauthor{\bsnm{Henry}, \binits{J.R.}},
\bauthor{\bsnm{Marcin}, \binits{L.R.}},
\bauthor{\bsnm{McIntosh}, \binits{M.C.}},
\bauthor{\bsnm{Scola}, \binits{P.M.}},
\bauthor{\bsnm{Harris~Jr}, \binits{G.D.}},
\bauthor{\bsnm{Weinreb}, \binits{S.M.}}:
\batitle{Mitsunobu reactions of n-alkyl and n-acyl sulfonamides-an efficient route to protected amines}.
\bjtitle{Tetrahedron letters}
\bvolume{30}(\bissue{42}),
\bfpage{5709}--\blpage{5712}
(\byear{1989})
\end{barticle}
\endbibitem

\bibitem[\protect\citeauthoryear{Thakkar et~al.}{2020}]{thakkar_ring_2020}
\begin{barticle}
\bauthor{\bsnm{Thakkar}, \binits{A.}},
\bauthor{\bsnm{Selmi}, \binits{N.}},
\bauthor{\bsnm{Reymond}, \binits{J.-L.}},
\bauthor{\bsnm{Engkvist}, \binits{O.}},
\bauthor{\bsnm{Bjerrum}, \binits{E.J.}}:
\batitle{“{Ring} {Breaker}”: {Neural} {Network} {Driven} {Synthesis} {Prediction} of the {Ring} {System} {Chemical} {Space}}.
\bjtitle{Journal of Medicinal Chemistry}
\bvolume{63}(\bissue{16}),
\bfpage{8791}--\blpage{8808}
(\byear{2020})
\doiurl{10.1021/acs.jmedchem.9b01919} .
\bcomment{Publisher: American Chemical Society}
\end{barticle}
\endbibitem

\bibitem[\protect\citeauthoryear{}{}]{Pistachio}
\begin{botherref}
The Pistachio dataset.
\url{https://www.nextmovesoftware.com/pistachio.html}
Accessed 2024-07-08
\end{botherref}
\endbibitem

\bibitem[\protect\citeauthoryear{}{}]{CASContent}
\begin{botherref}
The CAS reactions collection.
\url{https://www.cas.org/cas-data/cas-reactions}
Accessed 2024-07-08
\end{botherref}
\endbibitem

\bibitem[\protect\citeauthoryear{Antonchick and Burgmann}{2013}]{antonchick_direct_2013}
\begin{barticle}
\bauthor{\bsnm{Antonchick}, \binits{A.P.}},
\bauthor{\bsnm{Burgmann}, \binits{L.}}:
\batitle{Direct {Selective} {Oxidative} {Cross}-{Coupling} of {Simple} {Alkanes} with {Heteroarenes}}.
\bjtitle{Angewandte Chemie International Edition}
\bvolume{52}(\bissue{11}),
\bfpage{3267}--\blpage{3271}
(\byear{2013})
\doiurl{10.1002/anie.201209584} .
\bcomment{\_eprint: https://onlinelibrary.wiley.com/doi/pdf/10.1002/anie.201209584}
\end{barticle}
\endbibitem

\end{thebibliography}

\end{document}


\title[Article Title]{Supplementary information for ASKCOS: an open source software suite for synthesis planning}

\author[1]{\fnm{Zhengkai} \sur{Tu}}\email{ztu@mit.edu}
\author[2]{\fnm{Sourabh J.} \sur{Choure}}\email{sjchoure@mit.edu}
\author[2]{\fnm{Mun Hong} \sur{Fong}}\email{fong410@mit.edu}
\author[2]{\fnm{Jihye} \sur{Roh}}\email{jroh99@mit.edu}
\author[3]{\fnm{Itai} \sur{Levin}}\email{itail@mit.edu}
\author[4]{\fnm{Kevin} \sur{Yu}}\email{kyu3@mit.edu}
\author[2]{\fnm{Joonyoung F.} \sur{Joung}}\email{jjoung@mit.edu}
\author[2]{\fnm{Nathan} \sur{Morgan}}\email{knathan@mit.edu}
\author[2]{\fnm{Shih-Cheng} \sur{Li}}\email{scli@mit.edu}
\author[2]{\fnm{Xiaoqi} \sur{Sun}}\email{xiaoqis@mit.edu}
\author[2]{\fnm{Huiqian} \sur{Lin}}\email{linhq@mit.edu}
\author[2]{\fnm{Mark} \sur{Murnin}}\email{murninm@mit.edu}
\author[2]{\fnm{Jordan P.} \sur{Liles}}\email{jliles24@mit.edu}
\author[5]{\fnm{Thomas J.} \sur{Struble}}\email{Thomas.Struble@bms.com}
\author[6]{\fnm{Michael E.} \sur{Fortunato}}\email{mike.fortunato@novartis.com}
\author{\fnm{Mengjie} \sur{Liu}\textsuperscript{2,}\footnote[2]{Current affiliation: AstraZeneca. Work done while at MIT.}}\email{mjliu@mit.edu}
\author[2]{\fnm{William H.} \sur{Green}}\email{whgreen@mit.edu}
\author[2]{\fnm{Klavs F.} \sur{Jensen}}\email{kfjensen@mit.edu}
\author*[1,2]{\fnm{Connor W.} \sur{Coley}}\email{ccoley@mit.edu}

\affil[1]{\orgdiv{Department of Electrical Engineering and Computer Science}, \orgname{Massachusetts Institute of Technology}, \orgaddress{\street{77 Massachusetts Ave}, \city{Cambridge}, \state{MA}, \postcode{02139}, \country{USA}}}

\affil[2]{\orgdiv{Department of Chemical Engineering}, \orgname{Massachusetts Institute of Technology}, \orgaddress{\street{77 Massachusetts Ave}, \city{Cambridge}, \state{MA}, \postcode{02139}, \country{USA}}}

\affil[3]{\orgdiv{Department of Biological Engineering}, \orgname{Massachusetts Institute of Technology}, \orgaddress{\street{77 Massachusetts Ave}, \city{Cambridge}, \state{MA}, \postcode{02139}, \country{USA}}}

\affil[4]{\orgdiv{Center for Computational Science and Engineering}, \orgname{Massachusetts Institute of Technology}, \orgaddress{\street{77 Massachusetts Ave}, \city{Cambridge}, \state{MA}, \postcode{02139}, \country{USA}}}

\affil[5]{\orgname{Bristol Myers Squibb}, \orgaddress{\street{250 Water Street, \city{Cambridge}, \state{MA}, \postcode{02141}, \country{USA}}}}

\affil[6]{\orgname{Novartis Institutes for BioMedical Research, Inc.}, \orgaddress{\street{250 Massachusetts Avenue}, \city{Cambridge}, \state{MA}, \postcode{02139}, \country{USA}}}

\maketitle

\section{Details of solubility prediction and solvent screening}\label{results_solubility}

\begin{figure}[h!]
\centering
\includegraphics[width=1.0\textwidth]{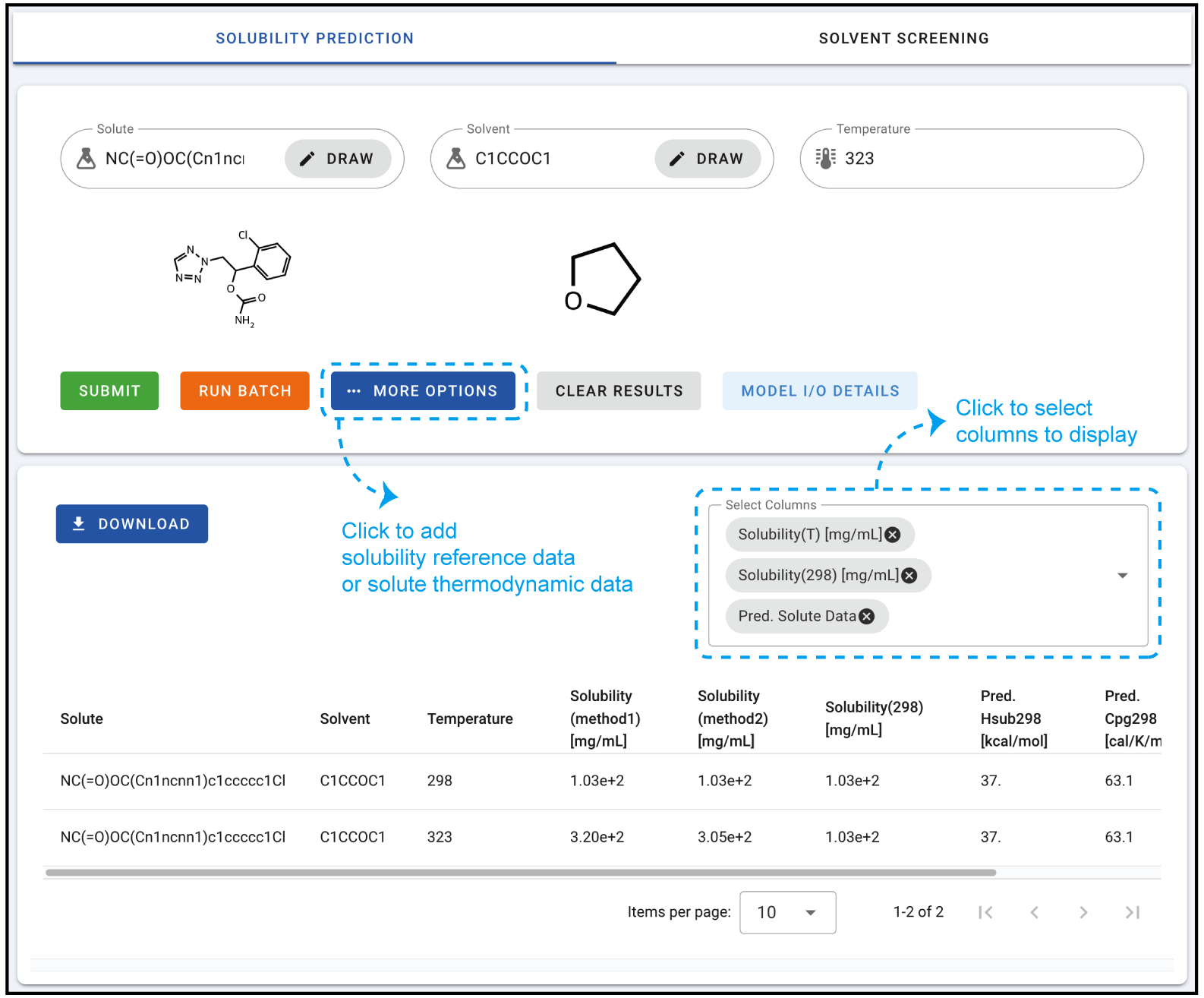}
\caption{Annotated screenshot of solubility prediction results in ASKCOS. The SMILES strings of the solute and the solvent are input in the \texttt{Solute} and \texttt{Solvent} panels. The desired temperature (in K) for the solubility prediction is input in the \texttt{Temperature} panel. The predicted solubility and solvation properties of solute cenobamate, defined by the SMILES string \texttt{NC(=O)OC(Cn1ncnn1)c1ccccc1Cl}, in solvent THF, defined by the SMILES string \texttt{C1CCOC1}, at 298 K and 323 K are displayed. The property values to display (e.g., $\log S$, Abraham parameters, etc.) can be selected from the \texttt{Select Columns} panel. }\label{fig_solubility_prediction}
\end{figure}

The required input for the solubility prediction model is a solute and a solvent molecule, which are specified using a SMILES string or by using the drawing functionality (Figure \ref{fig_solubility_prediction}). After clicking the \texttt{SUBMIT} button, the predicted solubility in milligrams per milliliter is shown in a table at the bottom of the screen. The default prediction temperature is 298 K, but this can be adjusted by using the temperature field on the main screen. 
Other optional input fields are found by clicking \texttt{MORE OPTIONS}. This opens a window where users can supply a few thermodynamic parameters to improve the model's predictions. The model makes solubility predictions by first predicting the aqueous solubility at 298 K ($\log\left(S_{aq, 298\, \mathrm{K}}\right)$) and then by using other thermochemical quantities to correct that solubility to the desired solvent and temperature. The known solubility of the solute in another solvent (a reference solubility) can be given to the model to use instead of the aqueous solubility prediction, which is especially useful for solutes that have low aqueous solubility. The solubility is corrected from the reference solvent to the target solvent using ML predicted solvation free energies ($\Delta G_{solv, 298\, \mathrm{K}}$) in both solvents. $\Delta G_{solv, 298\, \mathrm{K}}$ in the target solvent can also be viewed in the output table by selecting its column in the drop down menu, which is found by clicking the arrow in the \texttt{Select Columns} field. 

The solubility in the target solvent at 298 K, $\log\left(S_{298\, \mathrm{K}}\right)$, is corrected to other temperatures using the solute's dissolution enthalpy in the target solvent, $\Delta H_{diss, \mathrm{T}}$, which in turn is estimated using ML predictions of more common thermochemical quantities in a thermodynamic cycle. These quantities include the solute's sublimation enthalpy at 298 K ($\Delta H_{sub, 298 \,\mathrm{K}}$), solid phase heat capacity at 298 K ($C_{p,s}$), gas phase heat capacity at 298 K ($C_{p,g}$), and solvation enthalpy either at 298 K ($\Delta H_{solv, 298\, \mathrm{K}}$) or the target temperature ($\Delta H_{solv,\mathrm{T}}$). All of these can be viewed in the output table by selecting them in the \texttt{Select Columns} drop down menu, similar to $\Delta G_{solv, 298\, \mathrm{K}}$. The first three quantities are predicted via correlations that use ML predicted Abraham solute parameters~\citep{chung_group_2022}. These parameters can also be viewed in the output table. Known values for the first three quantities can be given to the model to use instead of the correlations from the solute parameters. This is done using the same window as the reference solubility. 

The $\Delta H_{solv}$ used in the thermodynamic cycle can be predicted in two ways, which gives two temperature dependent solubility estimates. These are labeled ``method1" and ``method2" in the output table. The first method neglects the temperature dependence of $\Delta H_{solv}$ and uses an ML model to predict $\Delta H_{solv, 298\, \mathrm{K}}$, which is used as $\Delta H_{solv,\mathrm{T}}$. This works best for temperatures below 350 K. The second method uses an ML model to predict $\Delta G_{solv, \mathrm{T}}$, from which $\Delta H_{solv, \mathrm{T}}$ and the temperature dependent entropy ($\Delta S_{solv, \mathrm{T}}$) can be calculated. This model requires the critical temperature and density of the solvent, which limits this method's use to around 100 solvents. These temperature dependent quantities can be viewed in the output table for supported solvents. 

\begin{figure}[h!]
\centering
\includegraphics[width=1.0\textwidth]{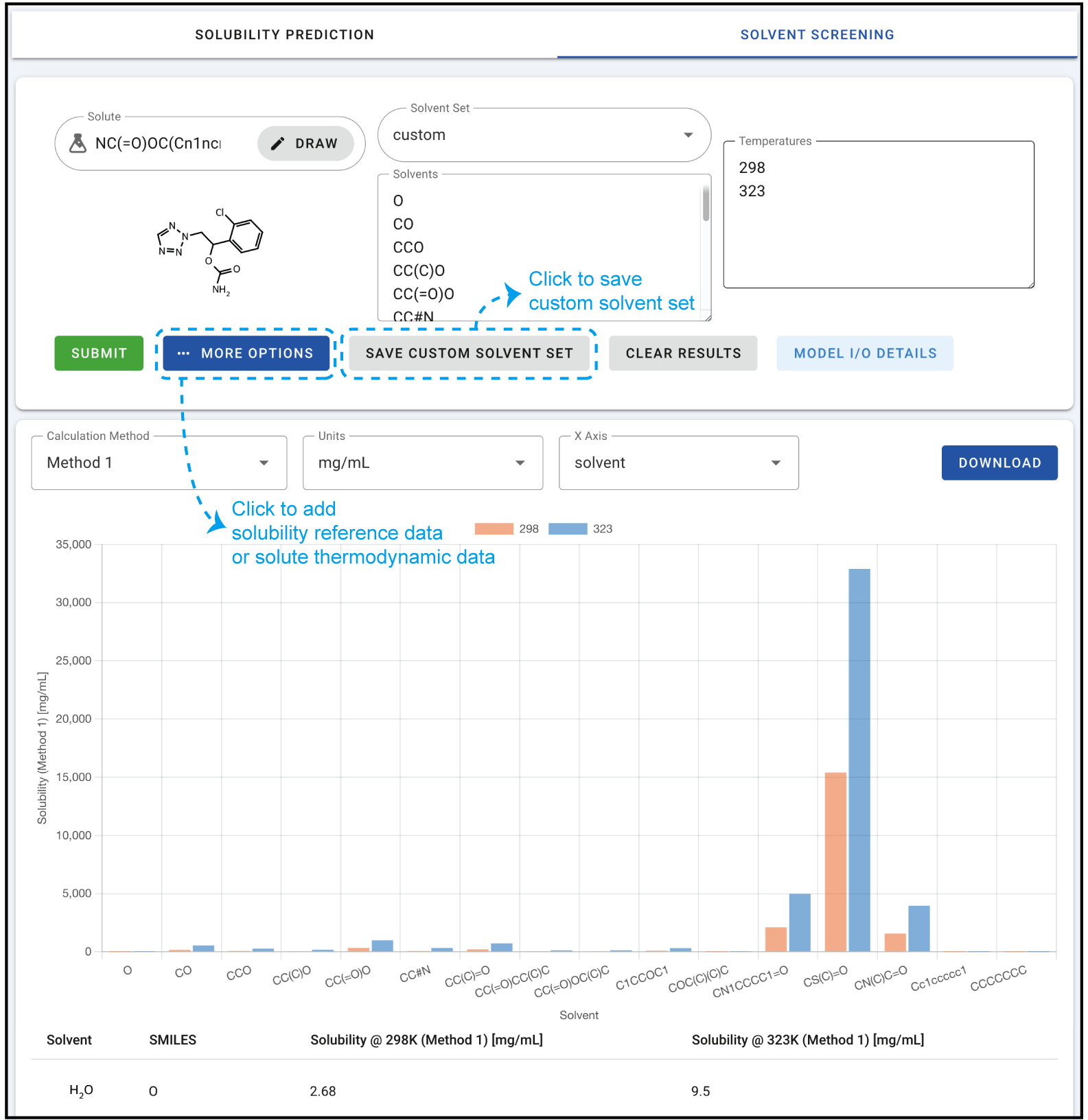}
\caption{Annotated screenshot of solvent screening results in ASKCOS. The SMILES string of the solute is input in the \texttt{Solute} panel. Users can select the solvents from the two predefined sets or create custom solvent sets in the \texttt{Solvent Set} panel. The desired temperatures (in K) at which to compute the solubility and solvation properties are input in the \texttt{Temperatures} panel. The predicted solubility for solute cenobamate defined by the SMILES string \texttt{NC(=O)OC(Cn1ncnn1)c1ccccc1Cl} in a custom set of solvents at 298 K and 323 K is displayed. }\label{fig_solvent_screening}
\end{figure}

A few other optional columns can be viewed in the output table, including the solubility in moles per liter, prediction uncertainty estimates, and any error or warning messages. The uncertainty estimates are the variance of predictions made by an ensemble of models. Each model in the ensemble starts with a different random initialization and is trained on the same data. An ensemble of 10, 12, and 30 models is used for $\Delta G_{solv, 298\, \mathrm{K}}$, $\Delta H_{solv, 298\, \mathrm{K}}$, and $\log\left(S_{aq, 298\, \mathrm{K}}\right)$ respectively. The prediction values in the output table for these quantities are the mean of the ensemble's predictions. 

It is also possible to use the solubility utility to make bulk predictions. Clicking the \texttt{RUN BATCH} button opens a window where a csv or json file containing input data can be uploaded. Clicking \texttt{MODEL I/O DETAILS} shows the format of this file. In the output section, there is also a \texttt{DOWNLOAD} button that allows for downloading all the results as a csv or json file. 

The solvent screening utility, as shown in Figure \ref{fig_solvent_screening}, also allows for bulk solubility predictions of a single solute in a set of solvents and temperatures. Two sets of a variety of solvents are predefined, and users can also make a custom set. The output predictions are shown in both table and chart forms. A bar chart is used if solvent is plotted on the x-axis, whereas a line chart is used if temperature is plotted on the x-axis. 

\section{Details of QM descriptor prediction}\label{results_qm}

ML has been widely used for chemical property predictions, but its performance and generalizability often depends on the availability of large datasets. However, acquiring large-scale datasets can be expensive and time-consuming in scenarios such as \emph{de novo} drug and material design. Driven by the belief that QM descriptors can provide deeper physical insights, recent studies~\citep{guan_regio-selectivity_2021,li_when_2024,stuyver2022quantum} have tried augmenting ML models with QM descriptors. It has been shown that this approach can improve the accuracy and generalizability of the model, especially when trained on smaller datasets. Thus, this method could potentially accelerate the exploration of complex reaction pathways in retrosynthesis and the prioritization of candidates in high-throughput screening.

Since calculating QM descriptors for all the molecules of interest can be very expensive, we include trained models from Li et al.'s work~\citep{li_when_2024} in ASKCOS to predict 37 general QM descriptors on the fly. These models are directed message passing neural network (D-MPNN) models trained on QM descriptors computed using $\omega$B97XD/def2-SVP//GFN2-xTB. The 37 descriptors consist of 13 atom descriptors, 4 bond descriptors, and 20 molecular descriptors. The atom descriptors include NPA charges, Parr functions, NMR shielding constants, and valence orbital occupancies. The bond descriptors consist of bond order, bond length, bonding electrons, and bond natural ionicity. The molecular descriptors encompass energy gaps, ionization potential (IP), electron affinity (EA), and dipole and quadrupole moments. For partial charge prediction, the summation is constrained by the molecule's net charge, nucleophilic/electrophilic Parr functions sum to 1, and no constraints apply to other atom or bond descriptors.

\begin{figure}[h!]
\centering
\includegraphics[width=1.0\textwidth]{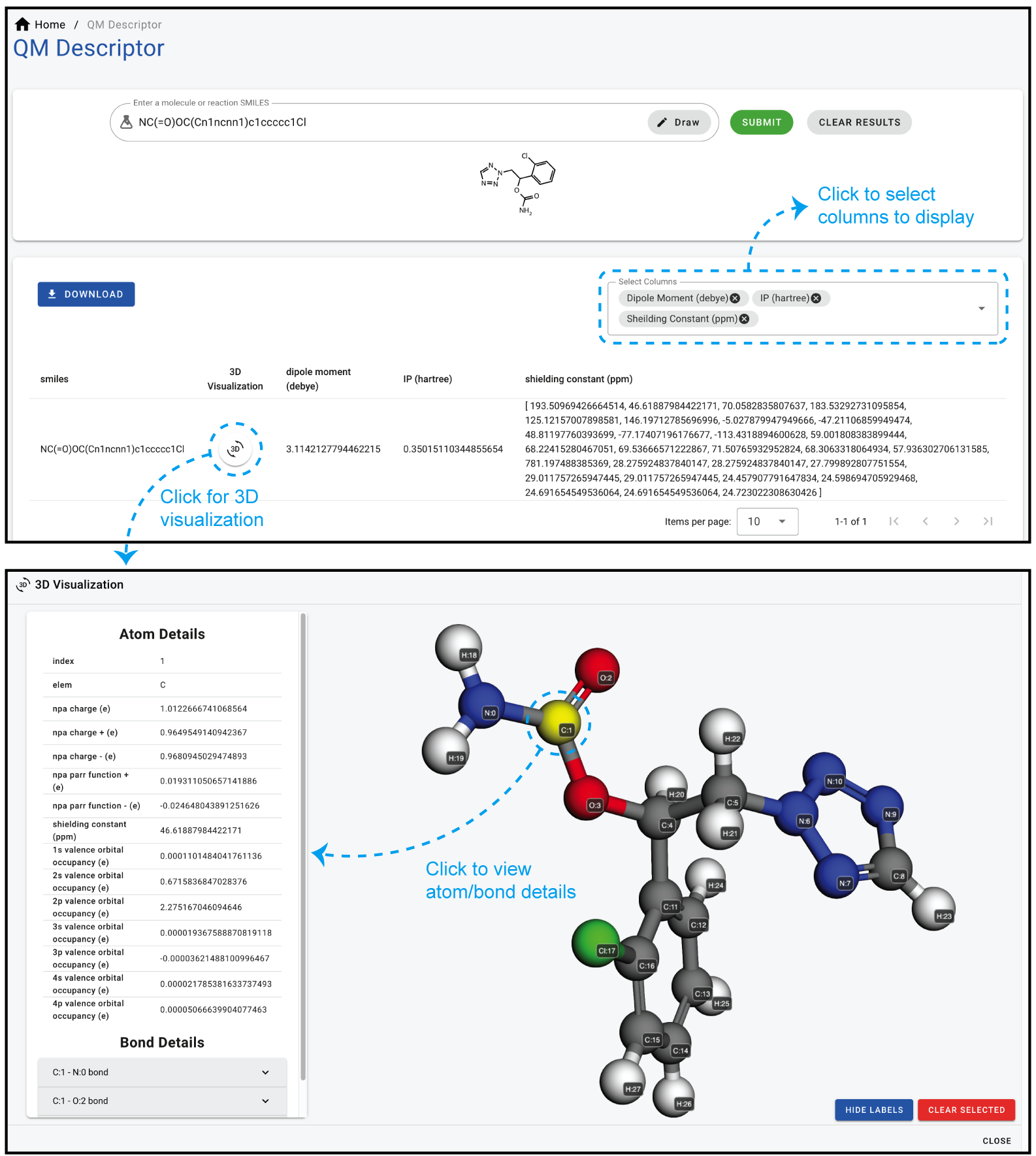}
\caption{Annotated screenshot of QM descriptor prediction results in ASKCOS. The predicted QM descriptors of cenobamate, defined by the SMILES string \texttt{NC(=O)OC(Cn1ncnn1)c1ccccc1Cl}, are displayed (top). The atom and bond descriptors can also be viewed individually with the 3D visualization page (bottom). The selected atom is highlighted in yellow, and the predicted atom properties for the selected atom is displayed on the left panel. This panel also provides predicted values for each bond connected to the selected atom.}\label{fig_qm_descriptors}
\end{figure}

The required input for QM descriptor prediction is either a SMILES string or the use of the drawing functionality. After clicking the \texttt{SUBMIT} button, the predicted QM descriptors will be displayed in a table at the bottom of the screen. By default, only the NPA charges and Parr functions are shown. Other predicted descriptors can be displayed in the output table by selecting them from the \texttt{Select Columns} drop-down menu. Each atom and bond property of a molecule is saved in a list, with the order defined by the RDKit~\citep{RDKit} molecule object. To better visualize these descriptors, users can click the 3D visualization button in the table to view the predicted values individually (Figure \ref{fig_qm_descriptors}) for the selected molecule. In the output section, there is also a \texttt{DOWNLOAD} button that allows you to download all the results as a CSV or JSON file.

\newpage
\section{Technical details of software engineering}\label{method_software}

\subsection{Refactor into a microservice-based architecture}

ASKCOS was originally developed as a pure Python 2 package beginning in 2016 to plan synthetic routes for a robotic flow chemistry platform as part of the DARPA Make-It program, with a full name of ``\textbf{A}utomated \textbf{S}ystem for \textbf{K}nowledge-based \textbf{C}ontinuous \textbf{O}rganic \textbf{S}ynthesis''. Various restful endpoints for modules were defined using the Django package, and the user interface was designed mostly using Javascript and basic Jinja templates. Since then, the number of different predictive models incorporated in ASKCOS has continued to grow. Given its long development history, the monolithic nature of ASKCOS pre-2023 presented a significant challenge in terms of maintainability and extensibility. Managing ever-changing dependencies and their deprecation (e.g., Keras, Tensorflow 1) and synchronizing environments across disparate models developed years-apart by different graduate students and postdocs became unsustainable.

The main deciding factor for a major refactor of ASKCOS was the need to resolve package dependency conflicts. As an extreme example, there is no straightforward way to run a Python program in which part of the code depends on Python 2, and another part depends on Python 3. This type of dependency conflicts became inevitable as we tried to integrate newer and more powerful modules from different contributors into ASKCOS. The de facto and somewhat obvious solution was to turn prediction modules into containerized microservices, which was exactly what has been done in the 2023 refactor, with careful decoupling and re-modularization to retain existing functionalities.

\subsection{Containerized microservices for backend modules} \label{microservices}

Prediction modules as microservices have been the cornerstone of ASKCOS, but it was not until the 2023 refactor that this philosophy was formalized. We enforce modularization and consistency of these microservices by, without loss of generality, having one containerized service per prediction module. No python function call between the modules is possible and any dependency call has to be made via http requests. For example, when the Tree Builder calls any one-step expansion model, it sends an API requests (via the API gateway) to that one-step prediction service. This preserves modularization for maintainability and extensibility in the long run, at little expense as we did not observe noticeable overhead from the conversion of function calls to API calls, especially when the services are hosted on the same machine.

The main challenges when transitioning into a microservice-based architecture was decoupling and modularization, as with code refactor in general. Once the prediction modules have been fully modularized, it is easy to replace function calls with http calls, which are now made by specialized API classes in our implementation. These modules can then be wrapped as services (i.e., as REST endpoints) using packages like FastAPI~\citep{FastAPI} and TorchServe~\citep{TorchServe}, and subsequently containerized with Docker. We refer the reader to the ASKCOS wiki~\citep{ASKCOSwiki} which has fully documented the process of converting python code into containerized services with guiding examples, under the Sections \texttt{Development-Packaging python codes as services} and \texttt{Development-Containerizing backend services}.

\subsection{Centralized API gateway}\label{method_api_gateway}

Calls to backend services are routed via the \emph{API gateway}, which is the central service in ASKCOS for API management. The API gateway is implemented with FastAPI, and consists mainly of \emph{wrappers} for routing API calls to backend services, as well as \emph{utils} which define endpoints for lightweight services such as drawing and authentication within the gateway itself. In order to take full advantage of FastAPI's capability to automatically generate API documentation based on endpoint definitions, these definitions have all been type hinted. In particular, each wrapper defines the expected input schema such as the name of the model used and other hyperparameters in a \texttt{dataclass}, which becomes visible from the auto-generated API documentation as will be illustrated in Section \nameref{method_api_usage}. The schema for the responses returned from the API gateway is defined in similar \texttt{dataclasses}. In this way, we maintain a standard pattern for all API endpoints in the gateway and make it easy to add wrappers for new prediction services: simply copy and paste the definition from an existing wrapper, and then specify the input and response schema.

We have also reworked the asynchronous workflows which are necessary for API calls from the frontend and for long-running tasks. The logic for asynchronous task management has been abstracted away and standardized as an additional endpoint, namely, \texttt{/call-async} within each wrapper. These asynchronous endpoints merely sub-call the corresponding synchronous endpoints in the same wrapper, generally with no other complication. This design keeps the effort for maintenance and extension to the minimum, while still allowing for customization if needed. The async workflows are implemented with the Celery package with RabbitMQ as the broker and Redis for storing results.

\subsection{Frontend}


The ASKCOS frontend is implemented in Vue 3~\citep{Vue} which offers a more flexible, maintainable, and performant architecture. Vue 3, with its composition API and enhanced reactivity system, allows developers to create scalable components and modularize code more effectively. In the context of ASKCOS, Vue 3 serves as the foundation for a dynamic user interface, where complex chemical retrosynthesis data can be seamlessly displayed and interacted with. The integration of Vuetify~\citep{Vuetify}, a robust Material Design component library, ensures a visually cohesive and responsive user interface. Vuetify components are highly customizable, enabling developers to build feature-rich and intuitive user interfaces while adhering to modern web standards. This synergy of Vue 3 and Vuetify helps create an engaging user experience that can handle complex data visualization (in particular, on the IPP canvas) while maintaining accessibility across various devices.

A few other tools are used for testing, application state management and routing. To ensure the reliability and robustness of this complex system, Cypress~\citep{Cypress} is employed for end-to-end testing. Cypress allows for the testing of every aspect of the application from user interaction to API calls, ensuring that ASKCOS functions as intended in real-world scenarios. Pinia~\citep{Pinia} serves as the state management tool to provide a more lightweight, modular, and type-safe approach to managing application state. This helps maintain data consistency across different components, especially in handling chemical datasets and user interactions. Furthermore, Vue Router~\citep{VueRouter} integrates seamlessly with the Vue 3 framework, managing complex navigational patterns, enabling smooth transitions between different sections of the application without reloading the page. These technologies together form the backbone of the ASKCOS web platform, ensuring high performance, reliability, and ease of use for chemists and researchers accessing the platform.

\subsection{Application monitoring and logging}


Several utilities have been designed to help monitor application status while ASKCOS is running. From the \texttt{Server Status} page of the ASKCOS user interface, users can see a status summary of the services for the celery workers, the database, and prediction modules. For the celery services, the status summary shows the numbers of their pending tasks as well as of available and busy workers. For the data collections (e.g., reactions, templates, and buyable building blocks) which are stored in the Mongo database, a description of each collection and the number of documents imported during database seeding are displayed. For backend services, metadata including model names and descriptions are populated from the config file used to deploy and start ASKCOS. The service status (online vs. offline) is checked with API calls and updated upon refresh.

The logs of the API calls made from the frontend are accessible from the \texttt{Logs} page, which can be particularly helpful for debugging. The storage of such logs is made possible by Pinia which allows a state to be shared across components and pages. When a user performs an action on any ASKCOS page, if requests to the backend APIs are made, Pinia can relay information about these requests to the \texttt{Logs} page. The logs can be cleared by refreshing any page in ASKCOS.

For superusers, another useful API endpoint is \texttt{/api-logging/get}, which counts the number of API calls to each endpoint and aggregates by date. It provides succinct statistics of the most used features in ASKCOS to help superusers optimize the deployment, e.g., by increasing the number of celery workers for more frequently used modules.

\section{Advanced features}\label{advanced_features}

\subsection{User-friendly and customizable deployment}\label{method_deployment}

ASKCOS can be deployed from scratch with the following five commands:
\begin{lstlisting}[language=bash]
$ mkdir ASKCOSv2
$ cd ASKCOSv2
$ git clone git@gitlab.com:mlpds_mit/askcosv2/askcos2_core.git
$ cd askcos2_core
$ make deploy
\end{lstlisting}
We refer the reader to the ASKCOS wiki~\citep{ASKCOSwiki} for the full instructions, including hardware and software requirements. The last command, \texttt{make deploy}, is the main deployment command which does the following in sequence:

\begin{enumerate}
    \item cloning all other repositories under \texttt{ASKCOSv2/} based on the central config file in \texttt{askcos2\_core}, while downloading data and model checkpoints if needed;
    \item generating deployment scripts for building Docker images, starting services and stopping services based on the central config file;
    \item building Docker images for all services using the generated script;
    \item downloading database data and seeding into the Mongo database;
    \item starting all services using using the generated script.
\end{enumerate}

A single centralized config file specifies all the required configurations. It is easy to \emph{turn off} unneeded modules for users who are more resource-constrained or have interest in only a subset of modules. Partial deployment has been documented in the ASKCOS wiki, and we include several sample config files for typical use cases (e.g., retro-only and/or backend-only).

\subsection{Model retraining and integration}\label{method_retraining}


A few models for one-step retrosynthesis and reaction outcome prediction have been designed for easy retraining and integration with new datasets. In particular, the retraining pipeline of the template relevance model has been streamlined. Only a list of reaction IDs and reaction SMILES are needed, and the user has the choice of either providing own train/validation/test splits, or letting the retraining engine handle the splitting. Thereafter, automated retraining and testing can be performed with the provided script (e.g., $\texttt{benchmark*.sh}$) after specifying the paths to the reaction files.

Integration of newly trained models into ASKCOS requires \emph{model archiving}, followed by updating the existing deployment config, and possibly seeding additional data into the database. Archiving models into a servable \texttt{.mar} file is done using \texttt{torch-model-archiver}, while the rest of the integration pipeline is mostly bookkeeping to place model files into the correct location and make the API gateway aware of the existence of the new model. The full process of retraining and integration has been documented in the ASKCOS wiki under the Section \texttt{Deployment-Model retraining and integration}.

\subsection{User customization}\label{method_customization}

ASKCOS provides options to customize many aspects of the application. The design of the deployment pipeline discussed in Section \nameref{method_deployment} allows users to easily customize their deployment. The complete application including the frontend and the backend is deployed by default, but if ASKCOS is being integrated with other systems where only API calls are being made and the user interface is not needed, backend-only deployment may be more suitable. Similarly, while some datasets are copied over and loaded into ASKCOS at first deployment, additional datasets can be added, or others removed, at a later date by simply invoking subroutines in the \texttt{deploy.sh} script in the \texttt{askcos2\_core} directory to modify the database. Additionally, as eluded to previously in Section \nameref{method_deployment}, superusers can chose to disable certain modules depending on usage and computational resources available.

Other aspects of the environment can be customized too, and most customizable environment (env) variables are found in the \texttt{.env.example} file. The user interface can be configured to use a different logo, for example, a company logo, and have a different welcome message which may include the company name. This is very important as ASKCOS is publicly accessible, and as such, there are no protections regarding intellectual property (IP). If users want to use ASKCOS on their proprietary, sensitive data, ASKCOS should be deployed behind their company firewalls. Customizing the user interface quickly informs users they are on their internal deployment and can safely and confidently enter potentially sensitive compound data. The superuser can also change the support email addresses (by modifying the \texttt{VITE\_CONTACT\_EMAIL} and \texttt{VITE\_SUPPORT\_EMAILS} env variables) from the default MIT support groups to internal support addresses to ensure sensitive data is not shared outside of the company.
 
Many companies have policies that block applications requesting data outside of their firewalls to safeguard intellectual property. ASKCOS has only one external lookup, calling on the NIH name resolving service~\citep{NIHNameResolver} to convert a compound name to its corresponding SMILES string. This service can be quickly disabled on each page by clicking the icon beside the target input area. It can permanently be disabled on deployment by setting the env variable \texttt{VITE\_ENABLE\_SMILES\_RESOLVER} to False.

ASKCOS allows users to ban chemicals and reactions. These lists are specific to that user account and not global. This allows users to ban compounds or reactions that are patented, that may be toxic or harmful, or that are otherwise considered undesirable. Chemicals or reactions can be added to the ban list directly from the IPP. Banned chemicals or reactions will not appear in the predicted pathways afterwards. Users can view, add, or delete items in their ban lists on \texttt{My Banlist} page accessible from the left hand sidebar.

The \texttt{Buyables} page and backend data structures have also been enhanced to allow superusers to add important metadata such as lead time and availability to their own building blocks. These custom metadata are easily viewable during interactive planning (via the \texttt{SEARCH BUYABLES} button in the chemical node detail panel) to assist chemists' decision making. Links to the procurement website can also be added to enable easier and quicker ordering. Superusers can add buyables individually or submit a list of buyables in a file for bulk upload.

\subsection{API usage}\label{method_api_usage}

As the frontend and the backend are fully separated in ASKCOS, it is possible to send http requests directly to the API gateway, e.g., from the command line or using some client library from any programming language. Using the APIs this way may be particularly useful for integrating some of ASKCOS' predictive modules into other workflows and for batch queries, for example, by checking whether routes with buyable building blocks can be found for a given list of (possibly many) molecules. The auto-generated API documentation mentioned in Section \nameref{method_api_gateway}, which is guaranteed to be up-to-date by design, can be accessed from the browser as an interactive web page for any locally deployed ASKCOS instance. The endpoints have been organized by groups on the documentation page, and many of these endpoints have been pre-filled with working sample queries which can be easily executed with a few clicks. The interactive nature of the documentation makes it easy for experimenting with various APIs, as well as understanding their schema and performance.

We refer the readers to the ASKCOS wiki~\citep{ASKCOSwiki} for detailed and more visual illustration of the API usage, under the Section \texttt{Advanced Usage-Using the APIs directly} where python script examples for API queries have also been provided.

\section{Illustration of applying ASKCOS to FDA-approved small molecule drugs from 2019 to 2023}\label{results_fda}

We perform a case study on FDA-approved New Chemical Entities from 2019 to 2023 (New Drug Applications only) to demonstrate how a chemist may use ASKCOS to plan routes for new targets. The compilation of raw data is retrieved from the FDA website~\citep{FDAApproval}. The raw chemical names of drug components are converted to SMILES using the NIH name resolving API~\citep{NIHNameResolver}. Unresolvable names and components found in the buyable database of ASKCOS are dropped, and the remaining list of targets are deduplicated based on the canonical SMILES. The final list contains 75 unique targets in total. The raw and processed data files, as well as the scripts for data processing and for sending tree building queries are provided with instructions under \href{https://gitlab.com/mlpds\_mit/askcosv2/askcos2\_core/-/tree/main/examples}{https://gitlab.com/mlpds\_mit/askcosv2/askcos2\_core/-/tree/main/examples}.

\subsection{Initial automated tree building with typical settings}

\begin{figure}[h!]
    \captionsetup[subfigure]{labelformat=empty}
    \begin{subfigure}[t]{1.0\textwidth}
        \includegraphics[scale=0.725]{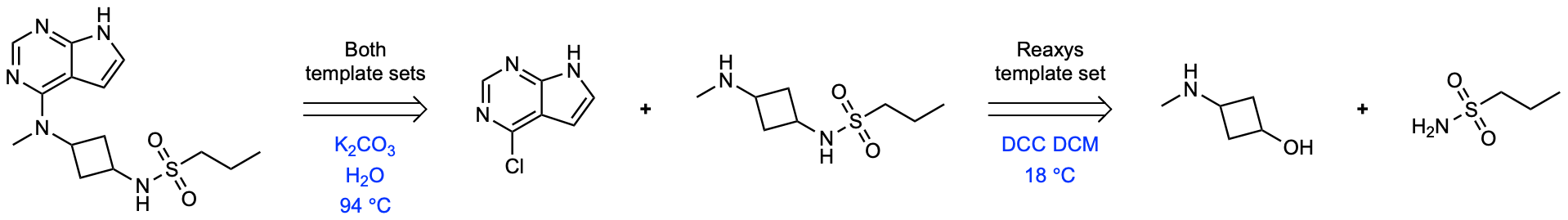}
        \caption{Abrocitinib: \\ \small CCCS(=O)(=O)NC1CC(N(C)c2ncnc3[nH]ccc23)C1}
    \end{subfigure}
    \hfill
    \vspace{1cm}
    \begin{subfigure}[t]{1.0\textwidth}
        \includegraphics[scale=0.725]{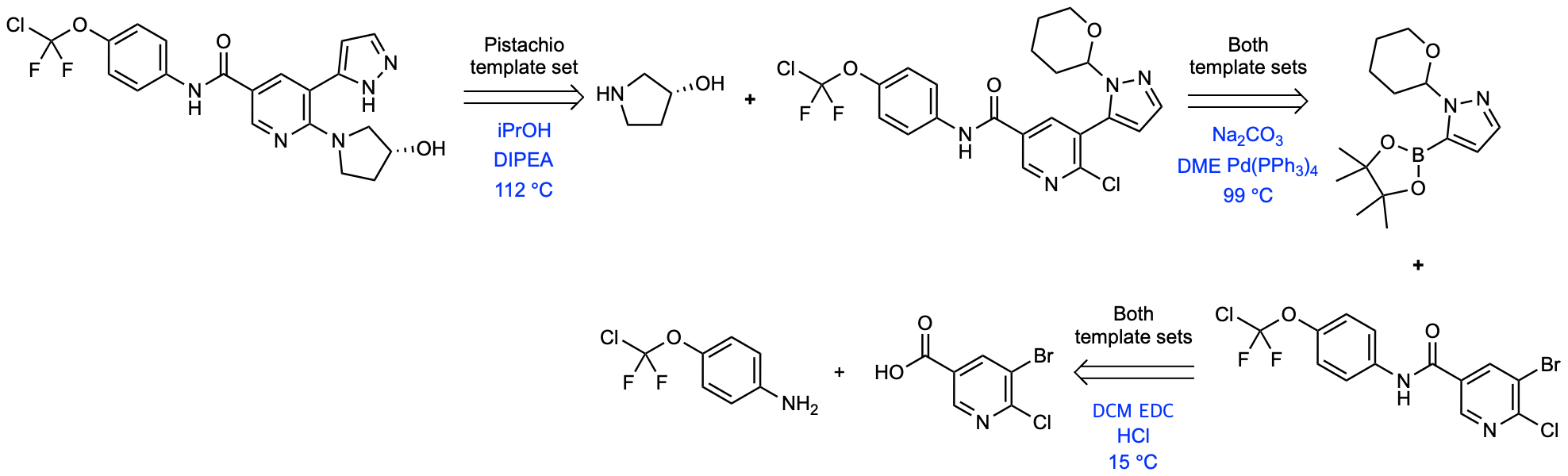}
        \caption{Asciminib: \\ \small O=C(Nc1ccc(OC(F)(F)Cl)cc1)c1cnc(N2CC[C@@H](O)C2)c(-c2ccn[nH]2)c1}
    \end{subfigure}
    \hfill
    \vspace{1cm}\begin{subfigure}[t]{1.0\textwidth}
        \includegraphics[scale=0.725]{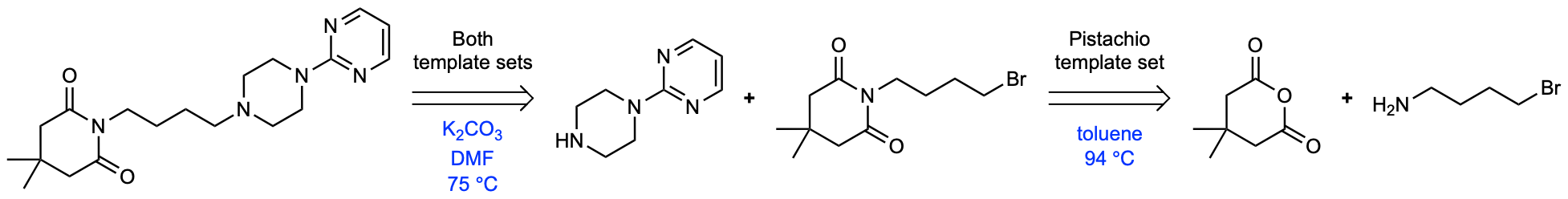}
        \caption{Gepirone: \\ \small CC1(C)CC(=O)N(CCCCN2CCN(c3ncccn3)CC2)C(=O)C1}
    \end{subfigure}
    \hfill
    \vspace{1cm}
    \begin{subfigure}[t]{1.0\textwidth}
        \includegraphics[scale=0.725]{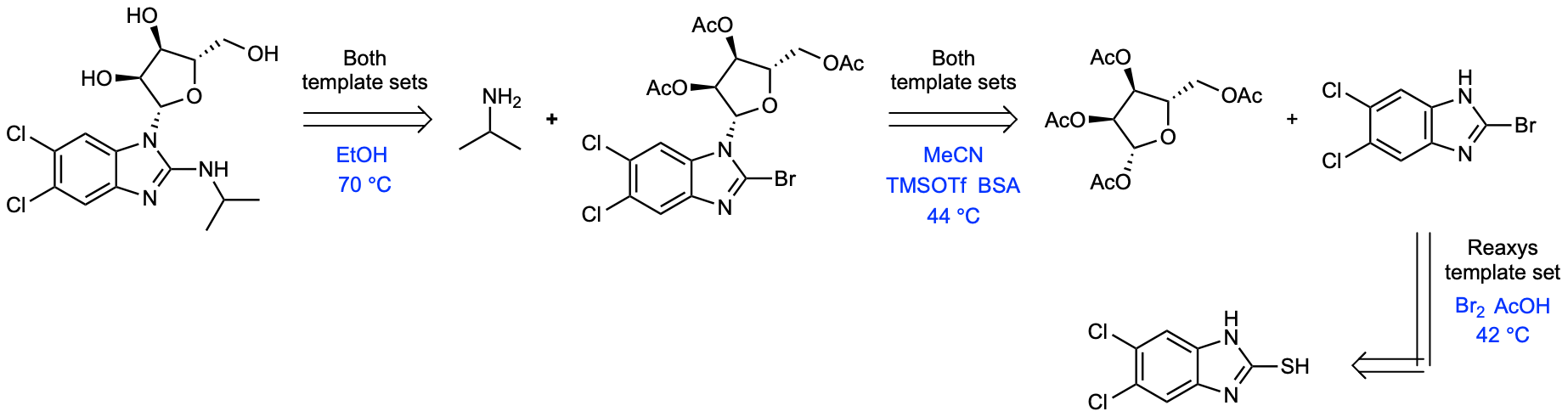}
        \caption{Maribavir: \\ \small CC(C)Nc1nc2cc(Cl)c(Cl)cc2n1[C@H]1O[C@@H](CO)[C@H](O)[C@@H]1O}
    \end{subfigure}
    \hfill
    \caption{Shortest retrosynthetic routes suggested by ASKCOS for FDA-approved small molecule drug components, part I. Top-1 recommendations by the condition recommender (V1) are shown in blue below each arrow. Ions are written in salt form. Abbreviations are defined in Table \ref{table:abbreviations}.}
    \label{fig:fda_study_1}
\end{figure}

\begin{figure}[h!]
    \captionsetup[subfigure]{labelformat=empty}
    \begin{subfigure}[t]{1.0\textwidth}
        \includegraphics[scale=0.725]{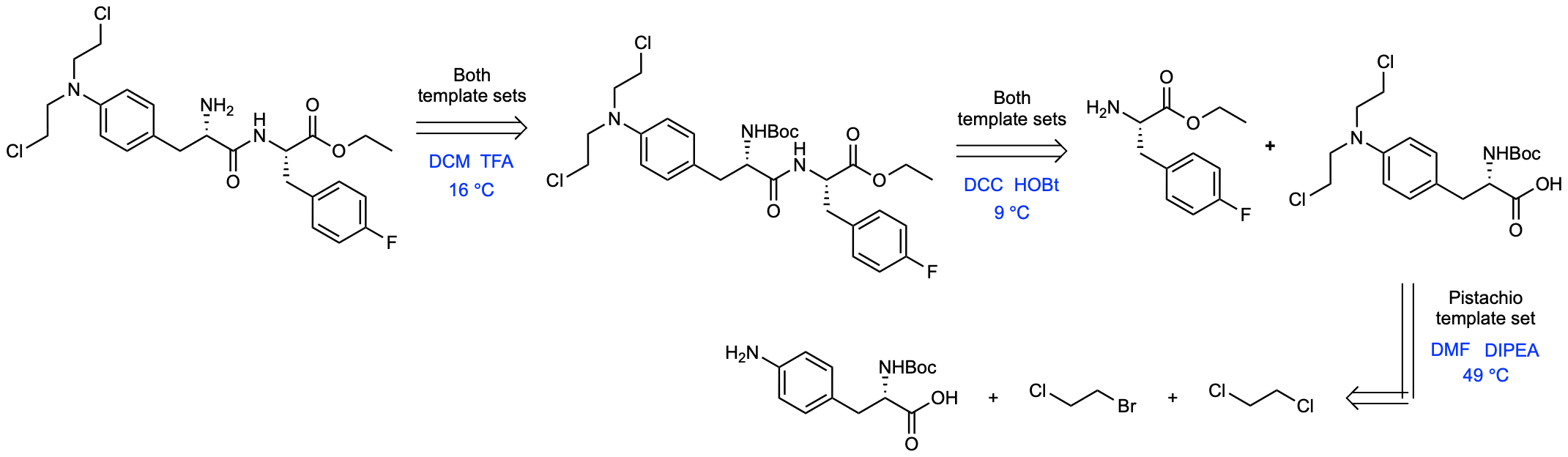}
        \caption{Melphalan flufenamide: \\ \small CCOC(=O)[C@H](Cc1ccc(F)cc1)NC(=O)[C@@H](N)Cc1ccc(N(CCCl)CCCl)cc1}
    \end{subfigure}
    \hfill
    \vspace{1cm}
    \begin{subfigure}[t]{1.0\textwidth}
        \includegraphics[scale=0.725]{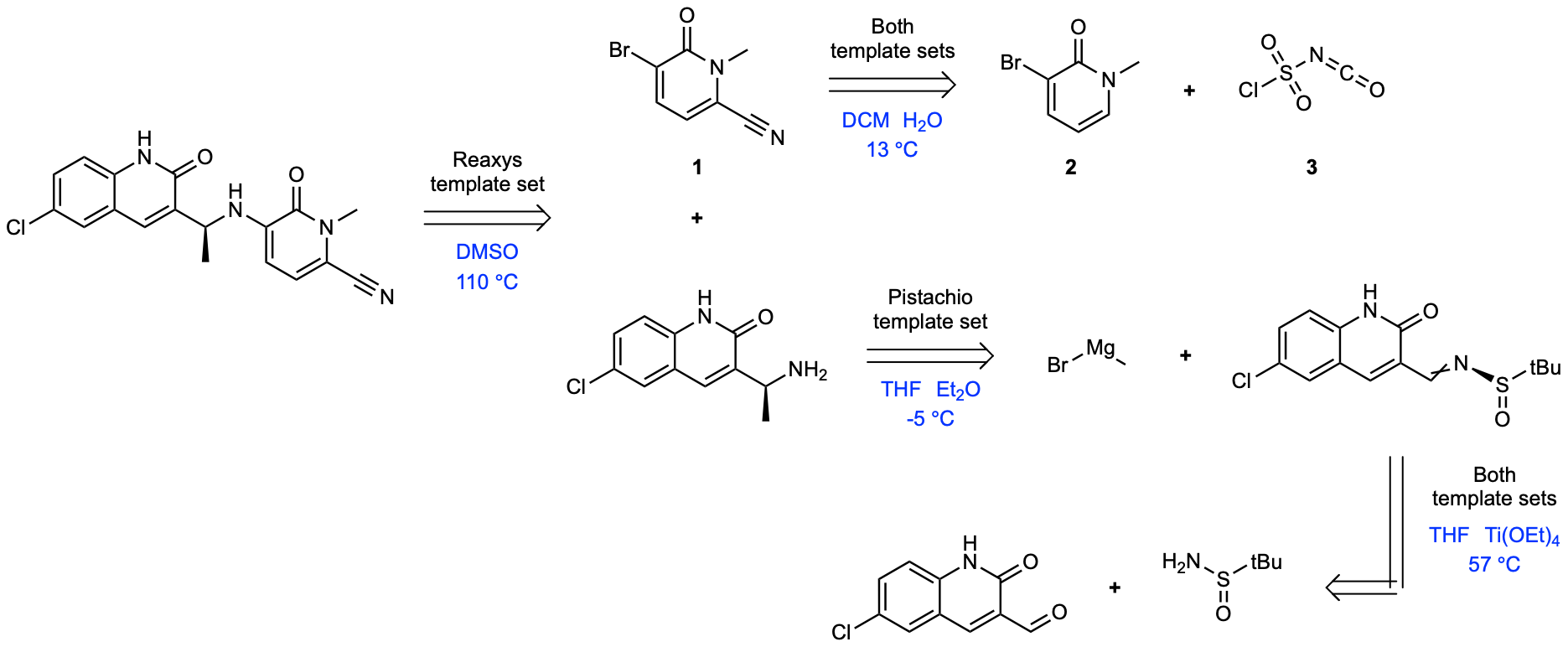}
        \caption{Olutasidenib: \\ \small C[C@H](Nc1ccc(C\#N)n(C)c1=O)c1cc2cc(Cl)ccc2[nH]c1=O}
    \end{subfigure}
    \hfill
    \vspace{1cm}
    \begin{subfigure}[t]{1.0\textwidth}
        \includegraphics[scale=0.725]{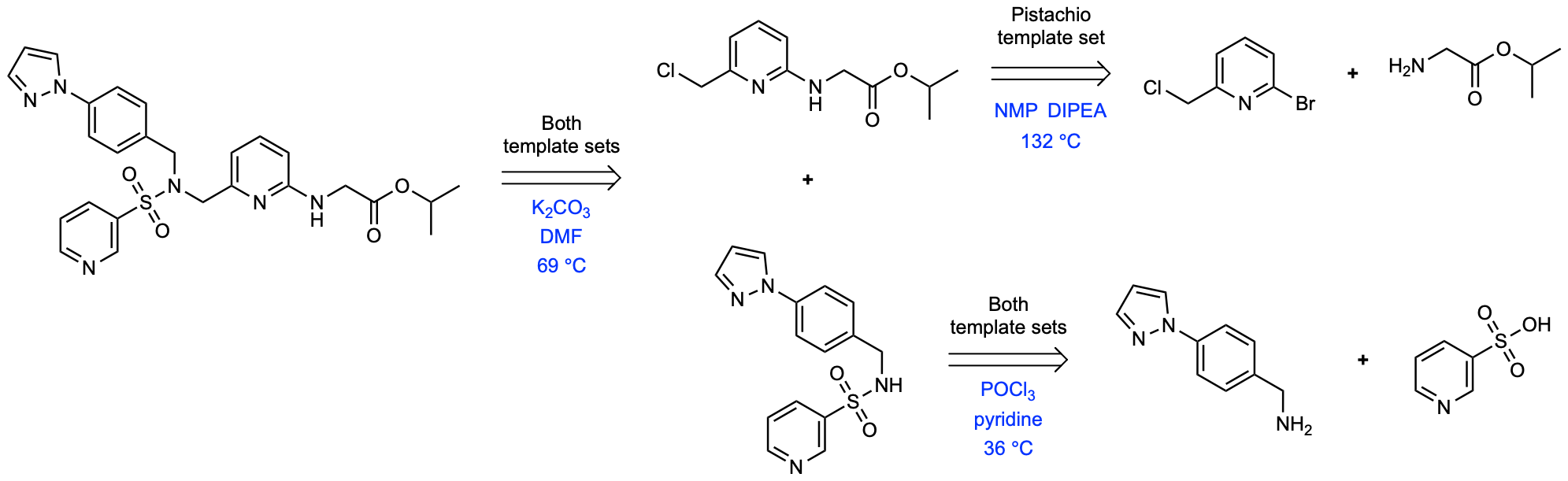}
        \caption{Omidenepag isopropyl: \\ \small CC(C)OC(=O)CNc1cccc(CN(Cc2ccc(-n3cccn3)cc2)S(=O)(=O)c2cccnc2)n1}
    \end{subfigure}
    \hfill
    \caption{Shortest retrosynthetic routes suggested by ASKCOS for FDA-approved small molecule drug components, part II. Top-1 recommendations by the condition recommender (V1) are shown in blue below each arrow. Ions are written in salt form. Abbreviations are defined in Table \ref{table:abbreviations}.}
    \label{fig:fda_study_2}
\end{figure}

\begin{figure}[h!]
    \captionsetup[subfigure]{labelformat=empty}
    \begin{subfigure}[t]{1.0\textwidth}
        \includegraphics[scale=0.725]{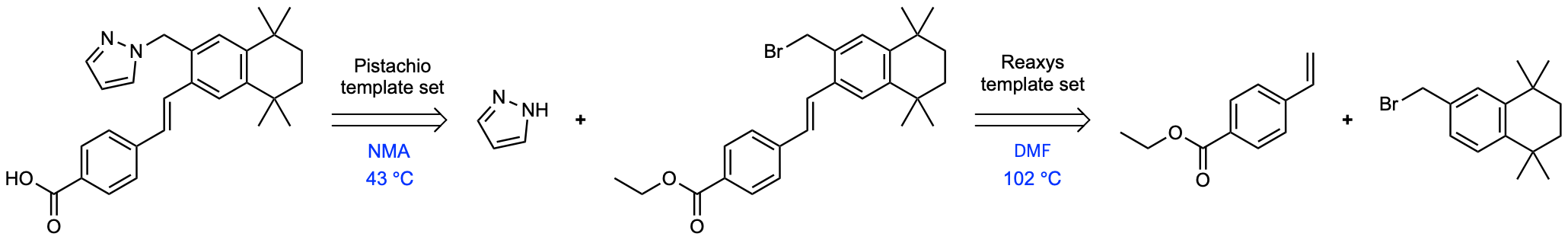}
        \caption{Palovarotene: \\ \small CC1(C)CCC(C)(C)c2cc(Cn3cccn3)c(/C=C/c3ccc(C(=O)O)cc3)cc21}
    \end{subfigure}
    \hfill
    \vspace{1cm}
    \begin{subfigure}[t]{1.0\textwidth}
        \includegraphics[scale=0.725]{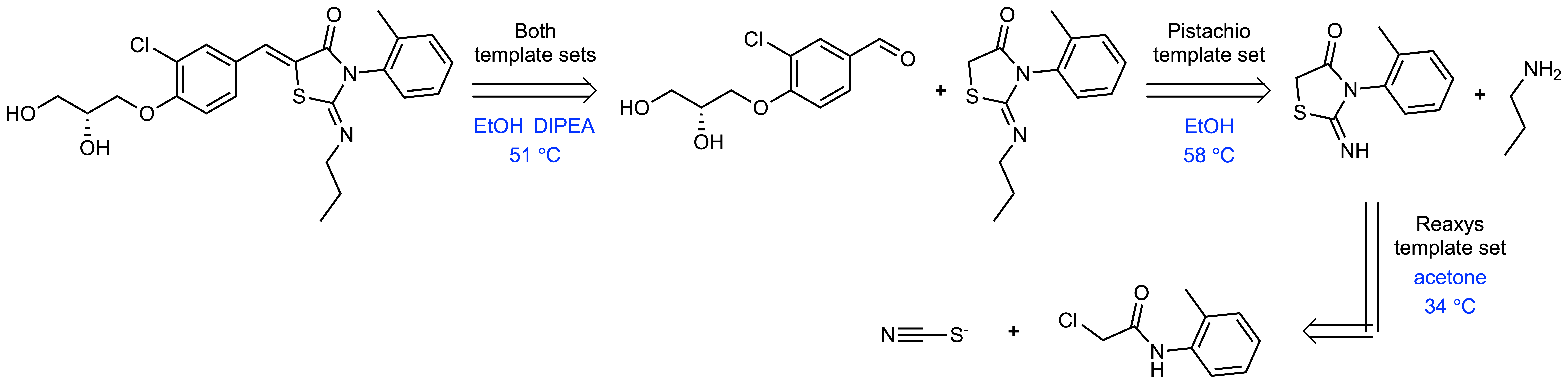}
        \caption{Ponesimod: \\ \small CCCN=C1S/C(=C$\backslash$c2ccc(OC[C@H](O)CO)c(Cl)c2)C(=O)N1c1ccccc1C}
    \end{subfigure}
    \hfill
    \caption{Shortest retrosynthetic routes suggested by ASKCOS for FDA-approved small molecule drug components, part III. Top-1 recommendations by the condition recommender (V1) are shown in blue below each arrow. Abbreviations are defined in Table \ref{table:abbreviations}.}
    \label{fig:fda_study_3}
\end{figure}

\begin{figure}[h!]
    \captionsetup[subfigure]{labelformat=empty}
    \begin{subfigure}[t]{1.0\textwidth}
        \includegraphics[scale=0.725]{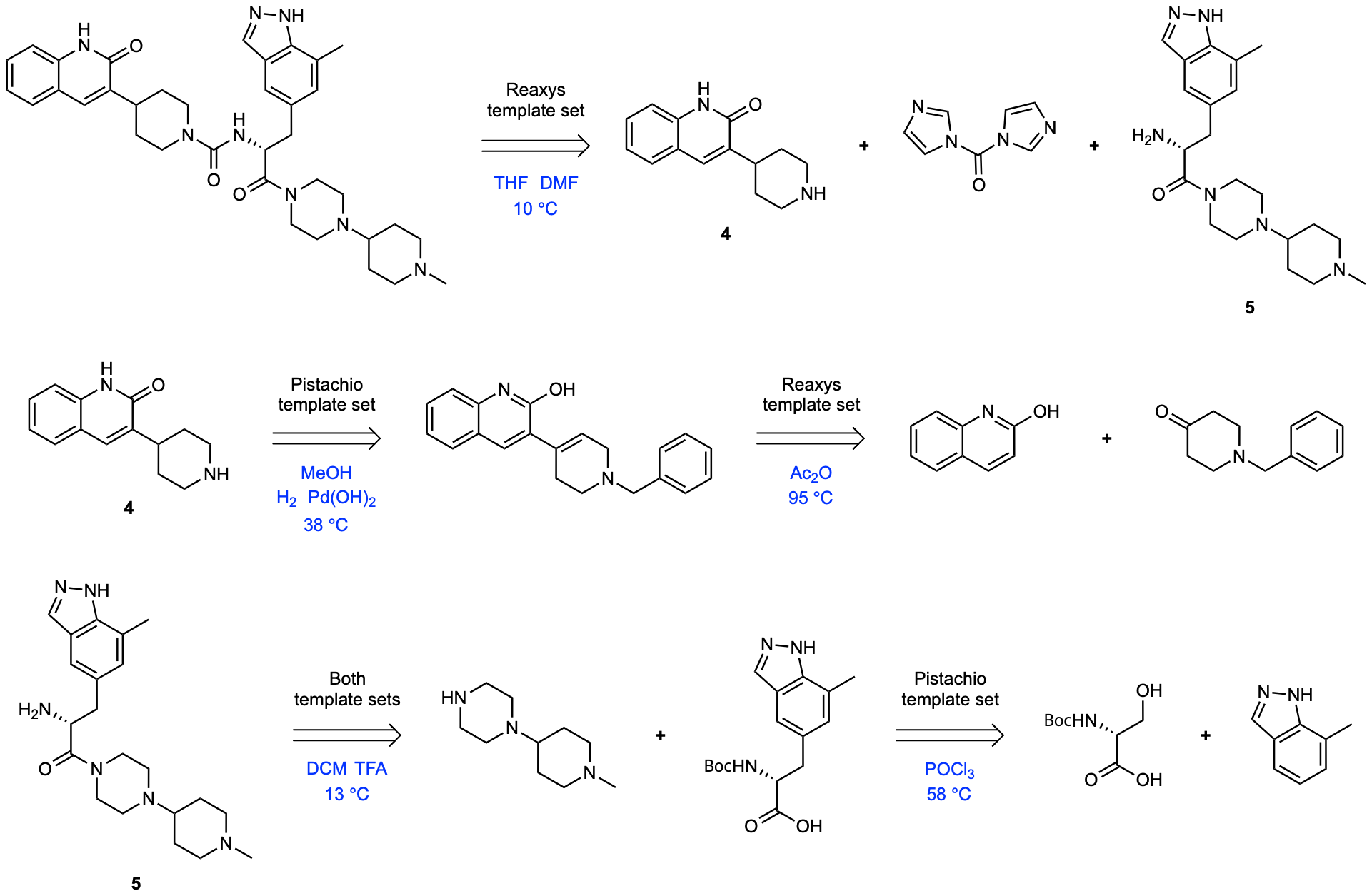}
        \caption{Zavegepant: \\ \small Cc1cc(C[C@@H](NC(=O)N2CCC(c3cc4ccccc4[nH]c3=O)CC2)C(=O)N2CCN \\ (C3CCN(C)CC3)CC2)cc2cn[nH]c12}
    \end{subfigure}
    \hfill
    \vspace{1cm}
    \caption{Shortest retrosynthetic routes suggested by ASKCOS for FDA-approved small molecule drug components, part IV. Top-1 recommendations by the condition recommender (V1) are shown in blue below each arrow. Ions are written in salt form. Abbreviations are defined in Table \ref{table:abbreviations}.}
    \label{fig:fda_study_4}
\end{figure}

As a baseline, we first query the MCTS endpoint with typical search settings. Specifically, Pistachio-trained and Reaxys-trained template relevance models are used simultaneously for one-step retrosynthetic expansion, both with a maximum of 1,000 templates and a maximum cumulative template probability of 0.999 per expansion step. The minimum threshold for the plausibility from the binary fast filter is set to 0.001. For the tree search, a maximum branching factor of 25 and a maximum search depth of 6 are used. The number of chemical nodes to be explored is capped at 5,000 and the maximum price for buyable building blocks is set at \$100/g, with no limit on expansion time. This baseline run takes about an hour to finish for all 75 targets on a typical desktop (with Intel i7-12700 CPU, 32 GB of RAM, and no GPU), corresponding to roughly one minute of wall time per molecule. After hypothetical retrosynthetic routes terminating in buyable building blocks are found for a target, routes are algorithmically sorted by length, by average plausibility based on the binary filter, and then by average template score. We include the rank-1 (by this definition) route for each of 10 sample targets in Figures \ref{fig:fda_study_1}, \ref{fig:fda_study_2}, \ref{fig:fda_study_3}, and \ref{fig:fda_study_4}. Note that these pathways are \emph{exactly} as returned by the Tree Builder and have not been postprocessed (i.e., filtered or rescored) by any other modules in ASKCOS. We show the top-1 proposed conditions for each step using the V1 condition recommendation model. Users can change the settings to view lower-ranking predictions by the V1 model and quantitative predictions by the V2 model.

\clearpage

\begin{table}[h!]
\caption{List of abbreviations used in Figures \ref{fig:fda_study_1}, \ref{fig:fda_study_2}, \ref{fig:fda_study_3}, and \ref{fig:fda_study_4}}\label{table:abbreviations}
\begin{tabular*}{\textwidth}{@{\extracolsep\fill}ll}
\toprule
Abbreviation & Full name \\
\midrule
BSA & benzenesulfonamide \\
DCC & 1,3-dicyclohexylcarbodiimide \\
DCE & 1,2-dichloroethane \\
DCM & dichloromethane \\
DEAD & 	diethylazodicarboxylate \\
DIPEA & \emph{N,N}-diisopropylethylamine \\
DMAP & \emph{N,N}-dimethyl-4-aminopyridine \\
DME & ethylene glycol dimethyl ether \\
DMF & \emph{N,N}-dimethylformamide \\
DMSO & dimethyl sulfoxide \\
DPP & diphenyl phosphate \\
EDC & 1-ethyl-3-(3-dimethylaminopropyl)carbodiimide \\
HATU & hexafluorophosphate azabenzotriazole tetramethyl uronium\\
HMPA & hexamethylphosphoric triamide \\
HOBt & hydroxybenzotriazole \\
LDA & lithium diisopropylamine \\
NMA & \emph{N}-methylacetamide \\
NMP & \emph{N}-methyl-2-pyrrolidone \\
TFA & trifluoroacetic acid \\
THF & tetrahydrofuran \\
TMSOTf & trimethylsilyl trifluoromethanesulfonate \\
\botrule
\end{tabular*}
\end{table}

\subsection{Step-wise verification and further analysis \emph{within} ASKCOS} \label{verification_and_analysis}

Successfully proposing routes for a target means that the recursive retrosynthetic tree search was able to identify hypothetical pathways that terminate in buyable starting materials, but does not necessarily mean that those hypothetical pathways are chemically plausible. Here, we discuss how ASKCOS facilitates cross-referencing with literature when needed, and how we can analyze suggestions more thoroughly.

At least a few steps among the routes in Figures \ref{fig:fda_study_1}, \ref{fig:fda_study_2}, \ref{fig:fda_study_3}, and \ref{fig:fda_study_4} are counter-intuitive or seemingly implausible. For example, in the route for olutasidenib (Figure \ref{fig:fda_study_2}), cyanopyridone \textbf{1} is prepared from the corresponding less substituted pyridone \textbf{2} in one step using chlorosulfonyl isocyanate (CSI) \textbf{3}. Since any step proposed by a template relevance model can be traced back to the associated template(s) and literature precedents, we can easily cross-reference the origin of the suggestion in the graphical interface by clicking the reaction node, clicking the template, and clicking the reference link sequentially as described in the Results Section. In this case, the precedent substrates for this cyanation~\citep{anderson_pyrrole_1985,elliott_intramolecular_2007,koovits_conformationally_2016} are all pyrroles. Visualization of the template shows that it captures only the requirement of an aromatic N-heterocycle without accounting for the ring size or other substituent effects. While this is typical for algorithmically-extracted templates, these types of errors can be identified and manually corrected. We exported the route into the Interactive Path Planner (IPP), removed this problematic step, and re-expanded manually with the Reaxys template set to find an alternative using trimethylsilyl cyanide. Checking the new template and some associated references~\citep{boogaard_ring_1994,ornstein_4-tetrazolylalkylpiperidine-2-carboxylic_1991,price_orally_2018} provides stronger evidence that cyanation with trimethylsilyl cyanide might afford \textbf{1} from \textbf{2}. It is worth noting that we can always manually replace this final step after exporting the route into the IPP. In a more general setting, if we are unsatisfied with some particular intermediate step(s), we can either re-expand the whole sub-tree, or pick another route with the unpromising step appearing more upstream in the synthesis so that we can ``fail early''.


Similar to the reaction steps, the validity and/or optimality of the top proposed reaction condition is not guaranteed. In particular, for the last step of the route for abrocitinib (Figure \ref{fig:fda_study_1}), DCC is suggested as part of the top conditions. As a reagent generally for amide coupling, it does not typically directly activate alcohols (into better leaving groups for nucleophilic attack). The V2 condition recommendation model instead proposes Mitsunobu conditions for this step (Table \ref{table:abrocitinib_conditions}) with sufficient literature precedent to suggest plausibility~\citep{henry1989mitsunobu}. For the last step of the route for asciminib (Figure \ref{fig:fda_study_1}), a mild, base-free set of conditions is initially proposed for the proposed amide coupling with EDC. These conditions may be improved by addition of a base, and as in the rank 2 and 3 predictions by the V1 condition recommender, DMAP and HOBt (Table \ref{table:asciminib_conditions}), which are commonly used in combination with EDC. For the second step of the route for maribavir (Figure \ref{fig:fda_study_1}), bis(trimethylsilyl)acetamide in the rank 3 condition by V1 may be a better base to use than benzenesulfonamide (BSA) in the rank 1 \& 2 recommended conditions by V1 (Table \ref{table:maribavir_conditions}). The final retrosynthetic step for zavegepant (Figure \ref{fig:fda_study_4}) would also warrant further investigation of potential protecting group chemistry and/or specialized conditions to achieve the desired site selectivity. Modifications of reactions conditions based on user expertise are common; in practice, we interpret recommended conditions as starting points for empirical screening and further optimization, rather than a claim that the reaction will/must proceed exactly as proposed.

\begin{table}[h!]
\caption{Recommended conditions for the last step in the proposed retrosynthesis of 
abrocitinib. The top 3 recommendations of the V2 condition recommender using fingerprint (fp) and graph representations of molecules are shown. }\label{table:abrocitinib_conditions}
\begin{tabular*}{\textwidth}{@{\extracolsep{\fill}}p{1.42cm}p{0.92cm}p{9.8cm}}
\toprule
\multicolumn{3}{c}{
\setlength{\fboxrule}{0pt}\fbox
{\includegraphics[width=0.6\textwidth]{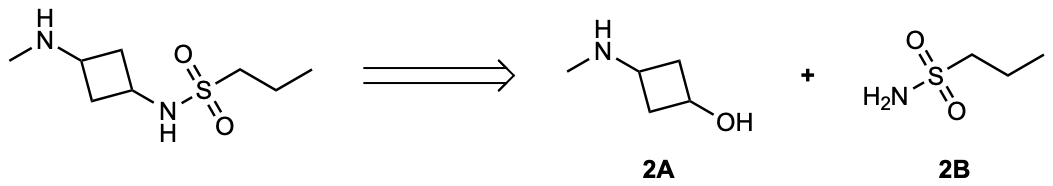}}} \\
\midrule
 \multirow{3}{*}{V1} & Rank 1 &  DCM, DCC at 18°C\\
 &Rank 2 & DCM, DCC, DMAP  at 17°C\\
 &Rank 3 &  THF, DCC at 24°C\\
 \midrule
 \multirow{5}{*}{V2 (fp)}  & Rank 1 & THF (110.71 equiv.), PPh\textsubscript{3} (0.57 equiv.), DEAD (1.52 equiv.)  at 22°C with \\
 & & reactants \textbf{2A} (1 equiv.), \textbf{2B} (1.06 equiv.)\\
 &Rank 2 & H\textsubscript{2}O (38.37), THF (70.34 equiv.), PPh\textsubscript{3} (0.71 equiv.), DEAD (1.39 equiv.)  \\
 & & at 24°C 
 with reactants \textbf{2A} (1 equiv.), \textbf{2B} (1.11 equiv.)\\
 &Rank 3 & THF (68.31 equiv.) at 36°C with reactants \textbf{2A} (1 equiv.), \textbf{2B} (1.34 equiv.) \\
\midrule
\multirow{3}{*}{V2 (graph)} & Rank 1 & pyridine (8.61 equiv.) at 8°C with reactants \textbf{2A} (1 equiv.), \textbf{2B} (1.15 equiv.) \\
 &Rank 2 & MeCN (75.16 equiv.) at 19°C with reactants \textbf{2A} (1 equiv.), \textbf{2B} (1.22 equiv.) \\
 &Rank 3 & DCM (12.96 equiv.) at 6°C with reactants \textbf{2A} (1 equiv.), \textbf{2B} (1.22 equiv.) \\

\botrule
\end{tabular*}
\end{table}

\begin{table}[h!]
\caption{Recommended conditions for the last step in the proposed retrosynthesis of 
asciminib. The top 3 recommendations of the V2 condition recommender using fingerprint (fp) and graph representations of molecules are shown.  }\label{table:asciminib_conditions}
\begin{tabular*}{\textwidth}{@{\extracolsep{\fill}}p{1.42cm}p{0.92cm}p{9.8cm}}
\toprule
\multicolumn{3}{c}{
\setlength{\fboxrule}{0pt}\fbox{
\includegraphics[width=0.6\textwidth]{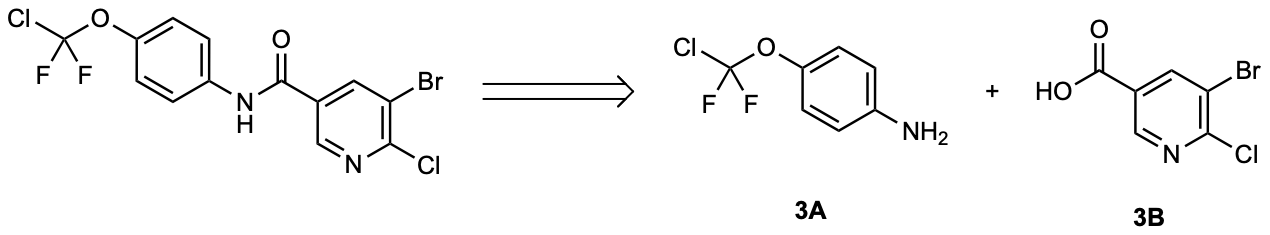}}} \\
\midrule
\multirow{3}{*}{V1} & Rank 1 & DCM, EDC, HCl at 15°C\\
 &Rank 2 & DCM, DMAP, EDC, HCl at 15°C \\
 &Rank 3 &  DCM, HOBt, EDC, HCl at 15°C \\
 \midrule
\multirow{6}{*}{V2 (fp)}  & Rank 1 & DMF (29.75 equiv.), H\textsubscript{2}O (2034.89 equiv.), HATU (1.64 equiv.), DIPEA (3.30 equiv.) at 37°C with reactants \textbf{3A} (1.06 equiv.), \textbf{3B} (1 equiv.) \\
 &Rank 2 &  DMF (21.93 equiv.), HATU (1.68 equiv.), DIPEA (3.10 equiv.) at 40°C with reactants \textbf{3A} (1.06 equiv.), \textbf{3B} (1 equiv.) \\
 &Rank 3 & DMF (18.34 equiv.), AcOEt (174.46 equiv.), HATU (1.69 equiv.), DIPEA (2.91 equiv.) at 39°C 
 with reactants \textbf{3A} (1.06 equiv.), \textbf{3B} (1 equiv.) \\
\midrule
 \multirow{5}{*}{V2 (graph)} & Rank 1 & H\textsubscript{2}O (162.66 equiv.), MeCN (111.64 equiv.), Na\textsubscript{2}CO\textsubscript{3} (2.65 equiv.) at 44°C with reactants \textbf{3A} (1.16 equiv.), \textbf{3B} (1 equiv.) \\
&Rank 2 &  MeCN (76.23 equiv.) at 43°C with reactants \textbf{3A} (1.38 equiv.), \textbf{3B} (1 equiv.) \\
 &Rank 3 &  MeCN (108.14 equiv.), Et\textsubscript{3}N (4.39 equiv.) at 44°C with reactants \textbf{3A} (1.16 equiv.), \textbf{3B} (1 equiv.) \\

\botrule
\end{tabular*}
\end{table}

\begin{table}[h!]
\caption{Recommended conditions for the second step in the proposed retrosynthesis of 
maribavir. The top 3 recommendations of the V2 condition recommender using fingerprint (fp) and graph representations of molecules are shown.}\label{table:maribavir_conditions}
\begin{tabular*}{\textwidth}{@{\extracolsep{\fill}}p{1.42cm}p{0.92cm}p{9.8cm}}
\toprule
\multicolumn{3}{c}{
\setlength{\fboxrule}{0pt}\fbox
{\includegraphics[width=0.6\textwidth]{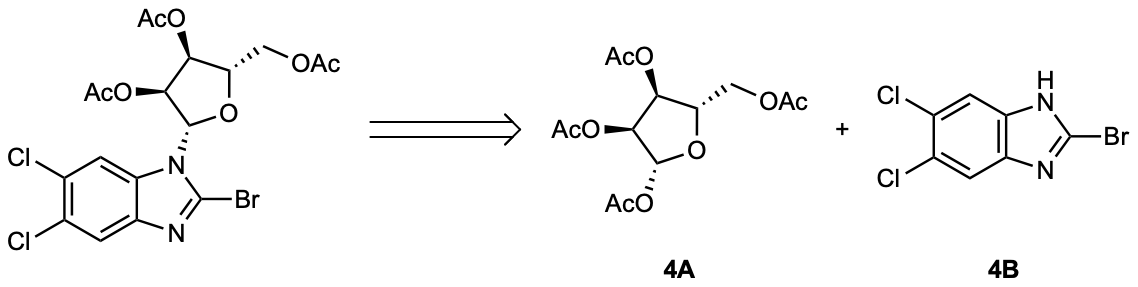}}} \\
\midrule
\multirow{3}{*}{V1} & Rank 1 & MeCN, TMSOTf, BSA at 44°C \\
 &Rank 2 & TMSOTf, BSA at 63°C \\
 &Rank 3 & MeCN, TMSOTf, bis(trimethylsilyl)acetamide at 45°C \\
 \midrule
 \multirow{5}{*}{V2 (fp)}  & Rank 1 & MeCN (166.38 equiv.) at 48°C with reactants \textbf{4A} (1 equiv.), \textbf{4B} (1 equiv.) \\
&Rank 2 & AcOEt (176.53 equiv.), MeCN (159.74 equiv.) at 51°C with reactants \textbf{4A} (1 equiv.), \textbf{4B} (1 equiv.) \\
 &Rank 3 & H\textsubscript{2}O (471.41 equiv.), AcOEt (148.80 equiv.), MeCN (143.76 equiv.) at 53°C with reactants \textbf{4A} (1 equiv.), \textbf{4B} (1 equiv.) \\
\midrule
 \multirow{5}{*}{V2 (graph)}  & Rank 1 &  THF (50.86 equiv.), \textit{n}-pentane (13.81 equiv.), \textit{t}-BuLi (1.43 equiv.) at -59°C 
 with reactants \textbf{4A} (1 equiv.), \textbf{4B} (1 equiv.) \\
&Rank 2 & O\textsubscript{2} (3.56 equiv.) at 62°C  with reactants \textbf{4A} (1 equiv.), \textbf{4B} (1 equiv.) \\
 &Rank 3 & DCM (51.48 equiv.), dimethyl sulfide (3.71 equiv.) at -6°C with reactants \textbf{4A} (1 equiv.), \textbf{4B} (1 equiv.) \\

\botrule
\end{tabular*}
\end{table}

\clearpage

\subsection{Re-running the automated Tree Builder with different search settings}

There are a number of known failure modes that may explain why routes are not found for all 75 targets under default search settings. For example, the tree search process may get stuck in some local optimum where it only explore paths following a particular disconnection it incorrectly thought to be highly promising; the template sets used (Pistachio and Reaxys) may not have templates corresponding to rarer (and arguably more interesting) transformations. One potential solution is straightforward: simply re-run the Tree Builder jobs with different settings, which only takes machine time. Like in the initial run, we queue up a large number of Tree Builder jobs via python scripts (available in the example folder mentioned at the start of this Section) and allow them to run to completion at the background. Results are automatically stored in the database for later inspection through the graphical interface.

We experimented three re-runs for demonstration purposes, where each run took about an hour to finish for all 75 targets, similar to the baseline. First, we change the maximum number of templates per expansion step to encourage the exploration of more diverse transformations. In an unconstrained setting (e.g., no limit on the number of reactions or chemicals explored) or during interactive planning, the maximum number of templates should be \emph{increased} so that each expansion may individually explore more transformations. In a constrained setting like the initial run where we cap the maximum number of chemicals explored to 5,000, however, it may be desirable to \emph{decrease} the maximum number of templates per expansion so that each may be explored more thoroughly. As an extreme numerical example, if the maximum number of templates per expansion is set to 5,000, the tree search may immediately reach the limit of 5,000 chemicals after a single expansion of the target and will terminate before it is able to consider any pathways of depth 2 or greater. In contrast, if the maximum number of templates per expansion is set to 10, the limit of 5,000 chemicals may be reached much more slowly, allowing for more thorough exploration of all of the 10 templates proposed for the target.

\begin{figure}[h!]
    \captionsetup[subfigure]{labelformat=empty}
    \begin{subfigure}[t]{1.0\textwidth}
        \includegraphics[scale=0.725]{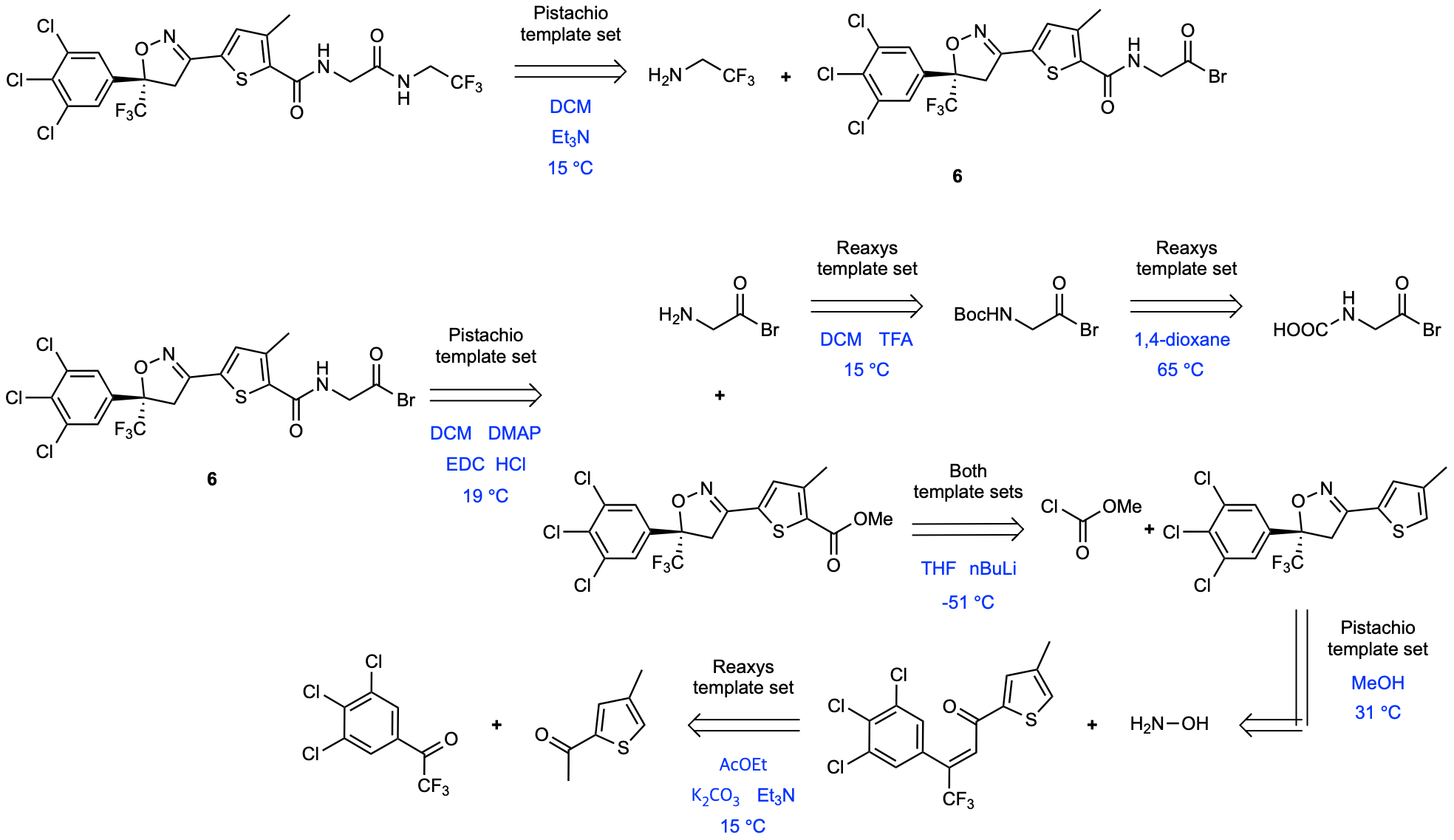}
        \caption{Lotilaner: \\ \small Cc1cc(C2=NO[C@@](c3cc(Cl)c(Cl)c(Cl)c3)(C(F)(F)F)C2)sc1C(=O)NCC(=O)NCC(F)(F)F}
    \end{subfigure}
    \hfill
    \vspace{1cm}
    \begin{subfigure}[t]{1.0\textwidth}
        \includegraphics[scale=0.725]{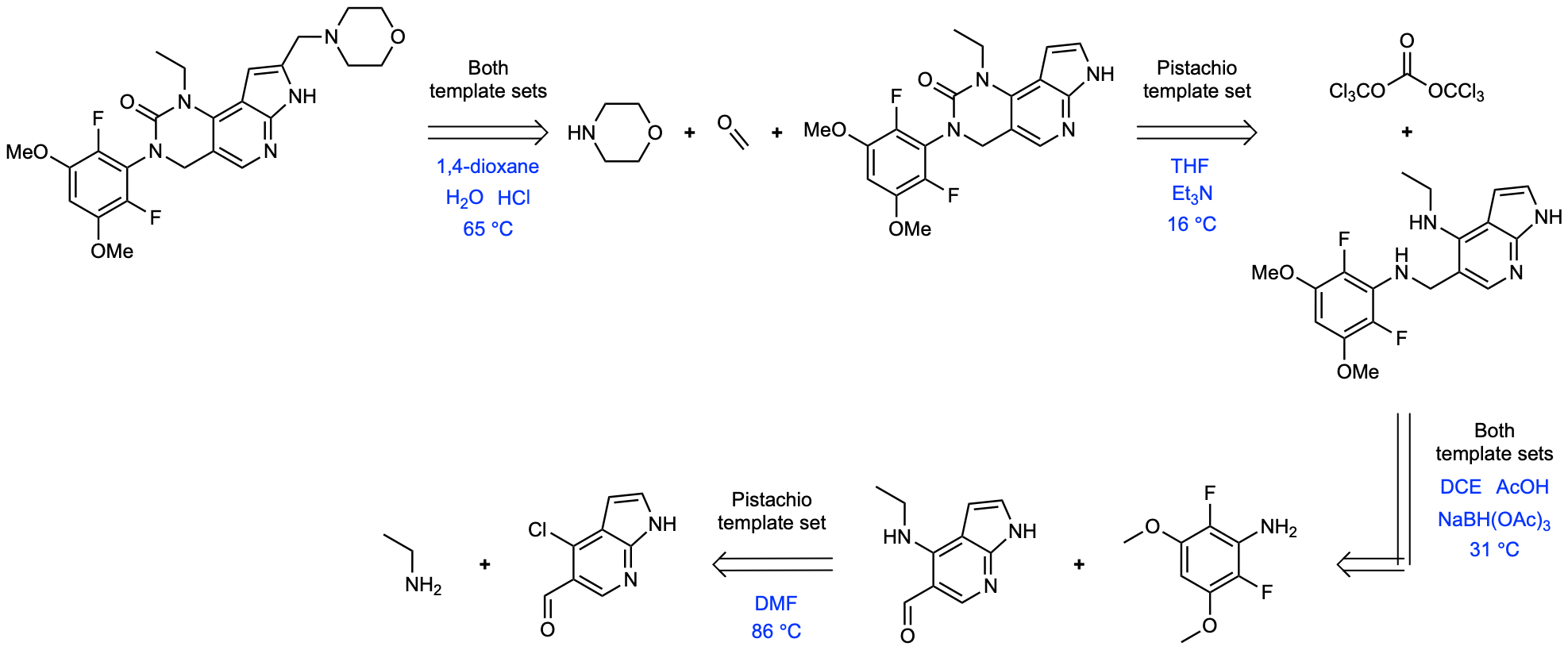}
        \caption{Pemigatinib: \\ \small CCN1C(=O)N(c2c(F)c(OC)cc(OC)c2F)Cc2cnc3[nH]c(CN4CCOCC4)cc3c21}
    \end{subfigure}
    \hfill
    \caption{Shortest retrosynthetic routes suggested by ASKCOS for lotilaner and pemigatinib when re-running with smaller numbers of templates per expansion step. Top-1 recommendations by the condition recommender (V1) are shown in blue below each arrow. Ions are written in salt form. Abbreviations are defined in Table \ref{table:abbreviations}.}
    \label{fig:fda_study_interesting_1}
\end{figure}

We re-ran with the maximum number of templates reduced to 100 while keeping the other settings exactly the same as in the baseline. Hypothetical routes could then be found for more targets. The shortest routes for lotilaner and pemigatinib are depicted in Figure \ref{fig:fda_study_interesting_1}. In both cases, the pathways returned by ASKCOS involve heterocycle forming steps. For lotilaner, the isoxazoline ring is prepared from an $\alpha$,$\beta$-unsaturated ketone with hydroxylamine; for pemigatinib, the cyclic urea in the fused ring system is prepared by an intramolecular cyclization with triphosgene. To re-iterate the typical disclaimer for template-based retrosynthetic proposals, these steps have some supporting precedents but may or may not be achievable experimentally  (e.g., due to the presumed stereoselectivity of the former). We can perform similar verification and further analysis as detailed in Section \nameref{verification_and_analysis}, and it is up to the user's discretion on whether to accept or reject proposed routes or steps.

\clearpage

\begin{figure}[h!]
    \captionsetup[subfigure]{labelformat=empty}
    \begin{subfigure}[t]{1.0\textwidth}
        \includegraphics[scale=0.725]{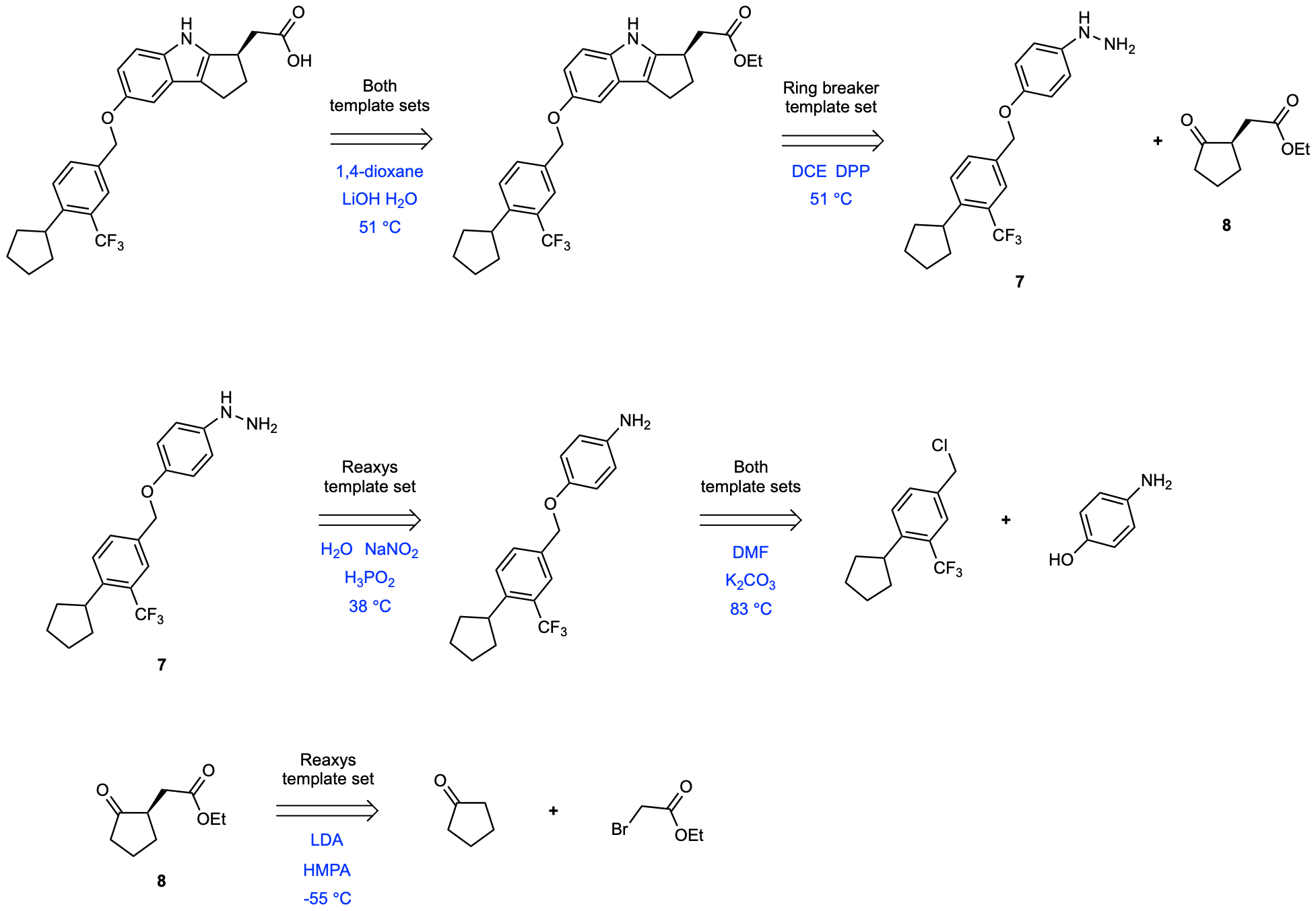}
        \caption{Etrasimod: \\ \small O=C(O)C[C@H]1CCc2c1[nH]c1ccc(OCc3ccc(C4CCCC4)c(C(F)(F)F)c3)cc21}
    \end{subfigure}
    \hfill
    \caption{Shortest retrosynthetic route suggested by ASKCOS for etrasimod when re-running with the addition of the ring breaker template set. Top-1 recommendations by the condition recommender (V1) are shown in blue below each arrow. Ions are written in salt form. Abbreviations are defined in Table \ref{table:abbreviations}.}
    \label{fig:fda_study_interesting_2}
\end{figure}

For the second re-run, we included the ring breaker~\citep{thakkar_ring_2020} template set in addition to Pistachio and Reaxys, while keeping all the other settings exactly the same as the baseline. While the ring breaker template set is derived from the Pistachio dataset~\citep{Pistachio} and overlaps with the Pistachio template set, it prioritizes (retrosynthetically) ring breaking transformation. As shown in Figure \ref{fig:fda_study_interesting_2}, in the shortest proposed route for etrasimod which was newly solved from this re-run, the 6-5-5 fused ring system is prepared via a Fischer indole synthesis between the phenylhydrazine and a chiral cyclic ketone. Templates corresponding to Fischer indole synthesis exist in the default template sets but may be better represented and prioritized by the ring breaker template relevance model.

\clearpage

\begin{figure}[h!]
    \captionsetup[subfigure]{labelformat=empty}
    \begin{subfigure}[t]{1.0\textwidth}
        \includegraphics[scale=0.725]{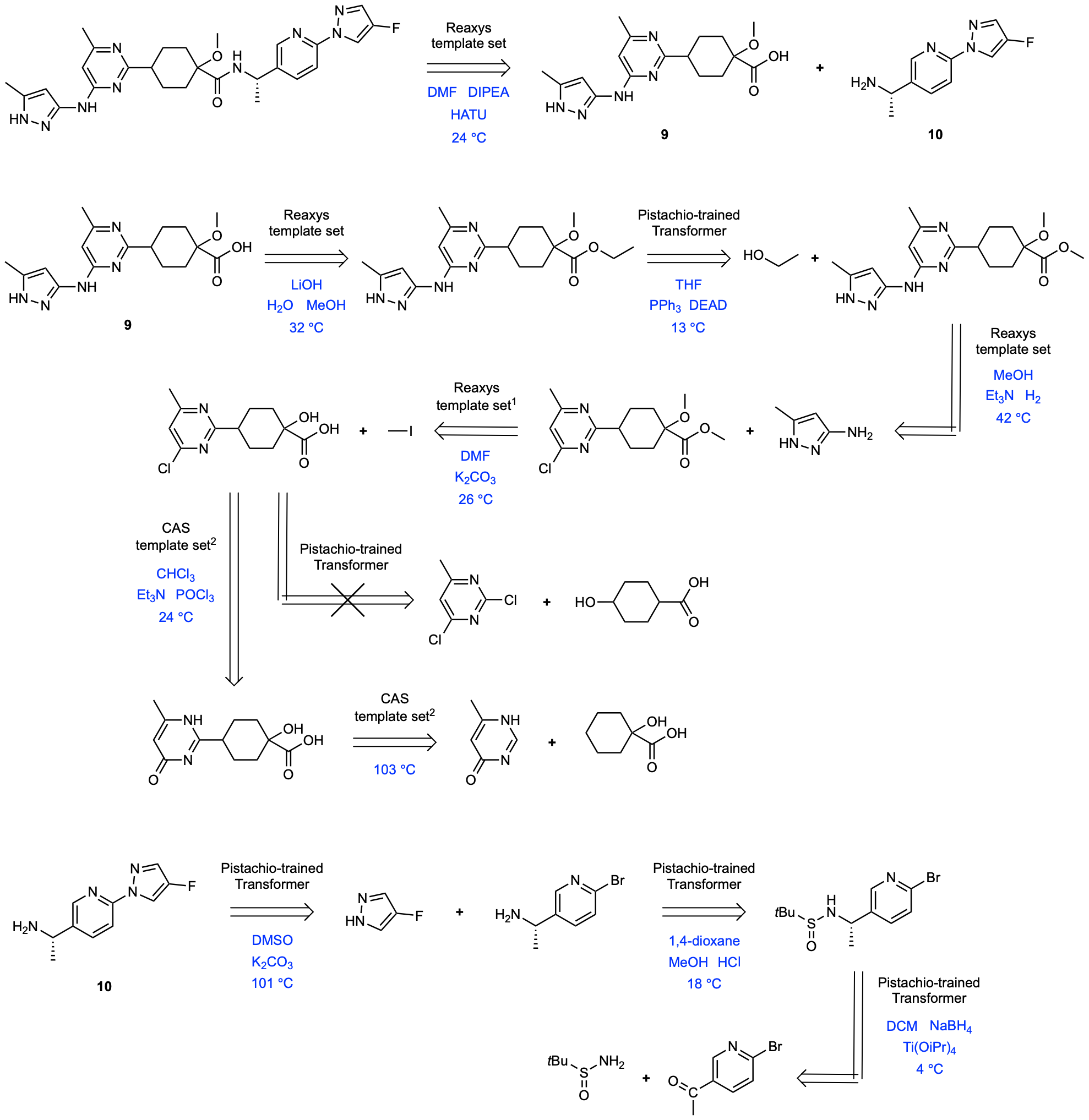}
        \caption{Pralsetinib: \\ \small COC1(C(=O)N[C@@H](C)c2ccc(-n3cc(F)cn3)nc2)CCC(c2nc(C)cc(Nc3cc(C)[nH]n3)n2)CC1}
    \end{subfigure}
    \hfill
    \caption{Shortest retrosynthetic route suggested by ASKCOS for pralsetinib when re-running with the addition of a template-free strategy. Top-1 recommendations by the condition recommender (V1) are shown in blue below each arrow. Ions are written in salt form. \textsuperscript{1}The original shortest route contains an alchemical step here (not shown) proposed by the Transformer model using a reactant with an extra nitrogen in the ring that has no way of disappearing in the product; excluding this reactant with one-click filters out all routes with this nonsensical step, and the replacement step from the new shortest route after filtering is shown. \textsuperscript{2}These two steps are obtained from manual expansion in the IPP upon rejecting the Transformer-proposed step marked with a crossed out arrow. Abbreviations are defined in Table \ref{table:abbreviations}.}
    \label{fig:fda_study_interesting_3}
\end{figure}

\clearpage

For the last re-run, we replaced the Pistachio-trained template relevance model with the Pistachio-trained Transformer, thereby combining a template-based expansion strategy with a template-free one; all the other settings remained exactly the same as the baseline. Template-free models are generally less constrained in formulation and \emph{may} generate more creative suggestions at the expense of chemical validity, helping the multi-step search converge towards buyable starting materials.

The shortest synthesis route for newly solved prasetinib is shown in Figure \ref{fig:fda_study_interesting_3} after filtering out a blatantly wrong intermediate with an extra nitrogen in the ring (not shown for succinctness). The inclusion of the Transformer model facilitates the solution of key intermediates \textbf{9} and \textbf{10}. For \textbf{9}, the Transformer model proposed a transesterification before making subsequent simplifying disconnections; while circuitous and seemingly unnecessary from a chemistry standpoint, from an algorithmic standpoint, the template-based model might have happened to generate better recommendations for the methyl ester than it could for the ethyl ester directly. The protection chemistry used in this route could be modified manually as a postprocessing step. There is another unusual suggestion from the Transformer in this route, which is marked with a cross. We chose to discard this step and re-expand manually in the IPP, but recommendations from the Pistachio and Reaxys template models did not propose to disconnect the two rings. We therefore made use of a proprietary template relevance model trained on data from the CAS Content Collection~\citep{CASContent}, which then successfully proposed a step based on analogy to metal-free cross coupling (e.g., as reported in~\citep{antonchick_direct_2013}, though missing the necessary NaN\textsubscript{3} and an external oxidant). Cross-referencing against literature precedents reveals that carboxylic acids seem to be out of the substrate scope; the proposal is perhaps more creative than feasible, but the underlying transformation can still be a starting point for a chemist to develop further. Notably, the interactive planning does not have to stop here; we can continue to modify, e.g., by (re-)expanding with different settings until we are satisfied with the route.

\subsection{Contextualizing the limitation of automatic planning}

\begin{figure}[h!]
    \captionsetup[subfigure]{labelformat=empty}
    \begin{subfigure}[t]{1.0\textwidth}
        \includegraphics[scale=0.725]{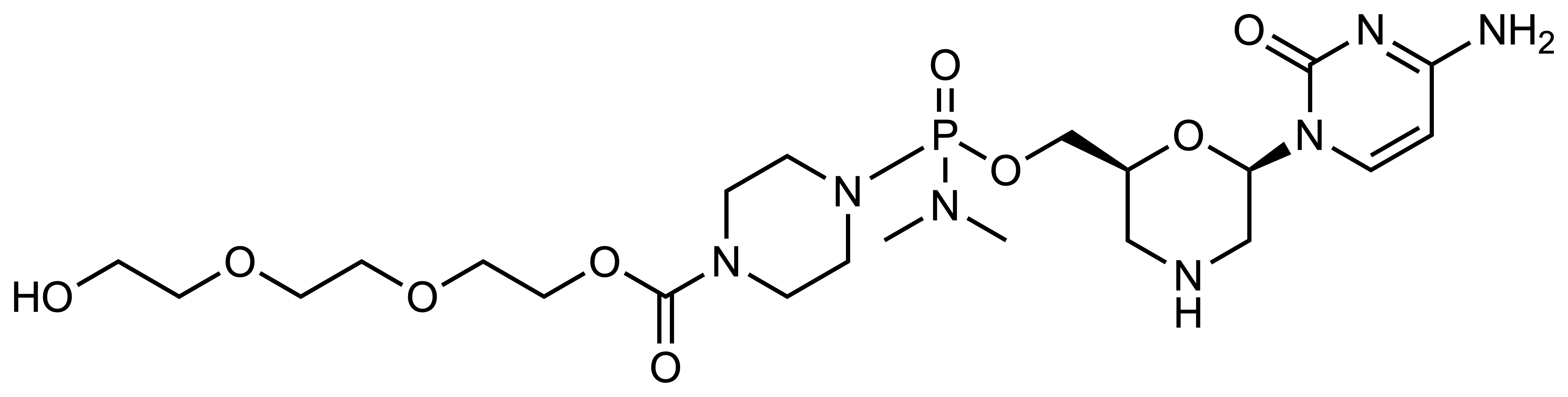}
        \caption{Casimersen: \\ \small CN(C)P(=O)(OC[C@@H]1CNC[C@H](n2ccc(N)nc2=O)O1)N1CCN \\ (C(=O)OCCOCCOCCO)CC1}
    \end{subfigure}
    \hfill
    \vspace{0.5cm}
    \begin{subfigure}[t]{1.0\textwidth}
        \includegraphics[scale=0.725]{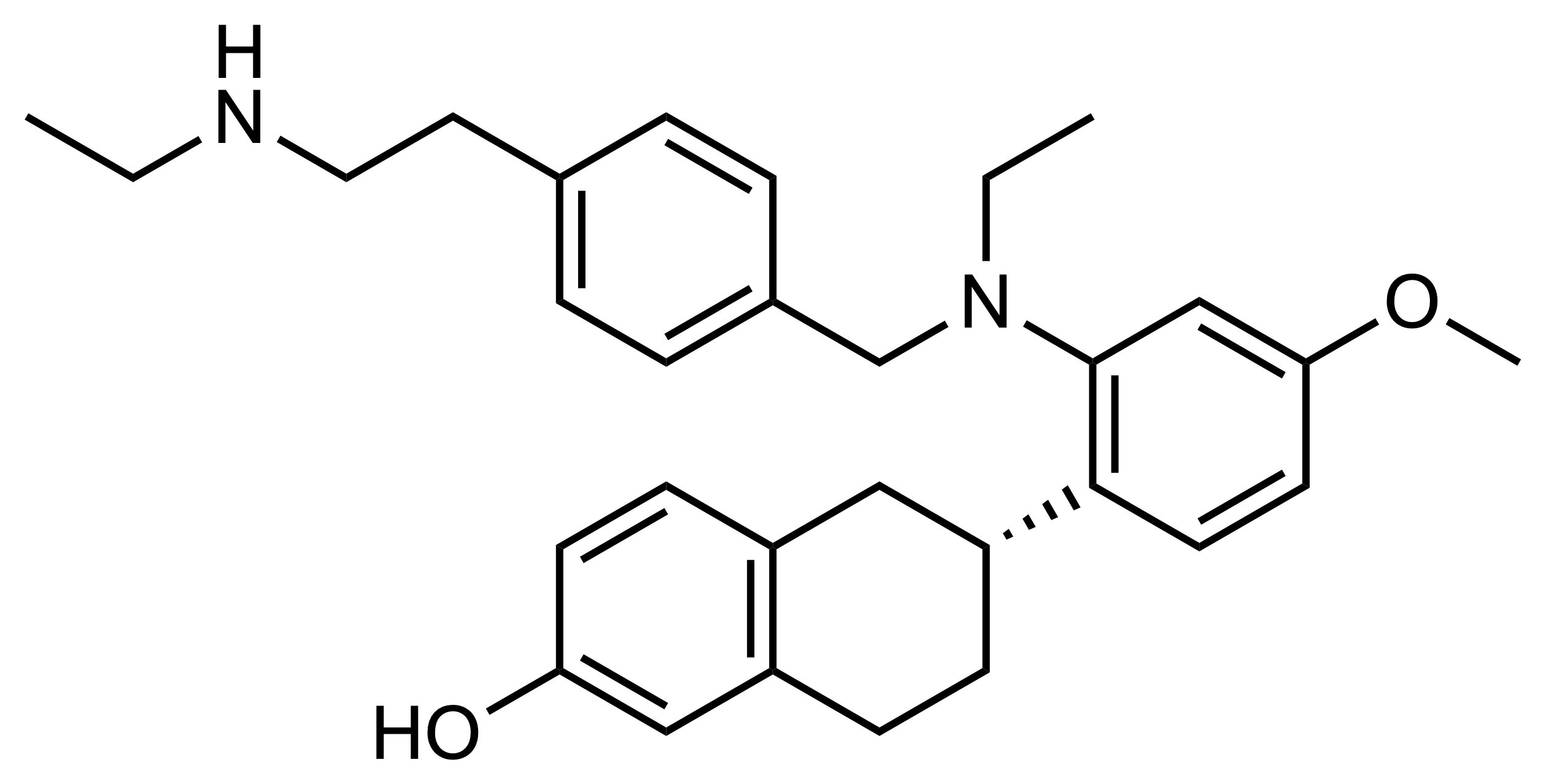}
        \caption{Elacestrant: \\ \small CCNCCc1ccc(CN(CC)c2cc(OC)ccc2[C@@H]2CCc3cc(O)ccc3C2)cc1}
    \end{subfigure}
    \hfill
    \vspace{0.5cm}
    \begin{subfigure}[t]{1.0\textwidth}
        \includegraphics[scale=0.725]{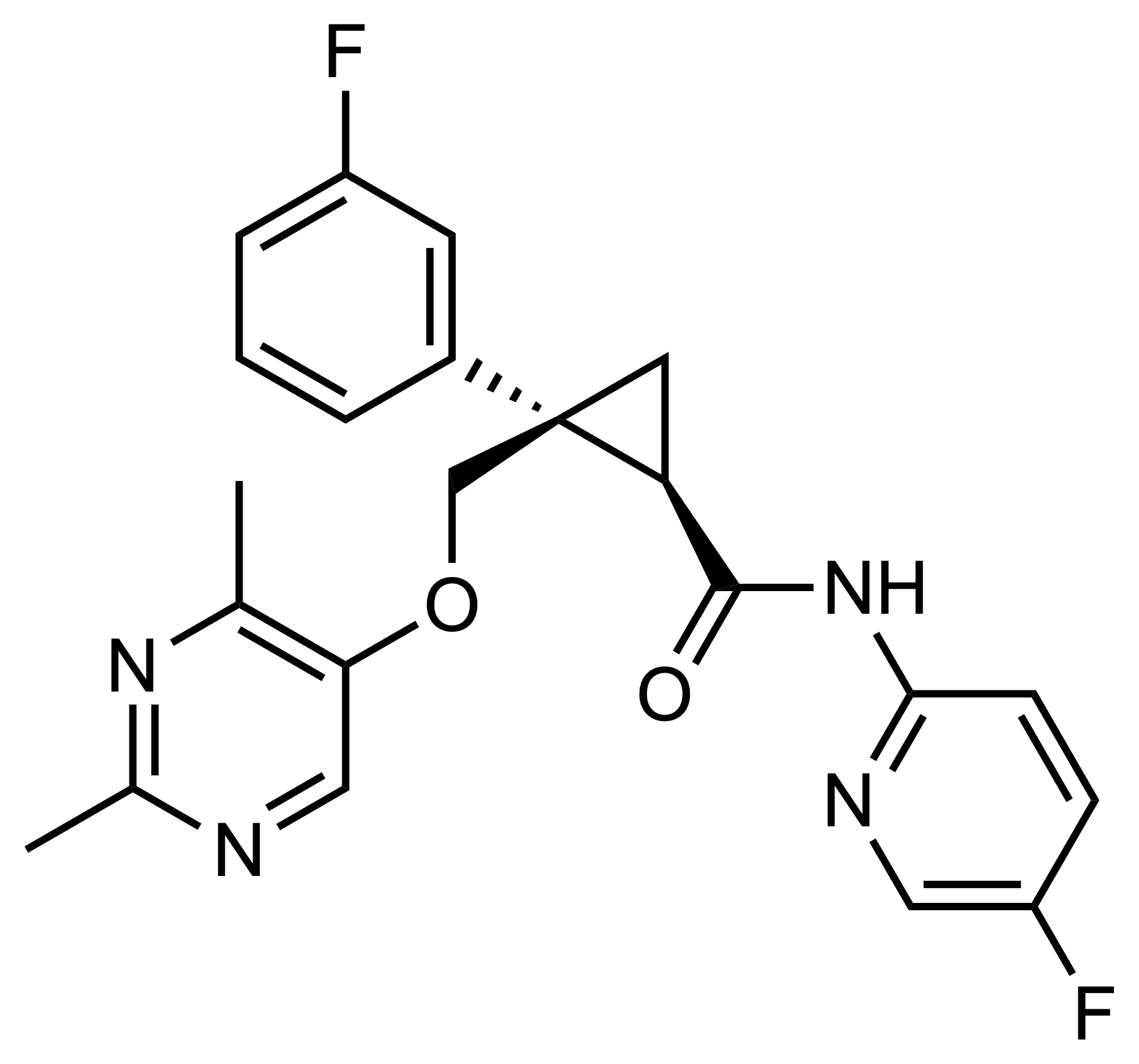}
        \caption{Lemborexant: \\ \small Cc1ncc(OC[C@@]2(c3cccc(F)c3)C[C@H]2C(=O)Nc2ccc(F)cn2)c(C)n1}
    \end{subfigure}
    \hfill
    \vspace{0.5cm}
    \begin{subfigure}[t]{1.0\textwidth}
        \includegraphics[scale=0.725]{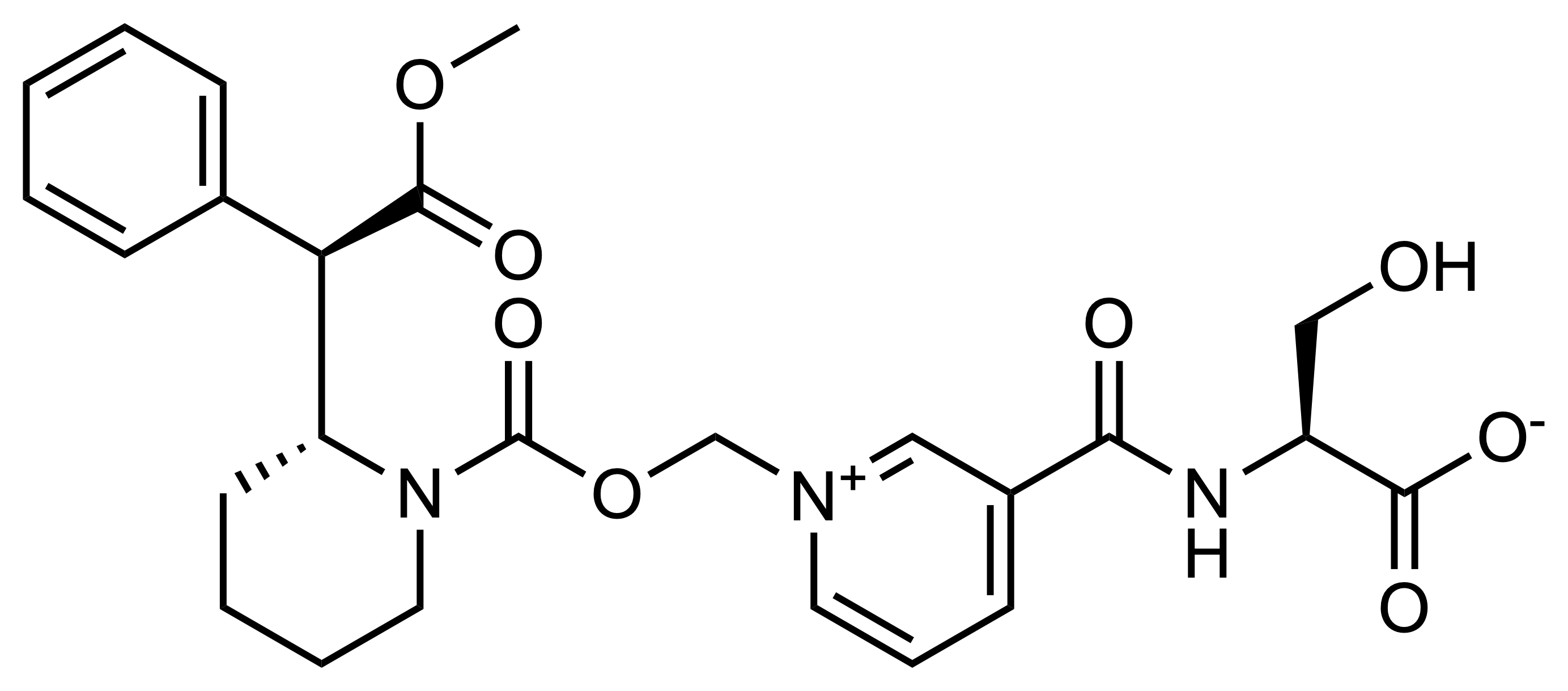}
        \caption{Serdexmethylphenidate: \\ \small COC(=O)[C@H](c1ccccc1)[C@H]1CCCCN1C(=O)OC[n+]1cccc(C(=O)N[C@@H](CO) \\ C(=O)[O-])c1}
    \end{subfigure}
    \hfill
    \vspace{0.5cm}
    \begin{subfigure}[t]{1.0\textwidth}
        \includegraphics[scale=0.725]{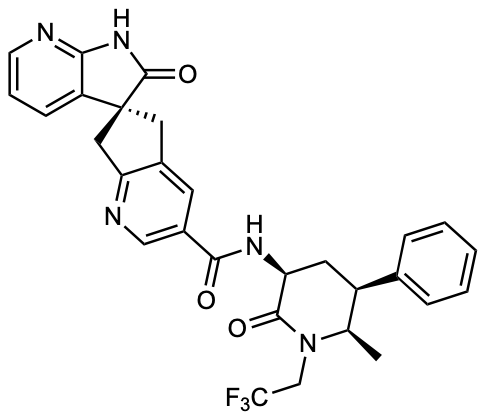}
        \caption{Ubrogepant: \\ \small C[C@@H]1[C@H](c2ccccc2)C[C@H](NC(=O)c2cnc3c(c2)C[C@@]2(C3)C(=O)Nc3ncccc32) \\ C(=O)N1CC(F)(F)F}
    \end{subfigure}
    \hfill
    \caption{Sample targets for which ASKCOS fails to propose any routes with buyable building blocks within the defined search criteria from all four automatic runs.}
    \label{fig:fda_study_fail}
\end{figure}

Even with a few different settings for the automated tree search, some targets still failed to yield retrosynthetic routes. Five such targets are shown in Figure \ref{fig:fda_study_fail}. Each may have a different explanation for \emph{why} the Tree Builder fails--missing chemistry, missing building blocks, or insufficient search time--which is not a particular focus for this case study. Interactive planning, as discussed in the Results Section, may be more helpful and insightful, as it offers the user full control over which intermediate to expand and thus the direction of retrosynthesis at a higher level. Moreover, since only one expansion step is performed per click, the user can explore different and more aggressive expansion settings (e.g., more templates, and/or more one-step strategies simultaneously, and/or more tolerant definition of building blocks). Last but not least, the interactive planning mode in ASKCOS is designed to cater even to the most general scenario of synthesis planning, by allowing the user to delete steps, manually add steps (by inputting SMILES as precursors), as well as to add step-wise notes and share routes with others.

\clearpage
\bibliography{sn-bibliography}